\documentclass[hidelinks,onefignum,onetabnum,nohypdvips]{siamart220329}

\usepackage{lipsum}
\usepackage{amsfonts}
\usepackage{graphicx}
\usepackage{epstopdf}
\usepackage{algorithmic}
\ifpdf
  \DeclareGraphicsExtensions{.eps,.pdf,.png,.jpg}
\else
  \DeclareGraphicsExtensions{.eps}
\fi

\usepackage[numbers]{natbib}

\newsiamremark{remark}{Remark}
\newsiamremark{hypothesis}{Hypothesis}
\crefname{hypothesis}{Hypothesis}{Hypotheses}
\newsiamthm{claim}{Claim}

\usepackage{amssymb}
\usepackage{amsmath}
\usepackage{mathtools}
\usepackage{subfig}

\usepackage{enumitem}

\usepackage{mathtools}

\renewcommand{\vec}[1]{\mathbf{#1}}
\newcommand{\vecgreek}[1]{\boldsymbol{#1}}
\newcommand{\mtx}[1]{\mathbf{#1}}

\DeclareMathOperator*{\argmax}{argmax}
\DeclareMathOperator*{\argmin}{argmin}

\DeclareMathOperator{\Tr}{Tr}
\newcommand{\mtxtrace}[1]{\Tr\left\lbrace{#1}\right\rbrace}
\newcommand{\expectation}[1]{\mathbb{E}\left[{{#1}}\right]} 
\newcommand{\expectationwrt}[2]{\mathbb{E}_{#2}\left[{{#1}}\right]}

\newcommand{\Ltwonorm}[1]{\left\Vert{{#1}}\right\Vert_2^2} 
\newcommand{\Frobnorm}[1]{\left\Vert{{#1}}\right\Vert_F^2}

\newcommand{\alphabold}{\boldsymbol{\alpha}}
\newcommand{\alphamintwo}{\widehat{\alphabold}}
\newcommand{\testerr}{{\mathcal{E}_{\sf test}}}
\newcommand{\testerrsignal}{{\mathcal{E}_{\sf test, sig}}}
\newcommand{\testerrnoise}{{\mathcal{E}_{\sf test,noise}}}
\newcommand{\testerrirred}{{\mathcal{E}_{\sf irred}}}
\newcommand{\testerrmissp}{{\mathcal{E}_{\sf test,missp}}}

\newcommand{\EE}{\mathbb{E}}
\newcommand{\UNIF}{\text{Unif}}

\newtheorem{example}{Example}

\newif\ifshowcomments

\showcommentstrue  

\headers{A Farewell to the Bias-Variance Tradeoff?}{Y.~Dar, V.~Muthukumar, R.~G.~Baraniuk}

\title{A Farewell to the Bias-Variance Tradeoff?\\  An Overview of the Theory of\\Overparameterized Machine Learning} 

\author{Yehuda Dar\thanks{Faculty of Computer and Information Science, Ben-Gurion University (\email{ydar@bgu.ac.il}).}
\and Vidya Muthukumar\thanks{Schools of Electrical and Computer Engineering, and Industrial and Systems Engineering,\protect\\ \hspace*{1.8em}Georgia Institute of Technology (\email{vmuthukumar8@gatech.edu}).}
\and Richard G.~Baraniuk\thanks{Electrical and Computer Engineering Department, Rice University (\email{richb@rice.edu}).}}

\usepackage{amsopn}

\ifpdf
\hypersetup{
  pdftitle={A Farewell to the Bias-Variance Tradeoff?  An Overview of the Theory of Overparameterized Machine Learning},
  pdfauthor={Y.~Dar, V.~Muthukumar, R.~G.~Baraniuk}
}
\fi

\begin{document}

\maketitle

\begin{abstract}
The last decade of progress in machine learning (ML), especially the deep learning era, has raised a number of scientific questions that challenge the longstanding dogma of the field. 
One of the most important riddles was the good empirical generalization of \emph{overparameterized} models.
Overparameterized models are highly complex with respect to the size of the training dataset, which enables them to perfectly fit (i.e., \textit{interpolate}) even noisy training data. 
Such interpolation of noisy data is traditionally associated with detrimental overfitting, and yet a wide range of interpolating models -- from simple linear models to deep neural networks -- have been observed to generalize remarkably well on fresh test data. 
Indeed, the discovery of the \emph{double descent} phenomenon has revealed that highly overparameterized models can improve over the best underparameterized model in test performance.

Understanding learning in this overparameterized regime required new theory and foundational empirical studies, even for the simplest case of the linear model.
The underpinnings of this understanding have been laid in foundational analyses of overparameterized linear regression and related statistical learning tasks, mostly published between 2018 and 2022, which resulted in precise analytic characterizations of double descent.
This paper provides an overview of the \emph{theory of overparameterized ML} (henceforth abbreviated as TOPML) by focusing on explaining the most foundational findings through a statistical signal processing perspective\footnote{A preliminary version of our paper appeared as an arXiv preprint in 2021: \url{https://arxiv.org/pdf/2109.02355v1}.}. 
We emphasize the unique aspects that define the TOPML research area as a subfield of modern ML theory and outline interesting open frontiers that remain.
\end{abstract}

\section{Introduction}
\label{sec:introduction}
Deep learning techniques have revolutionized the way many engineering and scientific problems are addressed, establishing the data-driven approach as a leading choice for practical success. 
Contemporary deep learning methods are large-scale, sophisticated versions of classical machine learning (ML) settings that were previously restricted by limited computational resources and insufficient availability of training data.
One prominent contemporary practice is to learn a highly complex deep neural network (DNN) from a set of training examples that, while itself large, is quite small \emph{relative} to the number of parameters in the DNN.
While such \emph{overparameterized} DNNs are common in the state-of-the-art ML practice, the fundamental reasons for this practical success were unclear.
Particularly mysterious were two empirical observations: i)~the explicit benefit (in generalization) of adding more parameters into the model, and ii)~the ability of these models to generalize well despite perfectly fitting even noisy training data.
These observations endure across diverse architectures in modern ML --- while the most high-profile of these observations was made for complex, state-of-the-art DNNs~\citep{zhang2017understanding,nakkiran2019deep}, they have also been unearthed in far simpler model families including wide neural networks~\cite{neyshabur2014search,belkin2019reconciling,spigler2019jamming,geiger2020scaling,advani2020high}, boosting and random forests~\cite{schapire1998boosting,wyner2017explaining}, kernel methods~\cite{belkin2018understand} and, subsequently, in linear models~\cite{belkin2019reconciling}.

In this paper, we survey the \emph{theory of overparameterized machine learning} (henceforth abbreviated as TOPML) that establishes foundational mathematical principles underlying phenomena related to \textit{interpolation} (i.e., perfect fitting) of training data. 
We will shortly provide a formal definition of overparameterized ML, but we first describe here some salient properties that a model must satisfy to qualify as overparameterized.
First, such a model must be highly complex such that interpolation of training data can be achieved; for linear models this typically takes the form of a model whose number of independently tunable (learnable) parameters\footnote{The number of parameters is an accepted measure of learned model complexity for simple cases such as ordinary least squares regression in the underparameterized regime and underlies classical complexity measures like Akaike's information criterion~\citep{akaike1998information}. For overparameterized models, the correct definition of learned model complexity is an open question at the heart of TOPML research.
See Section \ref{subsec:Open Question - complexity definition} for a detailed exposition.} 
is higher than the number of examples in the training dataset.
Second, the model must be trained to interpolate the training data; thus, any explicit or implicit regularization used in training must not prevent interpolation.

DNNs are popular instances of overparameterized models. 
They are highly complex and often trained to interpolate the training examples (i.e., achieve zero training error); such DNN training is commonly carried out without explicit regularization (see, e.g., \cite{zhang2017understanding,nakkiran2019deep}).
The training data is usually considered to be noisy realizations from an underlying data class (i.e., noisy data model). 
Hence, interpolating models perfectly fit both the underlying data and the noise in their training examples. 
Conventional statistical learning has always associated such perfect fitting of noise with poor generalization performance (e.g., \cite{friedman2001elements}); hence, it is remarkable that these
interpolating solutions often generalize well to new test data beyond the training dataset.

\begin{figure}
	\begin{center}
		{\includegraphics[width=\columnwidth]{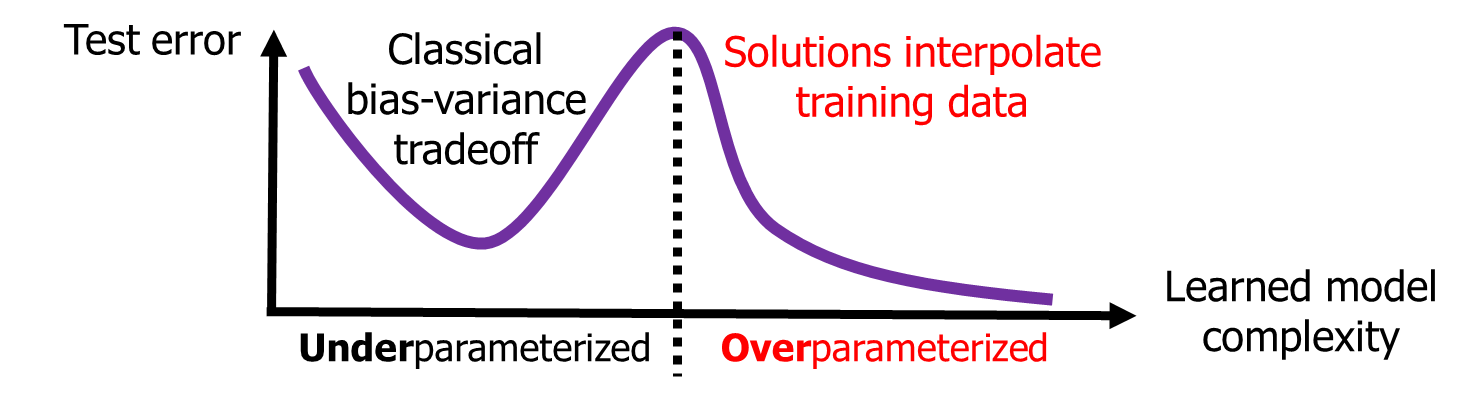}}
		\caption{Double descent of test error (i.e., generalization error) with respect to the complexity of the learned model. TOPML studies often consider settings in which the learned model complexity is expressed as the number of (independently tunable) parameters in the model. In this particular qualitative demonstration, the global minimum of the test error is achieved by maximal overparameterization. Note that overparameterization is not necessarily beneficial; Sections \ref{sec:Interpolation in Supervised Learning}-\ref{sec:regression} explain the learning problem aspects that promote beneficial overparameterization. }
		\label{fig:double_descent_conceptual_diagram}
	\end{center}
	\vspace*{-5mm}
\end{figure}

The observation that state-of-the-art neural networks tend to operate in the overparameterized regime, where they are capable of interpolating noise, was first elucidated in pioneering experiments in \cite{neyshabur2014search,zhang2017understanding}.
Moreover, it was shown in \cite{neyshabur2014search} that overparameterized (wide) neural networks are capable of achieving zero training error and yet generalizing well.
These findings sparked widespread conversation across deep learning practitioners and theorists, and inspired several new research directions in deep learning theory.
Although a comprehensive theory of overparameterized DNNs remains elusive, the discovery of analogous behavior in simpler models enabled a foundational theoretical program whose core results for linear models are now well developed.
Indeed, the way for foundational TOPML research was largely paved by the discovery of similar empirical behavior in far simpler parametric model families~\citep{belkin2019reconciling,geiger2020scaling,spigler2019jamming,advani2020high}.

In particular, the studies by Belkin et al.~\cite{belkin2019reconciling} and Spigler et al.~\cite{spigler2019jamming} explicitly evaluated the test error of these model families as a function of the number of tunable parameters and showed that it exhibits a remarkable \emph{double descent} behavior (pictured in Figure~\ref{fig:double_descent_conceptual_diagram}).
The first descent occurs in the underparameterized regime and is a consequence of the classical bias--variance tradeoff; indeed, the test error peaks when the learned model first becomes sufficiently complex to interpolate the training data.
More unusual is the behavior in the overparameterized regime: there, the test error is observed to decrease monotonically with the number of parameters, forming the second descent in the double descent curve.
\textbf{The double descent phenomenon suggests that the classical (U-shaped) bias--variance tradeoff, a cornerstone of conventional ML, is predictive only in the \emph{underparameterized} regime where the learned model is not sufficiently complex to interpolate the training data.}
A fascinating implication of the double descent phenomenon is that the best generalization error can be achieved by a highly overparameterized model even without explicit regularization, and despite perfect fitting of noisy training data.

Somewhat surprisingly, a good explanation for the double descent behavior did not exist until 2019 even for the simplest case of the overparameterized linear model --- for example, classical theories  that aim to examine the optimal number of variables in a linear regression task~(e.g., \cite{breiman1983many}) only considered the underparameterized regime. 
Particularly unexplained was the behavior of linear models that interpolate noise in data.
The aforementioned double descent experiments in~\cite{belkin2019reconciling,spigler2019jamming} inspired substantial TOPML research on the simple linear model\footnote{The pioneering experiments in~\cite{belkin2018understand} also inspired a parallel thread of mathematical research on harmless interpolation of noise by nonparametric and local methods beginning with~\cite{belkin2018overfitting,belkin2019does,liang2020just}. 
These models are not explicitly parameterized and are not surveyed in this paper, but possess possible connections to TOPML research.
}.
The theoretical foundations of this research originate in \cite{belkin2020two,bartlett2020benign,hastie2019surprises,muthukumar2020harmless} in early 2019. 
These studies provide precise analytical characterizations of the test error of minimum-norm solutions to the task of overparameterized linear regression with the squared loss (i.e., least squares regression, henceforth abbreviated as LS regression).
Of course, linear regression is a much simpler task than learning a deep neural network; however, it is a natural starting point for theory and itself of great independent interest for a number of reasons.
First, minimum $\ell_2$-norm solutions to LS regression do not include explicit regularization to avoid interpolating noise in training data, unlike more 
classical estimators such as ridge and the Lasso.
Second, the closed form of the minimum $\ell_2$-norm solution to LS regression is equivalent to the solution obtained by the popular gradient descent algorithm when the optimization process is initialized at zero~\citep{engl1996regularization}. 

\subsection{Contents of this paper}

In this paper, we survey TOPML with primary emphasis on the foundational wave of research in the first three to four years since 2018, while selectively discussing subsequent developments through 2026 where they clarify how the field has evolved. The maturation of this foundational program makes an elementary and accessible synthesis timely. Compared with the surveys~\citep{bartlett2021deep,belkin2021fit}, our treatment develops the central mechanisms through selected elementary derivations and a unique statistical signal processing perspective, and covers learning problems beyond regression and classification.

Formally, we define the TOPML research area as the subfield of ML theory where 
\begin{enumerate}
\item there is clear consideration of exact or near \textit{interpolation} of training data;
\item the \textit{learned} model complexity is high with respect to the training dataset size. Note that the complexity of the learned model is typically affected by (implicit or explicit) regularization aspects as a consequence of the learning process.
\end{enumerate}

Importantly, this definition highlights that while TOPML was inspired by observations in deep learning, several aspects of deep learning theory do not involve overparameterization.
More strikingly, TOPML is relevant to diverse model families other than DNNs, including relatively simple parametric and non-parametric model families. In this paper, our detailed treatment focuses primarily on feature-based linear models.  

Our interpolation-centered definition of overparameterization does not include every highly parameterized modern model. In particular, large language models (LLMs) are often pretrained in single-pass or few-pass regimes and need not attain near-zero loss on their large-scale pretraining dataset, even though substantial memorization can still occur. We therefore do not consider large-scale LLM pretraining as a canonical instance of TOPML.

Much of the first precise mathematical theory of TOPML was developed for linear regression; accordingly, our treatment mainly centers around a comprehensive survey of overparameterized linear regression.
However, TOPML goes well beyond the linear regression task.
Overparameterization naturally arises in diverse ML tasks, such as classification, e.g., \cite{muthukumar2021classification}, subspace learning for dimensionality reduction \cite{dar2020subspace},  data generation \cite{luzi2021interpolation}, and dictionary learning for sparse representations \cite{sulam2020recovery}.
In addition, overparameterization arises in various learning settings that are more complex than elementary fully supervised learning: unsupervised and semi-supervised learning~\cite{dar2020subspace}, transfer learning~\cite{dar2020double,dar2024common}, pruning of learned models~\cite{chang2021provable}, and others. 
We also survey foundational work on these topics.

This paper is organized as follows.
By expository mode, Sections \ref{sec:Interpolation in Supervised Learning}-\ref{sec:regression} have a tutorial style and include self-contained derivations, Sections \ref{sec:classification}-\ref{sec:additional problems} survey foundational and selected subsequent literature, and Section \ref{sec:Open Questions} synthesizes the main principles and identifies open frontiers. By topic, Sections \ref{sec:Interpolation in Supervised Learning}-\ref{sec:nonlinear} address regression, classification, and extensions to nonlinear models, whereas Sections \ref{sec:subspace learning}-\ref{sec:additional problems} address additional learning problems.

The first part of the paper (Sections \ref{sec:Interpolation in Supervised Learning}-\ref{sec:nonlinear}) focuses on the foundations of overparameterized learning as they pertain to fundamental ML tasks (regression and classification). 
In Section \ref{sec:Interpolation in Supervised Learning} we introduce the basics of interpolating solutions in overparameterized learning, as a machine learning domain that is outside the scope of the classical  bias--variance tradeoff. Section \ref{sec:Interpolation in Supervised Learning} also provides elementary simulations and transparent explanations for the bias--variance decomposition and the double descent phenomenon that may be of independent interest. 
In Section \ref{sec:regression} we overview foundational results on overparameterized linear regression. 
Here, we provide intuitive explanations of the fundamentals of overparameterized learning by taking a signal processing perspective. 
In Section \ref{sec:classification} we review foundational findings on overparameterized linear classification. 
In Section~\ref{sec:nonlinear} we provide a partial exposition on how these findings may generalize to more complex nonlinear models such as kernels and neural networks.

The second part of the paper surveys how overparameterization impacts more sophisticated learning problems beyond the basic tasks of regression and classification.
In Section \ref{sec:subspace learning} we overview foundational aspects of overparameterized subspace learning. 
In Section \ref{sec:additional problems} we discuss foundational research on overparameterized learning for generative models, transfer learning, model pruning, and dictionary learning of sparse representations.

Finally, in Section \ref{sec:Open Questions} we briefly discuss a couple of salient open frontiers in the theory of overparameterized ML. 

\section{Beyond the classical bias--variance tradeoff: The realm of interpolating solutions}
\label{sec:Interpolation in Supervised Learning}

We start by setting up basic notation and definitions for both underparameterized and overparameterized models.
Consider a supervised learning setting with $n$ training examples in the dataset ${\mathcal{D}=\{(\vec{x}_{i},y_{i})\}_{i=1}^{n}}$. 
We use a frequentist random-design formulation where the examples in $\mathcal{D}$ are random and reflect an unknown fixed (deterministic) functional relationship given by the function $f_{\sf true}: \mathcal{X} \rightarrow \mathcal{Y}$.
In its ideal, clean form, $f_{\sf true}$ maps a $d$-dimensional input vector ${\vec{x}\in\mathcal{X}\subseteq \mathbb{R}^{d}}$ to an output as 
\begin{equation}
\label{eq:ideal input-output mapping}
{y=f_{\sf true}(\vec{x})},    
\end{equation}
where the output domain $\mathcal{Y}$ depends on the specific problem (e.g., ${\mathcal{Y}=\mathbb{R}}$ in regression with scalar output values, and ${\mathcal{Y}=\{0,1\}}$ in binary classification). 
The mathematical notations and analysis in this paper consider one-dimensional output domain $\mathcal{Y}$, unless otherwise specified (e.g., in Section \ref{sec:subspace learning}). 
One can also perceive $f_{\sf true}$ as a \textit{signal}\footnote{In our frequentist random-design formulation, the target function $f_{\sf true}$ is treated as fixed, whereas the training dataset is random. Bayesian formulations, in which a prior is placed on $f_{\sf true}$ or its parameters, are outside our scope.} that should be estimated from the measurements in $\mathcal{D}$. 
Importantly, it is commonly the case that the output $y$ is degraded by noise. 
For example, a popular noisy data model for regression extends (\ref{eq:ideal input-output mapping}) into 
\begin{equation}
\label{eq:noisy input-output mapping}
{y=f_{\sf true}(\vec{x}) + {\epsilon}}
\end{equation}
where ${\epsilon \sim P_{\epsilon}}$ is a scalar-valued noise term that has zero mean and variance $\sigma_{\epsilon}^2$.
Moreover, the noise $\epsilon$ is independent of the input $\vec{x}\sim P_{\vec{x}}$. 

The input vector can be perceived as a random vector $\vec{x}\sim P_{\vec{x}}$ that, together with the unknown function $f_{\sf true}$ and the relevant noise model, induces a probability distribution $P_{\vec{x},y}$ over ${\mathcal{X}\times\mathcal{Y}}$ for the input-output pair 
$(\vec{x},y)$. 
Moreover, the $n$ training examples $\{(\vec{x}_{i},y_{i})\}_{i=1}^{n}$ in ${\mathcal{D}}$ are usually considered to be independent and identically distributed (i.i.d.)~random draws from $P_{\vec{x},y}$. 

The goal of supervised learning is to produce a function $f:\mathcal{X}\rightarrow\mathcal{Y}$ such that $f(\vec{x})$ constitutes a good estimate of the output $y$ for a new example ${(\vec{x},y)\sim P_{\vec{x},y}}$. 
This function $f$ is learned from the $n$ training examples in the dataset $\mathcal{D}$; consequently, a function $f$ that performs well on a new ${(\vec{x},y)\sim P_{\vec{x},y}}$ beyond the training dataset is said to \textit{generalize well}.
To quantify this performance, consider an error function ${\mathcal{E}: \mathcal{Y}\times\mathcal{Y} \rightarrow \mathbb{R}_{\ge 0}}$ that can evaluate the error $\mathcal{E}( f(\vec{x}), y )$ between the output estimate $f(\vec{x})$ and the output value $y$ from the data model (\ref{eq:noisy input-output mapping}) in the noisy case or (\ref{eq:ideal input-output mapping}) in the noiseless case. 
Then, the generalization performance of a function $f$ is evaluated using the \textit{test error}
\begin{equation}
\label{eq:test error - general definiton}
\mathcal{E}_{\sf test} (f) = \expectationwrt{ \mathcal{E}( f(\vec{x}), y )}{\vec{x},y}
\end{equation}
where $(\vec{x},y)\sim P_{\vec{x},y}$ is a random test data example.
During learning, one does not have the probability distribution $P_{\vec{x},y}$ but only $n$ training examples ${\mathcal{D}=\{(\vec{x}_{i},y_{i})\}_{i=1}^{n}}$ that were drawn i.i.d. from it. Accordingly, the test error expression (\ref{eq:test error - general definiton}) cannot be directly optimized during training.

Let us formulate the learning process. 
For this, a \text{loss function} ${L: \mathcal{Y}\times\mathcal{Y} \rightarrow \mathbb{R}_{\ge 0}}$ is chosen as a tractable surrogate for the error function ${\mathcal{E}: \mathcal{Y}\times\mathcal{Y} \rightarrow \mathbb{R}_{\ge 0}}$ of interest; for example, $L$ is usually convex in one of its arguments, while this may not be the case for $\mathcal{E}$ (e.g., in the case of classification where the $0-1$ error function, which is the binary indication of wrong label prediction, is non-convex). 
Then, the learning of $f$ is done by minimizing the $\textit{training loss}$, given by
\begin{equation}
\label{eq:training loss - general definiton}
\mathcal{L}_{\sf train} (f) = \frac{1}{n} \sum_{i=1}^{n}{ L ( f(\vec{x}_{i}), y_i ) }. 
\end{equation}
The training loss is simply the empirical average, over the training data, of the loss between the estimated output $f(\vec{x})$ and the actual output $y$.
The search for the function $f$ that minimizes $\mathcal{L}_{\sf train} (f) $ is done within a limited set $\mathcal{F}$ of functions that are induced by specific computational architectures. For example, in linear regression with scalar-valued output,  $\mathcal{F}$ denotes the set of all the linear functions from $\mathcal{X}=\mathbb{R}^{d}$ to $\mathcal{Y}=\mathbb{R}$. Accordingly, the training process can be written as 
\begin{equation}
\label{eq:learning optimization problem - general definiton}
\widehat{f} \in \argmin_{f\in\mathcal{F}} \mathcal{L}_{\sf train} (f) 
\end{equation}
where $\widehat{f}$ denotes a function with minimal training loss among the constrained set of functions $\mathcal{F}$. 
Note that the terminology in the field of statistical learning theory often refers to $\mathcal{F}$ as a hypothesis class and to (\ref{eq:learning optimization problem - general definiton}) as empirical risk minimization with respect to $\mathcal{F}$.

While the training loss $\mathcal{L}_{\sf train} (f)$ is used in the training process that learns the function $\widehat{f}$ in (\ref{eq:learning optimization problem - general definiton}), the training error 
\begin{equation}
\label{eq:training error - general definiton}
\mathcal{E}_{\sf train} (f) = \frac{1}{n} \sum_{i=1}^{n}{ \mathcal{E} ( f(\vec{x}_{i}), y_i ) }. 
\end{equation}
can be used to understand the proximity to or exact occurrence of perfect fitting of the training data based on the error function, although it does not necessarily participate directly in the learning objective. When the loss and error functions coincide, the training loss and training error are identical.
Note that we refer here to perfect fitting of training data purely based on the training error value. This training-data notion of perfect fitting is consistent with the ability of overparameterized models to perfectly fit their training data while generalizing well to test data.

Recall that the goal is to learn a function that generalizes well to test data beyond its training data. The best generalization performance corresponds to the lowest test error (\ref{eq:test error - general definiton}), which is achieved by the optimal function 
\begin{equation}
\label{eq:optimal mapping optimization problem - general definiton}
f_{\sf opt} = \argmin_{f:\mathcal{X}\rightarrow\mathcal{Y}} \mathcal{E}_{\sf test} (f).
\end{equation}
Note that the optimal function as defined above does not posit any restrictions on the function class.
For the purposes of this paper, we will consider the solution space $\mathcal{F}$ of the constrained optimization problem (\ref{eq:learning optimization problem - general definiton}), which underlies training, to be a parametric function class.
Further, if $\mathcal{F}$ includes the optimal function $f_{\sf opt}$, the learning architecture is considered as \textit{well specified}. 
Otherwise, the training procedure cannot possibly induce the optimal function $f_{\sf opt}$, and the learning architecture is said to be \textit{misspecified}.

The learned function $\widehat{f}$ depends on the specific training dataset $\mathcal{D}$ that was used in the training procedure (\ref{eq:learning optimization problem - general definiton}). 
If we use a larger function class $\mathcal{F}$ for training, we naturally expect the training error $\mathcal{E}_{\sf train}$ to improve.
However, what we are really interested in is the function's performance on independently drawn test data, which is unavailable at training time.
The corresponding test error, $\mathcal{E}_{\sf test} (\widehat{f})$, is also known as the \emph{generalization error} in statistical learning theory, as indeed this error reflects performance on ``unseen" examples beyond the training dataset.
For ease of exposition, throughout this paper we will consider the expectation of the test error over the training dataset, denoted by $\expectationwrt{\mathcal{E}_{\sf test} (\widehat{f})}{\mathcal{D}}$.
We note that several of the theoretical analyses that we survey actually provide expressions for the test error $\mathcal{E}_{\sf test}$ that hold with high probability over the training dataset (recall that both the input $\vec{x}_i$ and the output noise $\epsilon_i$ are random\footnote{As will become clear by reading the type of examples and results provided in this paper, the overparameterized regime necessitates a consideration of random design on training data to make the idea of good generalization even possible. Traditional analyses involving underparameterization or explicit regularization characterize the error of random design in terms of the fixed design (training) error, which is typically non-zero. In the overparameterized regime, the fixed design error is typically zero, but little can be said about the test error solely from this fact. See Section~\ref{sec:shouldweinterpolate} and the discussions on uniform convergence in the survey paper~\citep{belkin2021fit} for further discussion.}; in fact, independently and identically distributed across $i = 1,\ldots,n$).
See the discussion at the end of Section~\ref{sec:summary} for further details on the nature of these high-probability expressions.

The choice of loss function $L(\cdot,\cdot)$ and error function $\mathcal{E}(\cdot,\cdot)$ to evaluate training and test error, respectively, involves several nuances, and is both context and task-dependent\footnote{The choice is especially interesting for the classification task: for interesting classical and modern discussions on this topic, see~\citep{bartlett2006convexity,zhang2004statistical,ben2012minimizing,muthukumar2021classification,hsu2021proliferation}.}.
The simplest case is a least-squares regression task (with real-valued output space $\mathcal{Y}=\mathbb{R}$) where the squared loss function ${L ( f(\vec{x}), y )=(f(\vec{x})-y)^2}$ is the same as the squared error function ${\mathcal{E} ( f(\vec{x}), y )=(f(\vec{x})-y)^2}$. Informally, the prediction performance is evaluated by the same function for both training and test data. 
Such regression tasks possess the following important property: the optimal function $f_{\sf opt}$ as defined in Equation~\eqref{eq:optimal mapping optimization problem - general definiton} is the conditional mean of $y$ given $\vec{x}$, i.e., ${f_{\sf opt}(\vec{x})=\expectation{y | \vec{x}}=f_{\sf true}(\vec{x})}$  for both data models in (\ref{eq:ideal input-output mapping}) and (\ref{eq:noisy input-output mapping}).
Moreover, the expected test error $\expectationwrt{\mathcal{E}_{\sf test} (\widehat{f})}{\mathcal{D}}$ has the following bias-variance decomposition (similar decompositions are given, e.g., in \cite{geman1992neural,yang2020rethinking}):
\begin{equation}
\label{eq:bias variance decomposition - general definiton for squared error}
\expectationwrt{\mathcal{E}_{\sf test} \left(\widehat{f}\right)}{\mathcal{D}}= {{\sf bias}^{2}\left(\widehat{f}\right)} + {\sf var}\left(\widehat{f}\right) + {\mathcal{E}_{\sf irred}}.
\end{equation}
Above, the squared bias term is defined as 
\begin{equation}
\label{eq:bias variance decomposition - general definiton for squared error - bias}
{{\sf bias}^{2}\left(\widehat{f}\right)} \triangleq \expectationwrt{\left({\expectationwrt{\widehat{f}(\vec{x})}{\mathcal{D}} - f_{\sf opt}(\vec{x})}\right)^2}{\vec{x}},
\end{equation}
which is the expected squared difference between the estimates produced by the learned and the optimal functions (where the learned estimate is under expectation with respect to the training dataset).
For several popular estimators used in practice such as the ordinary least squares estimator (which is, in fact, the maximum likelihood estimator under Gaussian noise in an underparameterized setting with a unique solution to least squares), the expectation of the learned estimator is equal to the best model \emph{within the function class} $\mathcal{F}$; therefore, the bias can be thought of in these simple cases as the approximation-theoretic gap between this best-in-class model $\expectationwrt{\widehat{f}(\vec{x})}{\mathcal{D}}$ and the optimal model given by $f_{\sf opt}(\cdot)$.
Note that a mathematically well-defined expectation $\expectationwrt{\widehat{f}(\vec{x})}{\mathcal{D}}$ is required for having a well-defined bias-variance decomposition; we will discuss this in more detail in Section \ref{subsec:misspecification of learned models} and Appendix \ref{appendix:subsec:The Expected Learned Predictor is Undefined at the interpolation threshold}.

The variance term is defined as 
\begin{equation}
\label{eq:bias variance decomposition - general definiton for squared error - variance}
{\sf var}\left(\widehat{f}\right) \triangleq \mathbb{E}_{\vec{x}}\expectationwrt{\Big( \widehat{f}(\vec{x}) - \expectationwrt{\widehat{f}(\vec{x})}{\mathcal{D}} \Big)^2}{\mathcal{D}},
\end{equation}
which reflects how much the estimate can fluctuate due to randomness in the dataset used for training.
The irreducible error term is defined as 
\begin{equation}
\label{eq:bias variance decomposition - general definiton for squared error - irreducible error}
\mathcal{E}_{\sf irred} = \expectationwrt{(y - f_{\sf opt}(\vec{x}) )^2}{\vec{x},y}
\end{equation}
which quantifies the portion of the test error that does not depend on the learned function (or the training dataset). 
The model in (\ref{eq:noisy input-output mapping}) implies that $f_{\sf opt}=f_{\sf true}$ and ${\mathcal{E}_{\sf irred}} = \sigma_{\epsilon}^2$.

The test error decomposition in (\ref{eq:bias variance decomposition - general definiton for squared error}) demonstrates a prominent and classical concept in conventional statistical learning: the \textit{bias-variance tradeoff} as a function of the ``complexity" of the function class (or set of functions) $\mathcal{F}$.
The idea is that increased model complexity will \emph{decrease} the bias, as better approximations of $f_{\sf opt}(\cdot)$ can be obtained; but will \emph{increase} the variance, as the magnitude of dataset-induced variability around the best-in-class model increases.
Accordingly, the function class $\mathcal{F}$ is chosen to minimize the sum of the bias (squared) and its variance; in other words, to optimize the bias-variance tradeoff.
Traditionally, this measure of complexity is given by the number of free parameters in the model\footnote{The number of free parameters is an intuitive measure of the \emph{worst-case} capacity of a model class, and has seen fundamental relationships with model complexity in learning theory~\citep{vapnik2013nature} and classical statistics~\citep{akaike1998information} alike. In the absence of additional structural assumptions on the model or data distribution, this capacity measure is tight and can be matched by fundamental limits on performance. 
There are several situations in which the number of free parameters is not reflective of the true model complexity that involve additional structure posited both on the true function~\citep{candes2006robust,bickel2009simultaneous} and the data distribution~\citep{bartlett2005local}. As we will see through this paper, these situations are intricately tied to the double descent phenomenon.}.
In the underparameterized regime, where the learned model complexity is insufficient to interpolate the training data, this bias-variance tradeoff is ubiquitous in examples across model families and tasks. However, as we will see, the picture is significantly different in the overparameterized regime where the learned model complexity is sufficiently high to allow interpolation of training data.

\subsection{Underparameterized vs.~overparameterized regimes: Examples for linear models in feature space}
\label{subsec:Generalization in Underparameterized vs Overparameterized Regimes}
We now illustrate generalization performance using two examples of parametric linear function classes $\{\mathcal{F}_p\}_{p \geq 1}$ that are inspired by popular models in statistical signal processing.
Both examples rely on transforming the $d$-dimensional input $\vec{x}\in\mathcal{X}$ into a $p$-dimensional feature vector $\vecgreek{\phi}$ that is used for the linear mapping to output.
Namely, for a fixed feature map $\varphi: \mathcal{X} \rightarrow \mathbb{R}^p$, a feature vector is defined as $\vecgreek{\phi}\triangleq \varphi\left({\vec{x}}\right)$, and the learned function $\widehat{f}(\cdot)$ corresponding to learned parameters $\widehat{\vecgreek{\alpha}}\in\mathbb{R}^{p}$ is given by 
\begin{equation}
\label{eq:learned linear mapping of feature vector}
\widehat{f}(\vec{x}) = \vecgreek{\phi}^{T} \widehat{\vecgreek{\alpha}}.
\end{equation}
On the other hand, as we will now see, the two modeling examples differ in their composition of feature map as well as the relationship between the input dimension $d$ and the feature dimension $p$.

\begin{example}[Linear feature map]
\label{example:function class is linear mappings with p parameters}
Consider an input domain $\mathcal{X}=\mathbb{R}^{d}$ with $d \geq n$, and an orthonormal basis ${\vec{u}_1,\dots,\vec{u}_d \in \mathbb{R}^{d}}$ for $\mathbb{R}^{d}$.
We define ${\mathcal{F}_{p}^{\sf lin}(\{\vec{u}_j\}_{j=1}^{d})}$ as a class of linear functions with ${p\in\{1,\dots,d\}}$ parameters, given by 
\begin{align}
\label{eq:function class is linear mappings with p parameters}
&\mathcal{F}_{p}^{\sf lin}(\{\vec{u}_j\}_{j=1}^{d}) \triangleq \Bigg\{ f(\vec{x})=\sum_{j=1}^{p}{\alpha_j \vec{u}_j^T \vec{x} } ~\Bigg\vert~ \vec{x}\in\mathbb{R}^{d}; \alpha_1,\dots,\alpha_p\in\mathbb{R} \Bigg\}. 
\end{align}
In this case, we can also write $f(\vec{x})=\vecgreek{\alpha}^T \mtx{U}_p^T \vec{x}$ where ${\vecgreek{\alpha}\triangleq [\alpha_1,\dots,\alpha_p]^T \in\mathbb{R}^p}$ and ${\mtx{U}_p\triangleq [\vec{u}_1,\dots,\vec{u}_p]}$ is a $d\times p$ matrix with orthonormal columns. We define the $p$-dimensional feature vector as $\vecgreek{\phi}\triangleq \varphi\left({\vec{x}}\right) \triangleq \mtx{U}_p^T \vec{x}$. Then, the training optimization problem (\ref{eq:learning optimization problem - general definiton}) constitutes least-squares (LS) regression and is given by  
\begin{align}
\label{eq:function class is linear mappings with p parameters - training optimization}
\widehat{\vecgreek{\alpha}} & \in \argmin_{\vecgreek{\alpha}\in\mathbb{R}^p} \sum_{i=1}^{n}{ ( \vecgreek{\alpha}^T \mtx{U}_p^T \vec{x}_{i} - y_i )^2 } \nonumber\\
& = \argmin_{\vecgreek{\alpha}\in\mathbb{R}^p} \Ltwonorm{ \mtx{\Phi} \vecgreek{\alpha} - \vec{y} }.
\end{align}
Here, $\mtx{\Phi}\triangleq \left[\varphi\left({\vec{x}_1}\right),\dots,\varphi\left({\vec{x}_n}\right)\right]^T$ is the $n\times p$ matrix whose $(i,j)$ entry is given by $\vec{u}_j^T \vec{x}_{i}$, and the training data output is given by the $n$-dimensional vector ${\vec{y}\triangleq [y_1,\dots,y_n]^T}$. 
\end{example}
\begin{example}[Nonlinear feature map]
\label{example:function class is nonlinear mappings}
~
Consider the input domain $\mathcal{X}=[0,1]^{d}$, which is the $d$-dimensional unit cube.
Let $\varphi_j:[0,1]^{d} \rightarrow \mathbb{R}$ for ${j=1,\dots,\infty}$ be a family of real-valued functions that are defined over the unit cube and orthonormal in function space, i.e., we have $\int_{\vec{z}\in[0,1]^{d}} \varphi_j (\vec{z})\varphi_k (\vec{z})d\vec{z} = \delta_{jk}$, where $\delta_{jk}$ denotes the Kronecker delta\footnote{One can consider a more general definition of orthonormality where $\int_{\vec{x}\in[0,1]^{d}} \varphi_j (\vec{x})\varphi_k (\vec{x})d\mu(\vec{x}) = \delta_{jk}$ and $\mu(\cdot)$ is the distribution over the input.}.
We define ${\mathcal{F}_{p}^{\sf nonlin}(\{{\varphi}_j\}_{j=1}^{\infty})}$ as a class of nonlinear functions with ${p\in\{1,\dots,\infty\}}$ parameters, given by 
\begin{align}
\label{eq:function class is nonlinear mappings with p parameters}
&\mathcal{F}_{p}^{\sf nonlin}(\{\varphi_j\}_{j=1}^{\infty}) \triangleq \Bigg\{ f(\vec{x})=\sum_{j=1}^{p}{\alpha_j \varphi_j(\vec{x}) } ~\Bigg\vert~ \vec{x}\in[0,1]^{d}; \alpha_1,\dots,\alpha_p\in\mathbb{R} \Bigg\}. 
\end{align}
In this case, we again define ${\vecgreek{\alpha}\triangleq [\alpha_1,\dots,\alpha_p]^T \in\mathbb{R}^p}$, but now the $p$-dimensional feature vector is defined as ${\vecgreek{\phi}\triangleq \varphi\left({\vec{x}}\right) \triangleq [\varphi_1(\vec{x}),\dots,\varphi_p(\vec{x})]^T}$. The training optimization in (\ref{eq:learning optimization problem - general definiton}) takes the form of 
\begin{align}
\label{eq:example 2 - training optimization}
\widehat{\vecgreek{\alpha}} &\in \argmin_{\alpha_1,\dots,\alpha_p \in\mathbb{R}} \sum_{i=1}^{n}{ \left( \sum_{j=1}^{p}{\alpha_j \varphi_j(\vec{x}_{i}) } - y_i \right)^2 } 
\nonumber\\
&= \argmin_{\vecgreek{\alpha}\in\mathbb{R}^p} \Ltwonorm{ \mtx{\Phi}\vecgreek{\alpha} - \vec{y} }.
\end{align}
Here, ${\mtx{\Phi}} \triangleq \left[\varphi\left({\vec{x}_1}\right),\dots,\varphi\left({\vec{x}_n}\right)\right]^T$ is the $n\times p$ matrix whose $(i,j)$ entry is  $\varphi_j(\vec{x}_{i})$, and the training data output is given by the $n$-dimensional vector ${\vec{y}\triangleq [y_1,\dots,y_n]^T}$. 
\end{example}

For Examples \ref{example:function class is linear mappings with p parameters} and \ref{example:function class is nonlinear mappings}, an optimal solution $\widehat{\vecgreek{\alpha}}$ to (\ref{eq:function class is linear mappings with p parameters - training optimization}) and (\ref{eq:example 2 - training optimization}), respectively, should satisfy 
\begin{equation}
    \label{eq:normal equations for feature learning}
    \mtx{\Phi}^T \mtx{\Phi} \widehat{\vecgreek{\alpha}} = \mtx{\Phi}^T \vec{y}
\end{equation}
which is known as the normal equations of least squares. Specifically, the optimality condition (\ref{eq:normal equations for feature learning}) stems from requiring equality to zero of the gradient (w.r.t.~$\vecgreek{\alpha}$) of the optimization objective of (\ref{eq:function class is linear mappings with p parameters - training optimization}), (\ref{eq:example 2 - training optimization}).

Let us consider overparameterized settings where $p$ is sufficiently large such that (\ref{eq:normal equations for feature learning}) and, therefore, (\ref{eq:function class is linear mappings with p parameters - training optimization}), (\ref{eq:example 2 - training optimization}) have infinitely many solutions. Mathematically, (\ref{eq:normal equations for feature learning}) can have infinite solutions due to rank deficiency of $\mtx{\Phi}^T \mtx{\Phi}$; specifically, for $p>n$, $\mtx{\Phi}^T \mtx{\Phi}$ is a $p\times p$ matrix whose rank is at most $n$. Assuming\footnote{A sufficient assumption is that $\mtx{\Phi}$ have full row
rank. For $p\geq n$, this condition holds almost surely when the rows of $\mtx{\Phi}$ are independent draws from a nondegenerate Gaussian distribution, as in the Gaussian feature models analyzed below.} that the training outputs vector $\vec{y}$ is in the range of the training feature matrix $\mtx{\Phi}$, then all the infinitely many solutions interpolate the training data and achieve zero training error (i.e., $\mathcal{E}_{\sf train} = 0$). 
Accordingly, we choose the minimum $\ell_2$-norm solution 
\begin{align}
\label{eq:example 1-2 - min-norm solution}
\widehat{\vecgreek{\alpha}} &= \left(\mtx{\Phi}^T \mtx{\Phi}\right)^+ \mtx{\Phi}^T \vec{y} \nonumber\\
&= \mtx{\Phi}^+ \vec{y}, 
\end{align}
where the superscript $+$ denotes the Moore–Penrose pseudoinverse. 
The minimum $\ell_2$-norm solution is straightforward to implement, either via direct pseudoinverse calculation and matrix-vector multiplication or via gradient descent~\citep{engl1996regularization}, and therefore is a natural starting point for the study of interpolating solutions.

Figures \ref{fig:Example1_aligned_beta} and \ref{fig:Example1_nonaligned_beta} demonstrate particular cases of solving LS regression using (\ref{eq:example 1-2 - min-norm solution}) and function classes as in Example \ref{example:function class is linear mappings with p parameters}. 
For both Figures \ref{fig:Example1_aligned_beta} and \ref{fig:Example1_nonaligned_beta}, the data model (\ref{eq:noisy input-output mapping}) takes the form of ${y=\vec{x}^{T} \vecgreek{\beta}+\epsilon}$ where $\vecgreek{\beta}\in\mathbb{R}^{d}$ is an unknown, deterministic (non-random) parameter vector, $d=128$ is the input space dimension, and the noise component is $\epsilon\sim\mathcal{N}(0,0.25)$. 
The number of training examples is $n=32$.
We denote by $\mtx{C}_d$ the $d\times d$ Discrete Cosine Transform (DCT) matrix whose columns are orthonormal and span $\mathbb{R}^d$. 
Then, we consider the input data $\vec{x} \sim \mathcal{N}(\vec{0},\mtx{\Sigma}_{\vec{x}})$ where ${\mtx{\Sigma}_{\vec{x}}=\mtx{C}_d \mtx{\Lambda}_{\vec{x}} \mtx{C}_d^T}$ and $\mtx{\Lambda}_{\vec{x}}$ is a $d\times d$ diagonal matrix with real non-negative entries. 
Consider any choice $1 \leq p \leq d$ for the feature dimension. 
By Example \ref{example:function class is linear mappings with p parameters}, $\vecgreek{\phi}\triangleq \mtx{U}_p^T \vec{x}$ is the $p$-dimensional feature vector where $\mtx{U}_p$ is a $d\times p$ real matrix with orthonormal columns that extract the features for the learning process.
Consequently, $\vecgreek{\phi} \sim \mathcal{N}(\vec{0},\mtx{\Sigma}_{\vecgreek{\phi}})$ where $\mtx{\Sigma}_{\vecgreek{\phi}}=\mtx{U}_p^T\mtx{C}_d \mtx{\Lambda}_{\vec{x}} \mtx{C}_d^T \mtx{U}_p$.

\subsection{The double descent phenomenon}

\begin{figure*}
    \centering
    \subfloat[]{\label{fig:Example1_trueDCT_featureDCT_betaAligned_truebeta_signaldom}\includegraphics[width=0.32\textwidth]{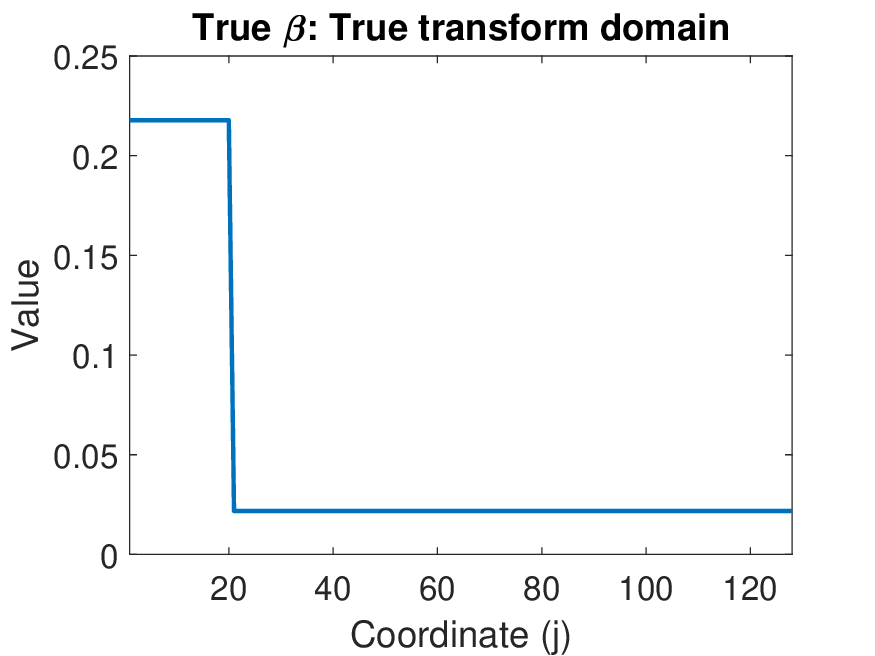}}
    \subfloat[]{\label{fig:Example1_trueDCT_featureDCT_betaAligned_truebeta_truetransdom}\includegraphics[width=0.32\textwidth]{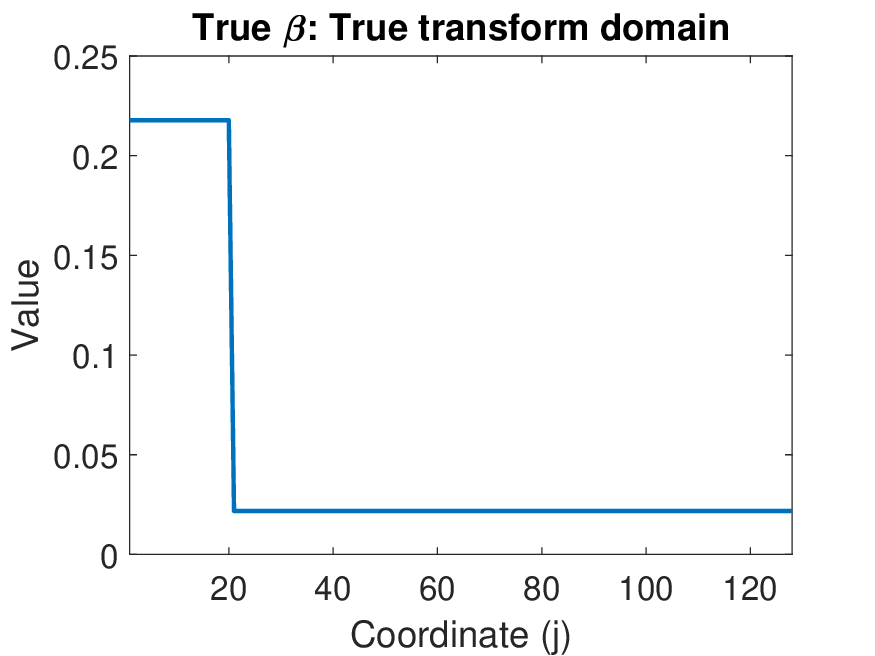}}
    \subfloat[]{\label{fig:Example1_trueDCT_featureDCT_betaAligned_InputCovarianceMatrix}\includegraphics[width=0.32\textwidth]{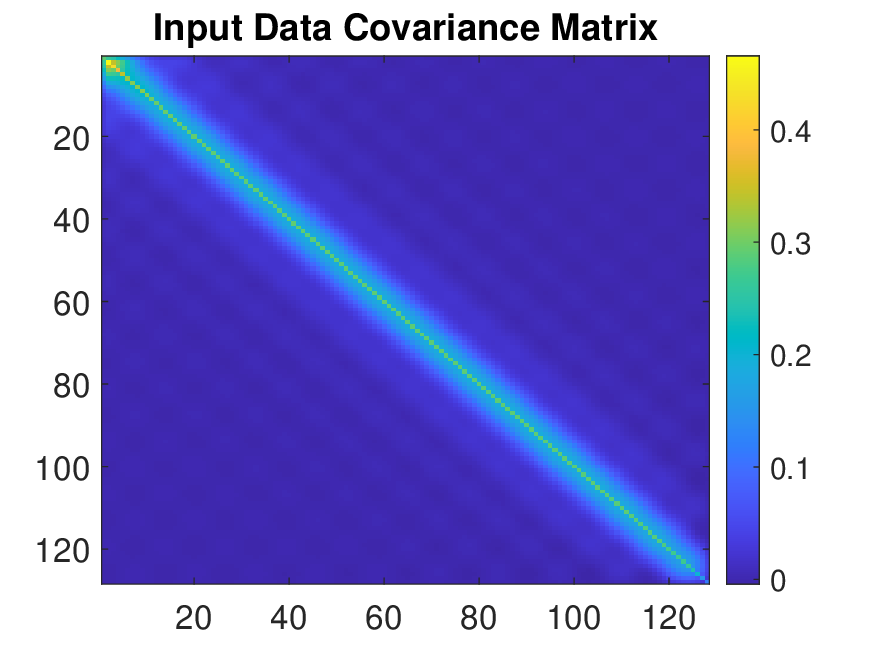}}
    \\
    \subfloat[]{\label{fig:Example1_trueDCT_featureDCT_betaAligned_curves}\includegraphics[width=0.32\textwidth]{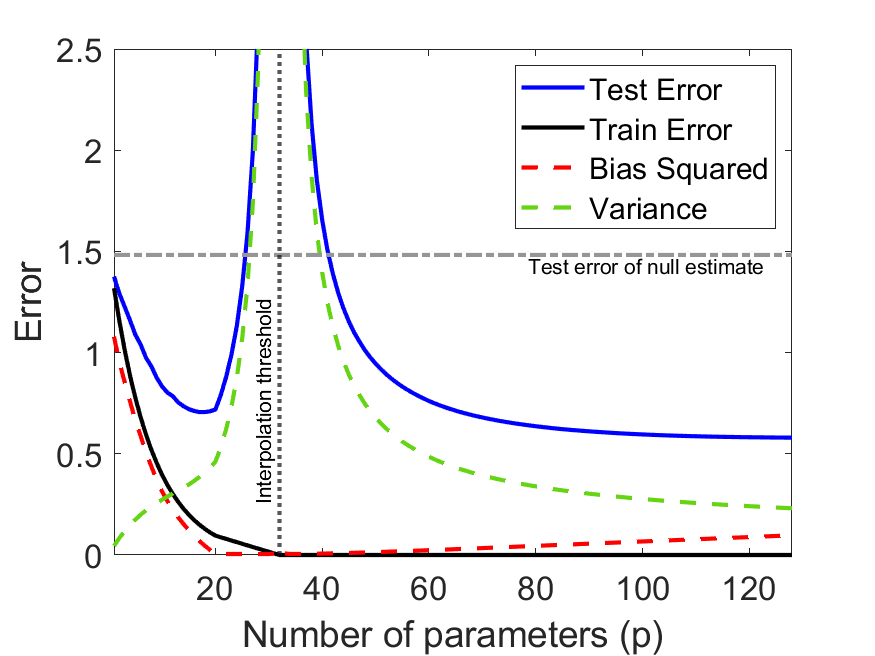}}
    \subfloat[]{\label{fig:Example1_trueDCT_featureDCT_betaAligned_truebeta_featuretransdom}\includegraphics[width=0.32\textwidth]{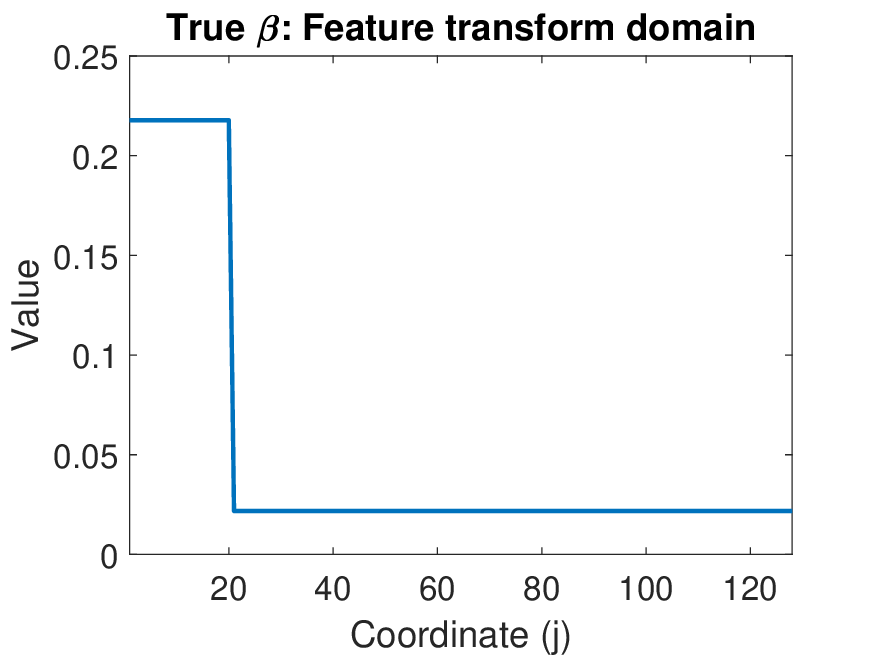}}
    \subfloat[]{\label{fig:Example1_trueDCT_featureDCT_betaAligned_truebeta_FeatureCovarianceMatrix}\includegraphics[width=0.32\textwidth]{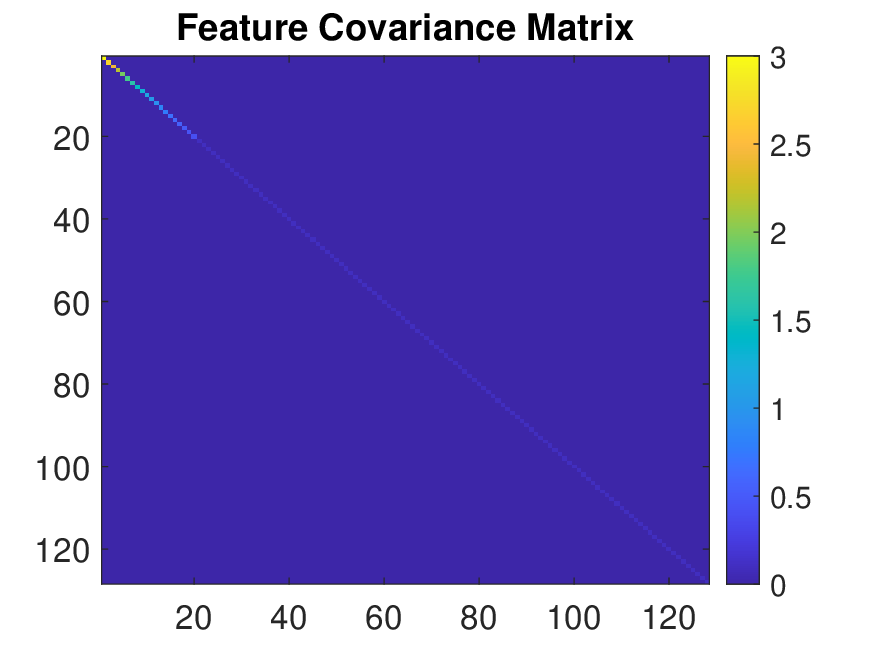}}
    \\
    \subfloat[]{\label{fig:Example1_trueDCT_featureHadamard_betaAligned_curves}\includegraphics[width=0.32\textwidth]{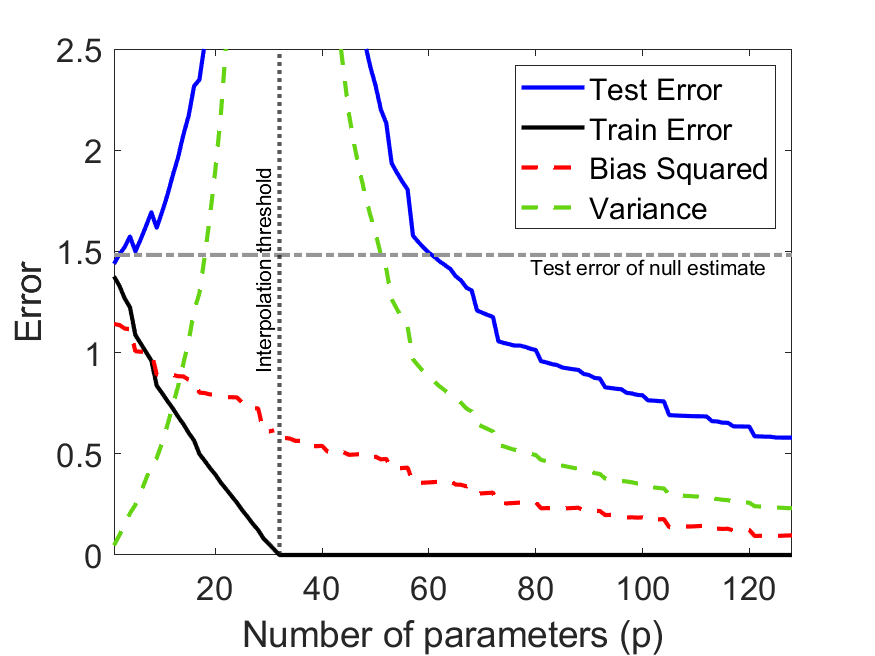}}
    \subfloat[]{\label{fig:Example1_trueDCT_featureHadamard_betaAligned_truebeta_featuretransdom}\includegraphics[width=0.32\textwidth]{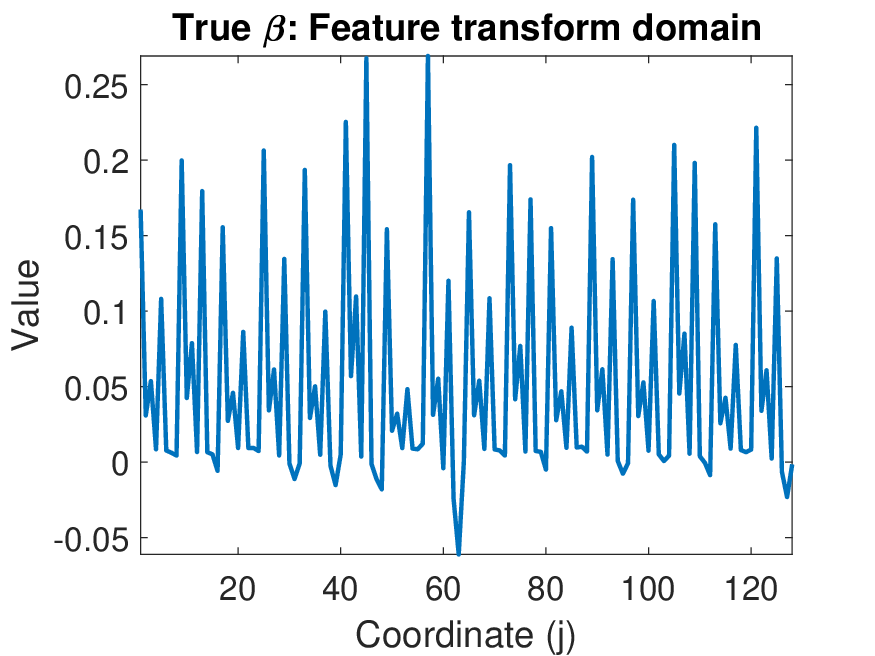}}
    \subfloat[]{\label{fig:Example1_trueDCT_featureHadamard_FeatureCovarianceMatrix}\includegraphics[width=0.32\textwidth]{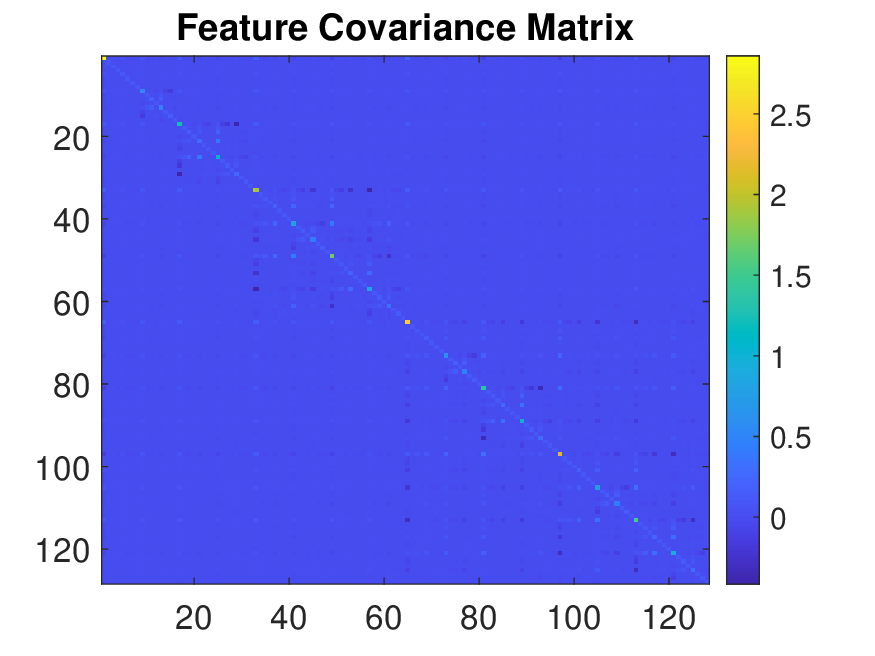}}
    \caption{Empirical results for LS regression based on a class $\mathcal{F}_{p}^{\sf lin}(\{\vec{u}_j\}_{j=1}^{d})$ of linear functions in the form described in Example \ref{example:function class is linear mappings with p parameters}. 
    The data model is ${y=\vec{x}^{T} \vecgreek{\beta}+\epsilon}$. (a) The components of $\vecgreek{\beta}$. (b) The components of the DCT transform of $\vecgreek{\beta}$, i.e.,     $\mtx{C}_d^T \vecgreek{\beta}$. This shows that \textbf{the majority of $\vecgreek{\beta}$'s energy is at the low-range DCT ``feature band" of $j=1,\dots,20$}. (c)~The input data covariance matrix $\mtx{\Sigma}_{\vec{x}}$.
    As explained in detail in the main text, the second row corresponds to $\mathcal{F}_{p}^{\sf lin}(\{\vec{u}_j\}_{j=1}^{d})$ with DCT features, and the third row corresponds to $\mathcal{F}_{p}^{\sf lin}(\{\vec{u}_j\}_{j=1}^{d})$ with Hadamard features. 
    }
    \label{fig:Example1_aligned_beta}
\end{figure*}

Figure~\ref{fig:Example1_aligned_beta} studies the relationship between model complexity (expressed by the number of parameters) and test error induced by linear regression with a linear feature map as in Example~\ref{example:function class is linear mappings with p parameters}.
We set the input dimension $d = 128$, number of training examples $n = 32$, and vary the number of parameters $p$ between $1$ and $d$.
We also consider $\vecgreek{\beta}$ to be an approximately $20$-sparse parameter vector such that the majority of its energy\footnote{The \textit{energy} of a vector $\vecgreek{\beta}\in\mathbb{R}^d$ is defined as its squared $\ell_2$-norm, namely, ${\sf Energy}\{\vecgreek{\beta}\}\triangleq\Ltwonorm{\vecgreek{\beta}}=\sum_{j=1}^{d}\beta_j^2$ where $\beta_j\in\mathbb{R}$ is the $j^{\sf th}$ component of $\vecgreek{\beta}$. The series of the $\beta_j^2$ values for $j=1,\dots,d$ reflects how well the representation basis organizes the information of $\vecgreek{\beta}$. 
}
is contained in the first $20$ DCT features (see Figure~\ref{fig:Example1_trueDCT_featureDCT_betaAligned_truebeta_truetransdom} for an illustration).
The input data covariance matrix $\mtx{\Sigma}_{\vec{x}}$ is also considered to be \emph{anisotropic}, as pictured in the heatmap of its values in Figure~\ref{fig:Example1_trueDCT_featureDCT_betaAligned_InputCovarianceMatrix}.

We consider two choices of linear feature maps, i.e., two choices of orthonormal bases $\{\vec{u}_j\}_{j=1}^d$: the discrete cosine transform (DCT) basis and the Hadamard basis.
For both choices of orthonormal feature basis, the model is \emph{well-specified} for $p = d$, since the complete set of $d$ orthonormal vectors spans the input space $\mathbb{R}^d$.
In the first case of DCT features, the information of the true parameter vector $\vecgreek{\beta}$ is concentrated in the first $20$ DCT-feature-coordinates  (see Figure~\ref{fig:Example1_trueDCT_featureDCT_betaAligned_truebeta_featuretransdom}); hence, the misspecification becomes relatively small if the first 20 DCT-feature coordinates are included in the learning, even though the misspecification vanishes exactly only at $p=d$.
As a consequence of the alignment between the DCT feature map and the above-defined DCT-based form of the input covariance matrix ${\mtx{\Sigma}_{\vec{x}}=\mtx{C}_d \mtx{\Lambda}_{\vec{x}} \mtx{C}_d^T}$, the feature covariance matrix $\mtx{\Sigma}_{\vecgreek{\phi}}$ is diagonal and its entries constitute the first $p$ entries of the diagonalized form of the input data covariance matrix $\mtx{\Lambda}_{\vec{x}}$.
We therefore write as shorthand $\mtx{\Sigma}_{\vecgreek{\phi}} = [\mtx{\Lambda}_{\vec{x}}]_{1:p}$; here, $[\mtx{\Lambda}_{\vec{x}}]_{1:p}$ denotes the leading $p\times p$ principal submatrix of the $d\times d$ matrix $\mtx{\Lambda}_{\vec{x}}$.

The ensuing \emph{spiked covariance structure} is depicted in Figure~\ref{fig:Example1_trueDCT_featureDCT_betaAligned_truebeta_FeatureCovarianceMatrix}.
Figure~\ref{fig:Example1_trueDCT_featureDCT_betaAligned_curves} plots the test error as a function of the number of parameters, $p$, in the model and shows a counterintuitive relationship.
When $p < n = 32$, we obtain non-zero training error and observe the classic bias-variance tradeoff.
When $p > n$, the model is able to perfectly interpolate the training data (with the minimum $\ell_2$-norm solution).
Although interpolation of noise is traditionally associated with overfitting, here we observe that the test error monotonically decays with the number of parameters $p$.
This is a manifestation of the \emph{double descent} phenomenon in an elementary example inspired by  signal processing perspectives.

The second choice of the Hadamard basis also gives rise to a double descent behavior, but is quite different both qualitatively and quantitatively.
The feature space utilized for learning (the Hadamard basis) is fundamentally mismatched to the feature space in which the original parameter vector $\vecgreek{\beta}$ is sparse (the DCT basis).
Consequently, as seen in Figure~\ref{fig:Example1_trueDCT_featureHadamard_betaAligned_truebeta_featuretransdom}, the energy of the true parameter vector $\vecgreek{\beta}$ is spread over the $d$ Hadamard features in an unorganized manner.
Moreover, the covariance matrix  $\mtx{\Sigma}_{\vecgreek{\phi}}$ of Hadamard features, depicted as a heatmap in Figure~\ref{fig:Example1_trueDCT_featureHadamard_FeatureCovarianceMatrix}, is significantly more correlated and dispersed in its structure.
Figure~\ref{fig:Example1_trueDCT_featureHadamard_betaAligned_curves} displays a stronger benefit of overparameterization in this model owing to the increased model misspecification arising from the choice of Hadamard feature map. Specifically, note that this misspecification slowly reduces the bias along the entire range of $p$ from $1$ to $d$.
Accordingly, unlike in the better-aligned case of DCT features, the overparameterized regime for Hadamard features significantly dominates the underparameterized regime in terms of test performance.

\begin{figure*}
    \centering
    \subfloat[]{\label{fig:Example1_trueDCT_featureDCT_betaNonaligned_truebeta_signaldom}\includegraphics[width=0.32\textwidth]{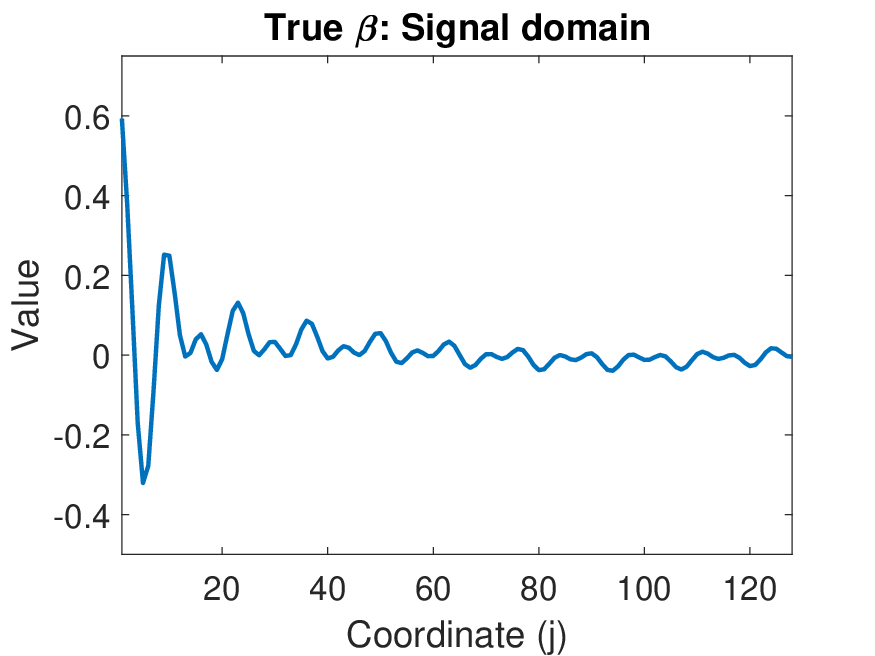}}
    \subfloat[]{\label{fig:Example1_trueDCT_featureDCT_betaNonaligned_truebeta_truetransdom}\includegraphics[width=0.32\textwidth]{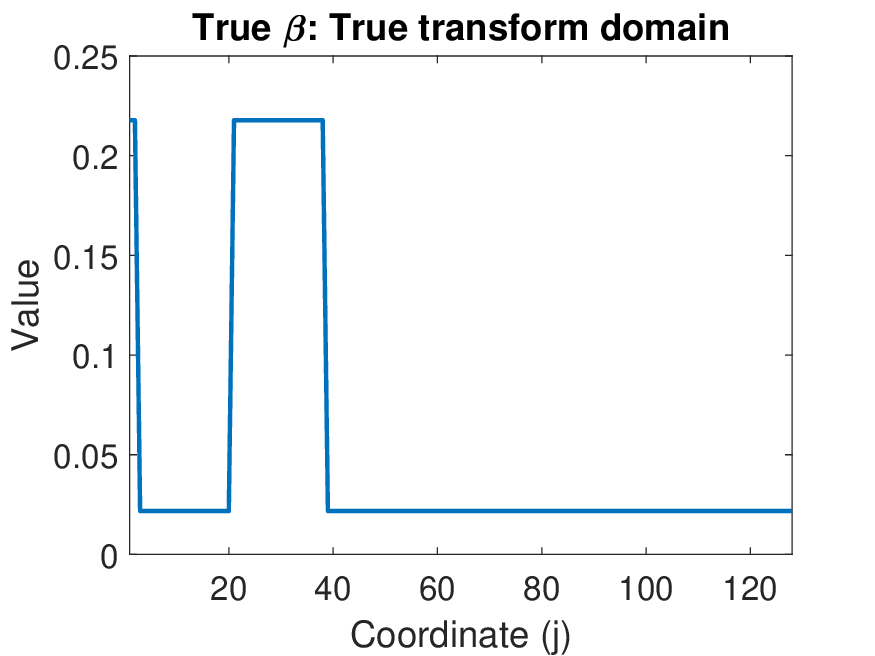}}
    \subfloat[]{\label{fig:Example1_trueDCT_featureDCT_betaNonaligned_InputCovarianceMatrix}\includegraphics[width=0.32\textwidth]{figures/trueDCT_featureDCT_InputCovarianceMatrix.eps}}
    \\
    \subfloat[]{\label{fig:Example1_trueDCT_featureDCT_betaNonaligned_curves}\includegraphics[width=0.32\textwidth]{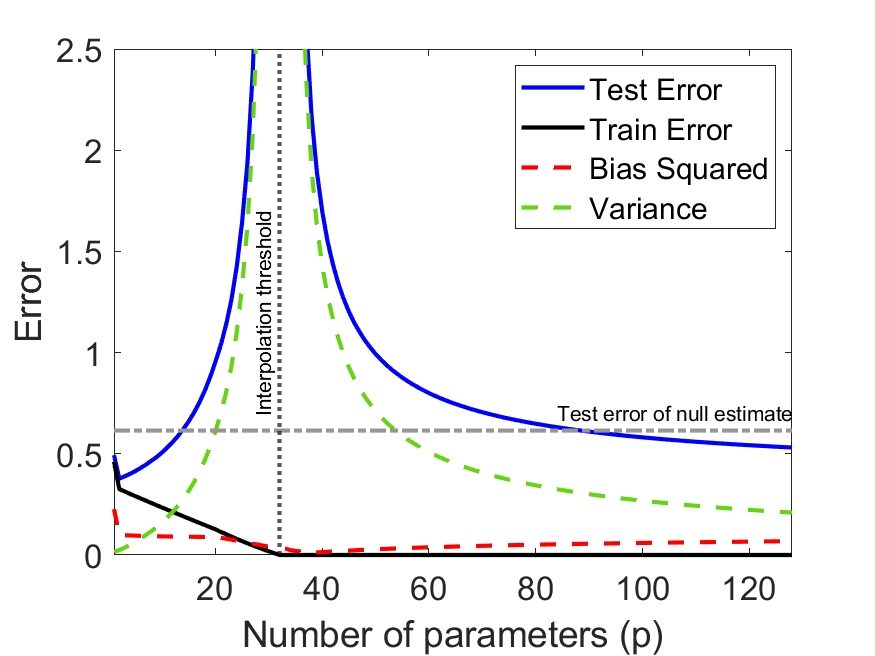}}
    \subfloat[]{\label{fig:Example1_trueDCT_featureDCT_betaNonaligned_truebeta_featuretransdom}\includegraphics[width=0.32\textwidth]{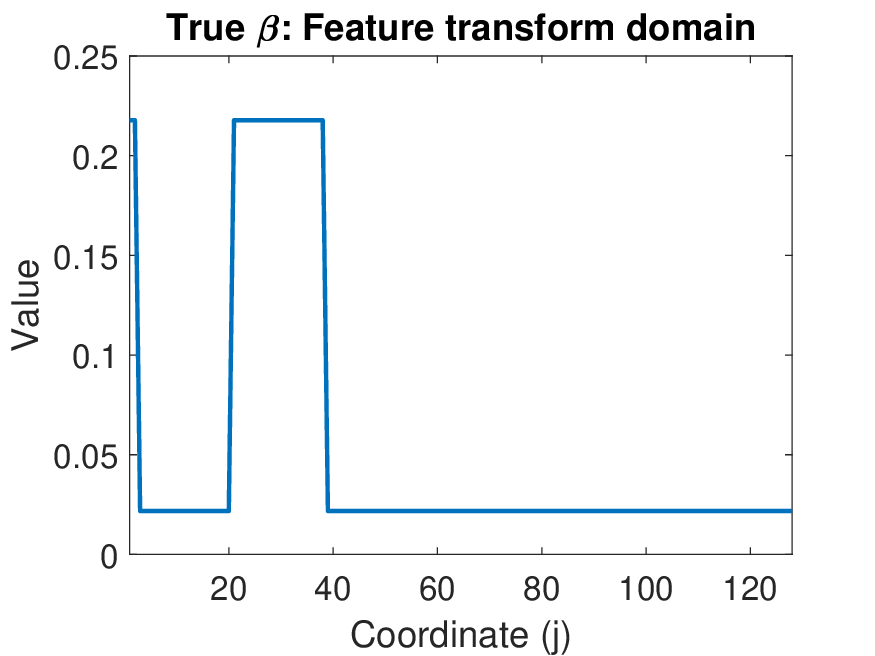}}
    \subfloat[]{\label{fig:Example1_trueDCT_featureDCT_betaNonaligned_truebeta_FeatureCovarianceMatrix}\includegraphics[width=0.32\textwidth]{figures/trueDCT_featureDCT_FeatureCovarianceMatrix.eps}}
    \caption{Empirical results for LS regression based on a class $\mathcal{F}_{p}^{\sf lin}(\{\vec{u}_j\}_{j=1}^{d})$ of linear functions in the form described in Example \ref{example:function class is linear mappings with p parameters}. The data model is ${y=\vec{x}^{T} \vecgreek{\beta}+\epsilon}$. (a) The components of $\vecgreek{\beta}$. (b) The components of the DCT transform of $\vecgreek{\beta}$, i.e.,     $\mtx{C}_d^T \vecgreek{\beta}$. This shows that besides two high-energy components at the first two features, \textbf{the majority of $\vecgreek{\beta}$'s energy is at the mid-range DCT ``feature band" of $j=21,\dots,38$}. (c)~The input data covariance matrix $\mtx{\Sigma}_{\vec{x}}$.
    As explained in detail in the main text, the second row of subfigures corresponds to $\mathcal{F}_{p}^{\sf lin}(\{\vec{u}_j\}_{j=1}^{d})$ with DCT features.
    (d) The training and test errors and the bias--variance components as functions of $p$. (e) The components of $\vecgreek{\beta}$ in the DCT feature space. (f) The DCT-feature covariance matrix, whose largest diagonal entries occur at $j=1,\ldots,20$. Thus, apart from $j=1,2$, the dominant signal coordinates $j=21,\ldots,38$ are outside the high-variance feature band.}
    \label{fig:Example1_nonaligned_beta}
\end{figure*}

Finally, the results in Figure~\ref{fig:Example1_nonaligned_beta} correspond to $\vecgreek{\beta}$ whose components are shown in Figure \ref{fig:Example1_trueDCT_featureDCT_betaNonaligned_truebeta_signaldom} and its DCT feature space representation is shown in Figure \ref{fig:Example1_trueDCT_featureDCT_betaNonaligned_truebeta_featuretransdom}. The majority of  $\vecgreek{\beta}$'s energy is in a mid-range ``feature band" that includes the 18 DCT features at coordinates $j=21,\dots,38$; in addition, there are two more high-energy features in coordinates $j=1,2$, and the remaining coordinates are of low energy (see Figure~\ref{fig:Example1_trueDCT_featureDCT_betaNonaligned_truebeta_featuretransdom}). 
In contrast, Figure \ref{fig:Example1_trueDCT_featureDCT_betaNonaligned_truebeta_FeatureCovarianceMatrix} shows that the largest DCT-feature variances occur at $j=1,\ldots,20$ (Figure \ref{fig:Example1_trueDCT_featureDCT_betaNonaligned_truebeta_FeatureCovarianceMatrix} shows the DCT-feature covariance matrix whose main diagonal values are the DCT-feature variances). Thus, except for
$j=1,2$, the directions carrying most of the signal energy (Figure~\ref{fig:Example1_trueDCT_featureDCT_betaNonaligned_truebeta_featuretransdom}) are largely distinct from the high-variance directions of the feature distribution (Figure \ref{fig:Example1_trueDCT_featureDCT_betaNonaligned_truebeta_FeatureCovarianceMatrix}).
This situation leads to the error curves presented in Figure \ref{fig:Example1_trueDCT_featureDCT_betaNonaligned_curves}, where double descent of test errors occurs, but the global minimum of test error is not achieved by an interpolating solution. This will be explained in more detail below.

\subsection{Why is double descent a surprise?\nopunct} 
As a consequence of the classical bias-variance tradeoff, the test error as a function of the number of parameters forms a U-shaped curve in the \emph{underparameterized} regime.
For example, Fig.~\ref{fig:Example1_trueDCT_featureDCT_betaAligned_curves} depicts this U-shaped curve in the underparameterized range of solutions $p<n$, i.e.,  to the left of the interpolation threshold (vertical dotted line at $p=n=32$).
This classical bias-variance tradeoff is usually induced by two fundamental behaviors of non-interpolating solutions:
\begin{itemize}
	\item The \textit{bias term usually decreases} because the optimal mapping $f_{\sf opt}$ can be better approximated by function classes  $\mathcal{F}$ that are more complex. In other words, models with more parameters or less regularization typically result in a lower (squared) bias component in the test error decomposition (\ref{eq:bias variance decomposition - general definiton for squared error}).
	As an example, see the red dashed curve to the left of the interpolation threshold in  Fig.~\ref{fig:Example1_trueDCT_featureDCT_betaAligned_curves}. 
	
	\item The \textit{variance term usually increases} because function classes $\mathcal{F}$ that are more complex have increased sensitivity to (or variation with respect to) the randomness in the training dataset $\mathcal{D}$. 
	Note that such noise can arise due to stochastic measurement errors and/or model misspecification~(see, e.g., \cite{rao1971some}).
	In other words, models with more parameters or less regularization typically result in a higher variance component in the test error decomposition (\ref{eq:bias variance decomposition - general definiton for squared error}).
	As an example, see the green dashed curve to the left of the interpolation threshold in  Fig.~\ref{fig:Example1_trueDCT_featureDCT_betaAligned_curves}.
\end{itemize}

Typically, the \emph{interpolating} solution of minimal complexity (e.g., $\mathcal{F}_{p}^{\sf lin}(\{\vec{u}_j\}_{j=1}^{d})$ for $p=n$) has a high test error. 
This inferior performance around the interpolation threshold (which can be mathematically explained by the poor conditioning of the training data matrix, or the training feature matrix $\mtx{\Phi}$, e.g., in (\ref{eq:normal equations for feature learning})) led to prior neglect of the study of the range of interpolating solutions, i.e., models such as Examples \ref{example:function class is linear mappings with p parameters}-\ref{example:function class is nonlinear mappings} with $p > n$.

In problems related to least squares, the learned solution involves an inverse or pseudoinverse of an empirical covariance (e.g., $\mtx{\Phi}^T\mtx{\Phi}$ in (\ref{eq:normal equations for feature learning})) or Gram matrix. Near the interpolation threshold, the smallest nonzero eigenvalue of this matrix can approach zero, strongly amplifying the variance component of the test error and thereby producing a peak in test error. Interpolation alone is therefore not sufficient to produce such a peak: when the solution avoids an explicit or implicit ill-conditioned inversion, the interpolation peak need not occur. Indeed, for the PCA-based subspace learning in Section \ref{sec:subspace learning}, interpolation can occur without applying an ill-conditioned inverse or pseudoinverse of the sample covariance matrix, and no corresponding interpolation-related peak of the test error appears in the setting considered there.

It is essential to note that we cannot expect double descent behavior (and beneficial overparameterization) to arise from \emph{any} interpolating solution; indeed, it is easy to find solutions that would generalize poorly even as more parameters are added.
Accordingly, we should expect the \emph{inductive bias}\footnote{A complementary, influential line of work (beginning with~\cite{soudry2018implicit,ji2019implicit}) shows a related phenomenon of \emph{implicit regularization} carried out by popular optimization algorithms. One prominent such result is that the max-margin support vector machine (SVM) (which minimizes the $\ell_2$-norm of the solution subject to correctly classifying all of the training examples with a hard margin constraint) arises naturally as a consequence of running gradient descent to minimize training loss. 
This demonstrates the relative ease of finding $\ell_2$-norm minimizing solutions in practice, and further motivates their study.} of the minimum $\ell_2$-norm interpolation to play a critical role in aiding generalization.
However, inductive bias provides intuition rather than a full explanation for the double descent behavior.
For one, it does not explain the seemingly harmless interpolation of noise in training data in the highly overparameterized regime.
For another, minimum $\ell_2$-norm interpolations will not generalize well for \emph{any} data distribution: indeed, classic experiments demonstrate catastrophic signal reconstruction in some cases by the minimum $\ell_2$-norm interpolation~\citep{chen2001atomic}.

An initial examination of the bias-variance decomposition of the test error in the overparameterized regime (Figures~\ref{fig:Example1_trueDCT_featureDCT_betaAligned_curves}, \ref{fig:Example1_trueDCT_featureHadamard_betaAligned_curves} and~\ref{fig:Example1_trueDCT_featureDCT_betaNonaligned_curves}) demonstrates significantly different behavior from the underparameterized regime, at least when minimum $\ell_2$-norm solutions are used: there appears to be a significant decrease in the variance component that dominates the increase in the bias component to cause an overall benefit in overparameterization.
It is also unclear whether the overparameterized regime is strictly beneficial for generalization in the sense that the global minimum of the test error is achieved by an overparameterized (interpolating) solution.
The simple experiments in this section demonstrate a mixed answer to this question: Figs.~\ref{fig:Example1_trueDCT_featureDCT_betaAligned_curves} and \ref{fig:Example1_trueDCT_featureHadamard_betaAligned_curves} demonstrate beneficial double descent behaviors, whereas Fig.~\ref{fig:Example1_trueDCT_featureDCT_betaNonaligned_curves} shows a double descent behavior but the underparameterized regime is the one that minimizes test error.

\subsection{The role of model misspecification in beneficial overparameterization}
\label{subsec:misspecification of learned models}
Misspecification was defined above as a learning setting where the function class $\mathcal{F}$ does not include the (unconstrained) optimal solution $f_{\sf opt}$ that minimizes the test error. 
In this section, we provide an elementary calculation to illustrate that this model misspecification significantly promotes the double descent behavior. 
Consider regression with respect to squared loss, data model ${y=\vec{x}^T\vecgreek{\beta}+\epsilon}$ and a function class ${\mathcal{F}_{p}^{\sf lin}(\{\vec{u}_j\}_{j=1}^{d})}$ as in Example \ref{example:function class is linear mappings with p parameters}. Then, the optimal solution is ${f_{\sf opt}(\vec{x})=\expectation{y | \vec{x}}=\vec{x}^{T}\vecgreek{\beta}}$. By the definition of the function class ${\mathcal{F}_{p}^{\sf lin}(\{\vec{u}_j\}_{j=1}^{d})}$, the learned function is applied on a feature vector $\vecgreek{\phi}= \mtx{U}_p^T \vec{x}$ that includes only $p$ features (out of the $d$ possible features) of the input $\vec{x}$. Assume a Gaussian input and that $\mtx{U}_d$ diagonalizes the input covariance $\mtx{\Sigma}_{\vec{x}}$. Then, the optimal solution in the function class ${\mathcal{F}_{p}^{\sf lin}(\{\vec{u}_j\}_{j=1}^{d})}$ is
\begin{equation}
\label{eq:optimal estimate in misspecified case of Example 1}
f_{{\sf opt},\mathcal{F}_{p}^{\sf lin}}(\vec{x})=\expectation{y | \vecgreek{\phi}}=\vecgreek{\phi}^{T}\mtx{U}_p^T\vecgreek{\beta} = \vec{x}^{T}\mtx{U}_p\mtx{U}_p^T\vecgreek{\beta}
\end{equation}
We can notice the difference between the optimal solution $f_{{\sf opt},\mathcal{F}_{p}^{\sf lin}}$ in ${\mathcal{F}_{p}^{\sf lin}(\{\vec{u}_j\}_{j=1}^{d})}$ and the unconstrained optimal solution $f_{\sf opt}$ as follows.  
Note that $\mtx{U}_p\mtx{U}_p^T$ in (\ref{eq:optimal estimate in misspecified case of Example 1}) is the orthogonal projection matrix onto the $p$-dimensional feature space that is utilized for learning and spanned by the $p$ orthonormal vectors $\{\vec{u}_j\}_{j=1}^{p}$. Hence, having $\vecgreek{\beta}$ or the entire distribution of $\vec{x}$ contained in the $p$-dimensional feature space (i.e., $\mtx{U}_p\mtx{U}_p^T\vecgreek{\beta}=\vecgreek{\beta}$ or $\mtx{U}_p\mtx{U}_p^T\vec{x}=\vec{x}$ for all $\vec{x}$ with strictly positive probability density) is a sufficient condition for (almost surely) avoiding misspecification that excludes the optimal solution $f_{\sf opt}$ from the function class ${\mathcal{F}_{p}^{\sf lin}(\{\vec{u}_j\}_{j=1}^{d})}$.

Throughout this subsection, assume that $\vec x\sim\mathcal N(\vec{0},\mtx{\Sigma}_{\vec{x}})$, $\mtx{\Sigma}_{\vec{x}}=\mtx{U}\mtx{\Lambda}_{\vec{x}} \mtx{U}^T$ 
where $\mtx{U}=[\vec{u}_1,\ldots,\vec{u}_d]$ is an orthogonal matrix and $\mtx{\Lambda}_{\vec{x}}$ is a diagonal matrix. The same matrix $\mtx{U}$ defines the linear feature map.  Then, $\mtx{U}^T\vec{x}\sim\mathcal N(\vec{0},\mtx{\Lambda}_{\vec{x}})$. Thus, diagonalization makes the transformed coordinates uncorrelated, and Gaussianity implies that they are mutually independent. In particular, the $p$-dimensional feature vector
$\vecgreek{\phi}=\mtx{U}_p^T\vec{x}$ is independent of the $d-p$ omitted features. 
Such learning settings with independent features appear in our above discussed example of DCT features and Gaussian input with covariance matrix that is diagonalized by the DCT matrix.

Moreover, we assume that the feature covariance matrix $\mtx{\Sigma}_{\vecgreek{\phi}}=\mtx{U}_p^T \mtx{\Sigma}_{\vec{x}} \mtx{U}_p$ is full rank for any $p$. This makes the Gaussian feature distribution nondegenerate and ensures that ${\rm rank}(\mtx{\Phi}) = \min\{n,p\}$ almost surely, hence the interpolation threshold is at $p=n$ rather than at a smaller effective feature dimension.

These assumptions allow us to further emphasize the effect of misspecification on the bias and variance of the learned solution from ${\mathcal{F}_{p}^{\sf lin}(\{\vec{u}_j\}_{j=1}^{d})}$. 
The important observation here is that the data model ${y=\vec{x}^{T}\vecgreek{\beta}+\epsilon}$ can be written as
\begin{equation}
\label{eq:misspecification perspective on noisy data model}
y=\vecgreek{\phi}^{T}\mtx{U}_{p}^{T}\vecgreek{\beta}+\xi + \epsilon    
\end{equation}
where $\vecgreek{\phi}=\mtx{U}_{p}^{T}\vec{x}$ is the $p$-dimensional feature vector of the input, and 
\begin{equation}
\label{eq:definition of the misspecification xi variable}
{\xi\triangleq\sum_{j=p+1}^{d}{\left(\vec{u}_j^{T}\vec{x}\right)\left(\vec{u}_j^{T}\vecgreek{\beta}\right)}}    
\end{equation}
 is a random variable which is independent of the feature vector $\vecgreek{\phi}$. 
Note that $\xi$ depends on the $d-p$ unused features $\{\vec{u}_j^{T}\vec{x}\}_{j=p+1}^{d}$. We consider a zero-mean Gaussian input $\vec{x}$, here, with covariance matrix ${\mtx{\Sigma}_{\vec{x}}=\mtx{U}\mtx{\Lambda}_{\vec{x}}\mtx{U}^{T}}$. Hence, the misspecification variable $\xi$ is also zero-mean Gaussian, but with variance 
\begin{equation}
\label{eq:variance of misspecification variable}
\sigma_{\xi}^{2} = \sum_{j=p+1}^{d}{\lambda_j\cdot \left(\vec{u}_j^{T}\vecgreek{\beta}\right)^2 },
\end{equation}
which sums over the $d-p$ feature directions that are not utilized in the $p$-dimensional feature space of ${\mathcal{F}_{p}^{\sf lin}(\{\vec{u}_j\}_{j=1}^{d})}$. Then, the model is (almost surely) well-specified if and only if $\sigma_{\xi}^{2}=0$.

Recall that the bias-variance decomposition formulas (\ref{eq:bias variance decomposition - general definiton for squared error}), (\ref{eq:bias variance decomposition - general definiton for squared error - bias}), (\ref{eq:bias variance decomposition - general definiton for squared error - variance}) require a mathematically defined expectation of the learned predictor $\expectationwrt{\widehat{f}(\vec{x})}{\mathcal{D}}$. Here, when $\sigma_{\xi}^2+\sigma_{\epsilon}^2>0$, the expected learned predictor is undefined for $p=n$, as a result of the non-integrable entries of the pseudoinverse of the Gaussian feature matrix $\mtx{\Phi}^+$ (see Appendix \ref{appendix:subsec:The Expected Learned Predictor is Undefined at the interpolation threshold}). Moreover, when $\sigma_{\xi}^2+\sigma_{\epsilon}^2>0$, the variance is infinite for $p\in\{n-1,n+1\}$ (see Appendix \ref{appendix:subsec:Proof - variance diverges near interpolation threshold}).  Accordingly, we continue the developments in this section by assuming that $p\ne n$ for the bias and its decomposition, and $p\notin \{n-1,n,n+1\}$ for the variance and its decomposition. 

We define the \textit{misspecification bias} (squared) as 
\begin{equation}
\label{eq:misspecification bias definition}
{\sf bias}_{\sf misspec}^{2} \triangleq \expectationwrt{\left(f_{{\sf opt},\mathcal{F}_{p}^{\sf lin}}(\vec{x}) - f_{\sf opt}(\vec{x})\right)^2}{\vec{x}},
\end{equation}
namely, the expected squared difference between the optimal solution in the function class ${\mathcal{F}_{p}^{\sf lin}(\{\vec{u}_j\}_{j=1}^{d})}$ and the optimal unconstrained solution. We also define the \textit{in-class bias} (squared) as 
\begin{equation}
\label{eq:in-class bias definition}
{\sf bias}_{\sf in class}^{2}\left(\widehat{f}\right) \triangleq \expectationwrt{\left(\expectationwrt{\widehat{f}(\vec{x})}{\mathcal{D}} - f_{{\sf opt},\mathcal{F}_{p}^{\sf lin}}(\vec{x})\right)^2}{\vec{x}},
\end{equation}
which is the expected squared difference between the learned solution (in expectation over the training dataset) and the optimal solution in the function class ${\mathcal{F}_{p}^{\sf lin}(\{\vec{u}_j\}_{j=1}^{d})}$.
Now, due to the orthogonality of the feature maps that form ${\mathcal{F}_{p}^{\sf lin}(\{\vec{u}_j\}_{j=1}^{d})}$ and independence of features, we can decompose the squared bias from (\ref{eq:bias variance decomposition - general definiton for squared error - bias}) as 
\begin{equation}
\label{eq:bias decomposition}
{\sf bias}^{2}\left(\widehat{f}\right) = {\sf bias}_{\sf misspec}^{2} + {\sf bias}_{\sf in class}^{2}\left(\widehat{f}\right).
\end{equation}
Specifically, the definition of ${\mathcal{F}_{p}^{\sf lin}(\{\vec{u}_j\}_{j=1}^{d})}$ in Example \ref{example:function class is linear mappings with p parameters} implies that
\begin{equation}
\label{eq:misspecification bias}
{\sf bias}_{\sf misspec}^{2} = \sigma_{\xi}^{2},
\end{equation}
which is the variance (\ref{eq:variance of misspecification variable}) of the misspecification variable $\xi$ defined in (\ref{eq:definition of the misspecification xi variable}) that is monotonically decreasing in $p$.
Moreover, the in-class bias can be expressed as 
\begin{equation}
\label{eq:in-class bias formula}
{\sf bias}_{\sf in class}^{2}\left(\widehat{f}\right) = \vecgreek{\beta}^T \mtx{U}_p  \bar{\mtx{P}}_{\mathcal{N}\left(\mtx{\Phi}\right)} \mtx{\Sigma}_{\vecgreek{\phi}} \bar{\mtx{P}}_{\mathcal{N}\left(\mtx{\Phi}\right)}\mtx{U}_p^T \vecgreek{\beta} 
\end{equation}
where $\bar{\mtx{P}}_{\mathcal{N}\left(\mtx{\Phi}\right)} \triangleq \mtx{I}_p - \expectationwrt{\mtx{\Phi}^{+}\mtx{\Phi}}{\mtx{\Phi}}$ is the expectation of the orthogonal projection matrix onto the nullspace of $\mtx{\Phi}$.
The proof outline of (\ref{eq:bias decomposition})-(\ref{eq:in-class bias formula}) is provided in Appendix \ref{appendix:subsec:Proof - The Bias Decomposition to In-Class and Misspecification Terms}.

Still considering independent features, we can also decompose the variance term (\ref{eq:bias variance decomposition - general definiton for squared error - variance}) to express its portion due to misspecification in ${\mathcal{F}_{p}^{\sf lin}(\{\vec{u}_j\}_{j=1}^{d})}$. 
For this, we first decompose the learned function (by extending the decomposition from (\ref{eq:misspecification perspective on noisy data model}) to the training outputs and using the least squares closed-form solution (\ref{eq:example 1-2 - min-norm solution}); see related formulations in Appendix \ref{appendix:subsec:The Expected Learned Predictor is Undefined at the interpolation threshold} before Eq.~(\ref{appendix:eq:expectation of the learned predictor - intermediate})) as 
\begin{equation}
\label{eq:learned funcation decomposition for devariance decomposition}
\widehat{f}\left(\vec{x}\right) =   \widehat{f}_{\sf misspec}\left(\vec{x}\right) + \widehat{f}_{\sf inclass}\left(\vec{x}\right) 
\end{equation}
where 
\begin{align}
\label{eq:learned funcation decomposition for devariance decomposition - misspecification term}
&\widehat{f}_{\sf misspec}\left(\vec{x}\right)\triangleq \vec{x}^T \mtx{U}_p \mtx{\Phi}^+ \vecgreek{\xi} ,    \\
\label{eq:learned funcation decomposition for devariance decomposition - inclass term}
&\widehat{f}_{\sf inclass}\left(\vec{x}\right)\triangleq \vec{x}^T \mtx{U}_p \mtx{\Phi}^+ \left(\mtx{\Phi}\mtx{U}_p^T\vecgreek{\beta} + \vecgreek{\epsilon} \right).
\end{align}
Here, we define $\mtx{U}_{c(p)}\triangleq[\vec{u}_{p+1},\dots,\vec{u}_d]$, and $\vecgreek{\xi}\triangleq\mtx{X}\mtx{U}_{c(p)}\mtx{U}_{c(p)}^T\vecgreek{\beta}$ is an $n$-dimensional vector whose $i^{\sf th}$ component corresponds to the computation in (\ref{eq:definition of the misspecification xi variable}) but for the $i^{\sf th}$ training input $\vec{x}_i$. Importantly, $\vecgreek{\xi}$, $\mtx{\Phi}$, and $\vecgreek{\epsilon}$ are independent. More details are available in Appendix \ref{appendix:subsec:Proof - The Variance Decomposition to In-Class and Misspecification Terms}.

Then, we can write the variance decomposition as 
\begin{equation}
\label{eq:variance decomposition}
{\sf var}\left(\widehat{f}\right) = {\sf var}_{\sf misspec}\left(\widehat{f}\right) + {\sf var}_{\sf in class}\left(\widehat{f}\right)
\end{equation}
where the variance component due to misspecification is defined as 
\begin{equation}
\label{eq:misspecification variance - definition}
{\sf var}_{\sf misspec}\left(\widehat{f}\right) \triangleq \mathbb{E}_{\vec{x}}\expectationwrt{\Big( \widehat{f}_{\sf misspec}\left(\vec{x}\right) - \expectationwrt{\widehat{f}_{\sf misspec}\left(\vec{x}\right)}{\mathcal{D}} \Big)^2}{\mathcal{D}}
\end{equation}
and the in-class variance term is defined as 
\begin{equation}
\label{eq:in-class variance - definition}
{\sf var}_{\sf inclass}\left(\widehat{f}\right) \triangleq \mathbb{E}_{\vec{x}}\expectationwrt{\Big( \widehat{f}_{\sf inclass}\left(\vec{x}\right) - \expectationwrt{\widehat{f}_{\sf inclass}\left(\vec{x}\right)}{\mathcal{D}} \Big)^2}{\mathcal{D}}.
\end{equation}
Note that we do not decompose the variance for $p\in\{n-1,n,n+1\}$, due to the undefined expected learned predictor at $p=n$ and the infinite variance at $p\in\{n-1,n+1\}$, as explained in Appendices \ref{appendix:subsec:The Expected Learned Predictor is Undefined at the interpolation threshold}, \ref{appendix:subsec:Proof - variance diverges near interpolation threshold}.

Further mathematical developments (presented in Appendix~\ref{appendix:subsec:Proof - The Variance Decomposition to In-Class and Misspecification Terms}) lead to the misspecification variance formulation of  
\begin{equation}
\label{eq:misspecification variance}
{\sf var}_{\sf misspec}\left(\widehat{f}\right) = \sigma_{\xi}^{2} \mtxtrace{\mtx{\Sigma}_{\vecgreek{\phi}}\expectationwrt{\left(\mtx{\Phi}^{T}\mtx{\Phi}\right)^{+}}{\mtx{\Phi}} }
\end{equation}
Recall that, in this setting, the feature covariance matrix is $\mtx{\Sigma}_{\vecgreek{\phi}}=[\mtx{\Lambda}_{\vec{x}}]_{1:p}$. Note that the misspecification aspect in (\ref{eq:misspecification variance}) is induced by $\sigma_{\xi}^{2}$ whose value is determined by the variances of the features unused in the learning (recall (\ref{eq:variance of misspecification variable})). 
The in-class variance can be formulated as 
\begin{align}
\label{eq:in-class variance}
{\sf var}_{\sf inclass}\left(\widehat{f}\right) = & \sigma_{\epsilon}^{2} \mtxtrace{\mtx{\Sigma}_{\vecgreek{\phi}}\expectationwrt{\left(\mtx{\Phi}^{T}\mtx{\Phi}\right)^{+}}{\mtx{\Phi}} } 
\\ \nonumber
& + \vecgreek{\beta}^T \mtx{U}_p {\expectationwrt{ \left(\mtx{\Phi}^{+}\mtx{\Phi} - \expectationwrt{\mtx{\Phi}^{+}\mtx{\Phi}}{\mtx{\Phi}}\right) \mtx{\Sigma}_{\vecgreek{\phi}}    \left(\mtx{\Phi}^{+}\mtx{\Phi} - \expectationwrt{\mtx{\Phi}^{+}\mtx{\Phi}}{\mtx{\Phi}}\right)   }{\mtx{\Phi}}} \mtx{U}_p^T \vecgreek{\beta}.
\end{align}
The proof of the variance decomposition in (\ref{eq:variance decomposition})-(\ref{eq:in-class variance}) is provided in Appendix \ref{appendix:subsec:Proof - The Variance Decomposition to In-Class and Misspecification Terms}.
Hastie et al.~\cite{hastie2019surprises} provide similar formulas and obtain particularly simple expressions in the case where both the input and the true parameter vector are isotropic random vectors.

\begin{figure*}
	\begin{center}
		{\includegraphics[width=0.95\textwidth]{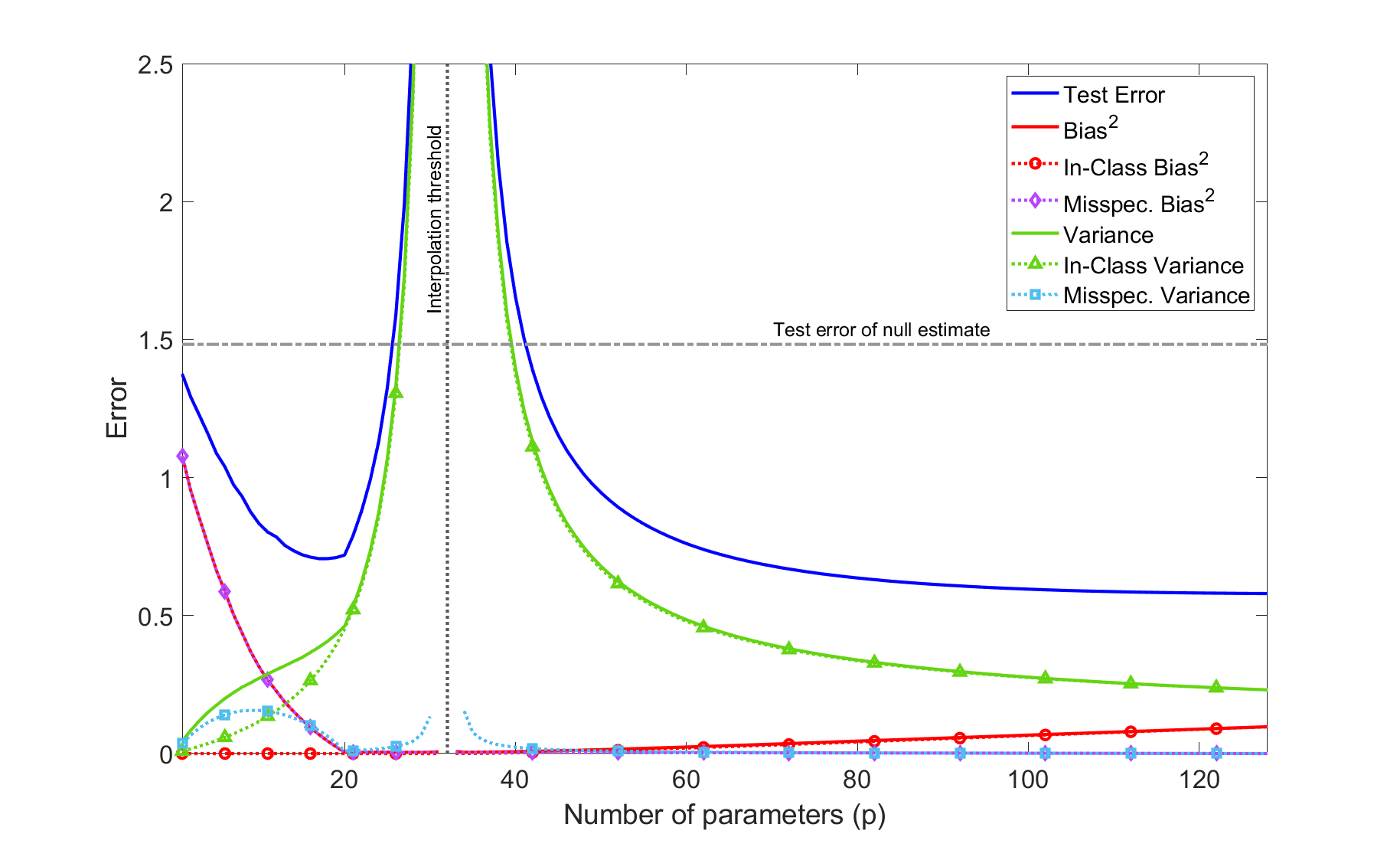}}
		\caption{Bias-variance decomposition of test error, including further decomposition into misspecification and in-class components as formulated in (\ref{eq:bias decomposition})-(\ref{eq:in-class variance}) for cases with independent features. This figure is an extension of Fig.~\ref{fig:Example1_trueDCT_featureDCT_betaAligned_curves}. These results are for LS regression with DCT features and Gaussian input with covariance that is diagonalized by the DCT matrix.}
		\label{fig:misspec_Example1_trueDCT_featureDCT_betaAligned_curves}
	\end{center}
\end{figure*}

Let us return to the example given in Fig.~\ref{fig:Example1_trueDCT_featureDCT_betaAligned_curves} for regression with DCT features that are independent due to the covariance form of the Gaussian input. This setting allows us to  demonstrate the misspecification components of the bias and variance from (\ref{eq:misspecification bias}) and (\ref{eq:misspecification variance}).  
Figure~\ref{fig:misspec_Example1_trueDCT_featureDCT_betaAligned_curves} extends Fig.~\ref{fig:Example1_trueDCT_featureDCT_betaAligned_curves} by illustrating the misspecification and in-class components of the bias and variance versus the number of parameters $p$ in the function class. We can draw the following conclusions from this figure:
\begin{itemize}
    \item The misspecification bias decreases as the function class is more parameterized. In contrast, the in-class bias is zero in the underparameterized regime, and increases with $p$ in the overparameterized regime. Yet, the reduction in the misspecification bias due to overparameterization is more significant than the corresponding increase in the in-class bias; this promotes the occurrence of the global minimum of test error in the overparameterized range.
    \item We can observe that both the in-class and misspecification variance terms peak around the interpolation threshold (i.e., $p=n$) due to the poor conditioning of the feature matrix. Here, the in-class variance peak is much more significant than the misspecification variance peak.
    \item Based on the structure of the true parameter vector (see Fig.~\ref{fig:Example1_trueDCT_featureDCT_betaAligned_truebeta_featuretransdom}), the only significant misspecification occurs for $p<20$ and this is apparent for the misspecification components of both the bias and variance.
\end{itemize}

Recall that, for a full-rank $\mtx{\Sigma}_{\vecgreek{\phi}}$ and $\sigma_{\xi}^{2}+\sigma_{\epsilon}^{2}>0$, the bias and variance are undefined at $p=n$ due to the undefined expectation of the learned predictor (Appendix \ref{appendix:subsec:The Expected Learned Predictor is Undefined at the interpolation threshold}), and the variance is infinite for $p\in\{n-1,n+1\}$ (Appendix \ref{appendix:subsec:Proof - variance diverges near interpolation threshold}). Hence, the graph does not show the bias and variance values (and their decompositions) for $p=n$ and also not the variance decomposition for $p\in\{n-1,n+1\}$.

More generally than the case of independent features, the examples in Figures \ref{fig:Example1_aligned_beta}-\ref{fig:Example1_nonaligned_beta} include misspecification for any $p<d$, and reveal the following insights:
\begin{itemize}
    \item This misspecification, for both DCT and Hadamard features, is due to the representation of the true parameter vector $\vecgreek{\beta}$ requiring all the $d$ features (see Figures \ref{fig:Example1_trueDCT_featureDCT_betaAligned_truebeta_featuretransdom}, \ref{fig:Example1_trueDCT_featureHadamard_betaAligned_truebeta_featuretransdom}, \ref{fig:Example1_trueDCT_featureDCT_betaNonaligned_truebeta_featuretransdom}). 
    \item Learning using DCT features benefits from the fact that $\vecgreek{\beta}$ has energy compaction\footnote{We say that a transform provides \textit{energy compaction} for a signal when a large fraction of the signal's squared $\ell_2$-norm is concentrated in a small number of transform coefficients.} in the DCT domain, which in turn significantly reduces the misspecification bias for $p$ that is large enough to include the majority of $\vecgreek{\beta}$'s energy.
    \item In contrast, in the case of learning using the Hadamard features, the representation of $\vecgreek{\beta}$ in the feature space has its energy spread in an unstructured pattern all over the features, without energy compaction, and therefore the effect of misspecification is further amplified in this case. 
    The amplified misspecification due to a poor selection of feature space actually promotes increased benefits from overparameterization (compared to underparameterized solutions, which are extremely misspecified) as can be observed in Fig.~\ref{fig:Example1_trueDCT_featureHadamard_betaAligned_curves}.
\end{itemize}

This section has examined how the test error and its bias and variance components evolve as the number of parameters crosses the interpolation threshold. In particular, the examples show how poor conditioning creates the interpolation peak and how decreasing model misspecification can make the second descent beneficial. The next section adopts a complementary perspective: it focuses attention on an interpolating linear model with a specific number of parameters $p>n$, and asks under what conditions such an interpolator can simultaneously preserve the underlying signal and fit the training noise without a substantial test-performance degradation.

\section{Overparameterized linear regression}
\label{sec:regression}

Precise theoretical explanations now exist for the phenomena of harmless interpolation of noise and double descent~\citep{belkin2020two,bartlett2020benign,hastie2019surprises,kobak2020optimal,muthukumar2020harmless,mitra2019understanding}.
In this section, we provide a brief recap of these results, all focused on minimum $\ell_2$-norm interpolators.

Additionally, we provide intuitive explanations for the harmless interpolation phenomenon in simplified toy settings that are inspired by statistical signal processing.
We will see that the test error can be upper bounded by a sum of two terms:
\begin{itemize}
    \item A \textit{signal-dependent} term, which expresses how much error is incurred by a solution that interpolates the noiseless part of our (noisy) training data. 
    \item A \textit{noise-dependent} term, which expresses how much error is incurred by a solution that interpolates only the noise in our training data. 
\end{itemize}
Situations that adequately trade off these terms give rise to an overall benign overfitting effect, and can be described succinctly for several families of linear models.
We will now describe these results in more detail.

\subsection{Summary of results}\label{sec:summary}

The minimum $\ell_2$-norm interpolator (MNI), denoted by the $p$-dimensional vector $\alphamintwo$ where $p > n$, has a closed-form expression (recall Eq.~(\ref{eq:example 1-2 - min-norm solution})).
In turn, this allows us to derive closed-form \textit{exact} expressions for the generalization error as a function of the training dataset; indeed, all aforementioned theoretical analyses of the minimum $\ell_2$-norm interpolator take this route.

We can state the diversity of results \cite{belkin2020two,bartlett2020benign,hastie2019surprises,kobak2020optimal,muthukumar2020harmless,mitra2019understanding} in a common framework by decomposing the test error of the MNI, denoted by $\alphamintwo$, into the respective test errors that would be induced by two related solutions. 
To see this, consider the noisy data model (\ref{eq:noisy input-output mapping}) with scalar output; the pure noise components of the $n$ training examples are organized as an $n$-dimensional vector $\vecgreek{\epsilon}$ (which will be used here for theoretical analysis while still being unavailable in the actual learning), and note that the MNI from (\ref{eq:example 1-2 - min-norm solution}) satisfies $\alphamintwo = \alphamintwo_{\sf{sig}} + \alphamintwo_{\sf{noise}}$ where  
\begin{itemize}
    \item $\alphamintwo_{\sf{sig}}\triangleq \mtx{\Phi}^+ \left({\vec{y} - \vecgreek{\epsilon}}\right)$ is the minimum $\ell_2$-norm interpolator on the \textit{noiseless training dataset}  ${\mathcal{D}_{\sf{sig}}=\{(\vec{x}_{i},f_{\sf true}(\vec{x}_{i}))\}_{i=1}^{n}}$. Note that \[{[f_{\sf true}(\vec{x}_{1}),\dots,f_{\sf true}(\vec{x}_{n})]^T =\vec{y} - \vecgreek{\epsilon} \in \mathbb{R}^n},\] where the vectors $\vec{y},\vecgreek{\epsilon}\in \mathbb{R}^n$ include the training data outputs and their error components, respectively.
    Further, denote the \emph{signal specific function fit} as $\widehat{f}_{\sf{sig}}(\vec{x}) \triangleq \vecgreek{\phi}^T \alphamintwo_{\sf{sig}}$ where $\vecgreek{\phi}$ is the feature vector of the input $\vec{x}$.
    \item $\alphamintwo_{\sf{noise}}\triangleq \mtx{\Phi}^+ \vecgreek{\epsilon}$ is the minimum $\ell_2$-norm interpolator on the \textit{pure noise} ${\vecgreek{\epsilon}\in \mathbb{R}^n}$ in the training data. In other words, $\alphamintwo_{\sf{noise}}$ is the minimum $\ell_2$-norm interpolator of the dataset ${\mathcal{D}_{\sf{noise}}=\{(\vec{x}_{i},\epsilon_{i})\}_{i=1}^{n}}$. Further, denote the \emph{noise specific function fit} as $\widehat{f}_{\sf{noise}}(\vec{x}) \triangleq \vecgreek{\phi}^T \alphamintwo_{\sf{noise}}$.
\end{itemize}
Recall that we defined $\testerr(\widehat{f}) \triangleq \EE_{\vec{x},y}\left[(y - \widehat{f}(\vec{x}))^2\right]$ as the test error for the LS regression problem.
Then, by the decomposition $\widehat{f}(\vec{x}) = \widehat{f}_{\sf{sig}}(\vec{x}) + \widehat{f}_{\sf{noise}}(\vec{x})$, the zero-mean error component and its independence of the input in the data model, we use the algebraic inequality $(a+b)^2 \leq 2a^2 + 2b^2$ to get the following upper bound on the test error: 
\begin{align}\label{eq:signalnoisedecomposition}
    \testerr \leq 2 \testerrsignal + 2 \testerrnoise + \testerrirred
\end{align}
where
\begin{itemize}
    \item $\testerrsignal$ is a \textit{signal specific} component that is determined by the noiseless data interpolator $\alphamintwo_{\sf{sig}}$, defined as $\testerrsignal \triangleq \EE_{\vec{x}} \left[(f_{\sf true}(\vec{x}) - \widehat{f}_{\sf sig}(\vec{x}))^2\right]$.
    We expect $\testerrsignal$ to be small (or to asymptotically vanish) if the fit on noiseless data nearly matches the true function, i.e., \emph{signal}. We colloquially refer to conditions under which $\testerrsignal$ is sufficiently small (or vanishing) as conditions for \emph{signal recovery}.
    \item $\testerrnoise$ is a \textit{noise specific} component that is determined by the pure noise interpolator $\alphamintwo_{\sf{noise}}$, defined as $\testerrnoise \triangleq \EE_{\vec{x}} \left[(\widehat{f}_{\sf noise}(\vec{x}))^2\right]$. We hope for the impact of noise (from the training data) on the function fit to be minimal, and therefore for $\testerrnoise$ to be sufficiently small.
    Accordingly, we colloquially refer to conditions under which $\testerrnoise$ is sufficiently small (or asymptotically vanishing) to correspond to \emph{harmless interpolation}.
    \item $\testerrirred$ is the irreducible error, which is defined in (\ref{eq:bias variance decomposition - general definiton for squared error - irreducible error}). For the data model (\ref{eq:noisy input-output mapping}), $\testerrirred = \sigma_{\epsilon}^2$.
\end{itemize}
Moreover, this decomposition is ``sharp" in the sense that both sources of error have to be controlled for good generalization; to see this, note that we would have $\testerr = \testerrsignal$ in the situation of noiseless data, while $\testerr = \testerrnoise + \testerrirred$ in the situation of pure-noise data (i.e., $f_{\sf true}\equiv0$ and hence $\vec{y} = \vecgreek{\epsilon}$).

The foundational results on minimum $\ell_2$-norm interpolation can be viewed from the perspective of (\ref{eq:signalnoisedecomposition}) such that they provide precise characterizations for the error terms $\testerrsignal$ and $\testerrnoise$ (or the resulting total test error) under the well-specified linear model (we will formulate such well-specification of the feature space in our proof in Appendix \ref{appendix:subsec:proof-signal-error-term} for (\ref{eq:signal error term bound - isotropic covariance - well specified - in expectation})).
Moreover, as we will see, these analytical results demonstrate that the goals of minimizing $\testerrsignal$ and $\testerrnoise$ can sometimes be at odds with one another. Accordingly, the number of examples $n$, the number of parameters $p > n$, and properties of the feature covariance matrix $\mtx{\Sigma}_{\vecgreek{\phi}}$ should be chosen to tradeoff these error terms.

In some studies, the ensuing error bounds are \emph{non-asymptotic} and can be stated as a closed-form function of these quantities. 
In other studies, the parameters $p$ and $n$ are grown together at a fixed rate and the \emph{asymptotic} test error is exactly characterized as a solution to either linear or nonlinear equations that is sometimes closed form, but at the very least typically numerically evaluable.
Such exact characterizations of the test error are typically called \emph{precise asymptotics}.

To simplify, we summarize foundational non-asymptotic results from the literature, i.e., error bounds that hold for finite values of $n$, $p$ and $d$ (in addition to their implied asymptotic scalings).
Accordingly, we characterize the behavior of the test error (and various factors influencing it) as a function of the parameters $n, p, d$ using two types of notation. 
First, considering an arbitrary function $h(\cdot,\cdot,\cdot)$, we will use ${\testerr = \mathcal{O}(h(n,p,d))}$ to mean that there exists a universal constant $C > 0$ such that $\testerr \leq C h(n,p,d)$ for all sufficiently large $n, p, d$. 
Second, we will use $\testerr = \Theta(h(n,p,d))$ to mean that there exist universal constants ${c, C > 0}$ such that ${c h(n,p,d) \leq \testerr \leq C h(n,p,d)}$ for all sufficiently large $n, p, d$.
Moreover, the bounds that we state typically hold \emph{with high probability} over the training data, in the sense that the probability that the bound holds goes to $1$ as $n \to \infty$ (and $p,d \to \infty$ with the desired proportions with respect to $n$ in order to maintain overparameterization). 

\subsubsection{When does the minimum $\ell_2$-norm solution enable harmless interpolation of noise?\nopunct}
\label{sec:consistency results}

We begin by focusing on the contribution to the upper bound on the test error that arises from a fit of pure noise, i.e., $\testerrnoise$, and identifying necessary and sufficient conditions under which this error is sufficiently small.
We will see that these conditions, in all cases, reduce to a form of high ``effective overparameterization" in the data relative to the number of training examples (in a sense that we will define shortly). 
In particular, these results can be described under the following two particular models for the feature covariance matrix $\mtx{\Sigma}_{\vecgreek{\phi}}$: 
\begin{itemize}
    \item The simplest result to obtain is in the case of \textit{isotropic covariance}, for which $\mtx{\Sigma}_{\vecgreek{\phi}} = \mathbf{I}_p$.
    For instance, this case occurs when the feature maps in Example \ref{example:function class is linear mappings with p parameters} are applied on input data $\vec{x}$ with isotropic covariance matrix $\mtx{\Sigma}_{\vec{x}} = \mathbf{I}_d$.
    If the features are also \textit{independent} (e.g., consider Example \ref{example:function class is linear mappings with p parameters} with isotropic Gaussian input data), the results in \cite{bartlett2020benign,hastie2019surprises,muthukumar2020harmless,belkin2020two}
    imply that  the noise error term scales, with high probability over the training data and for $p$ sufficiently larger than $n$, as
        \begin{equation}
            \label{eq:noise error term bound - isotropic covariance}
        \testerrnoise = \Theta\left(\frac{\sigma_{\epsilon}^{2} n}{p}\right)
        \end{equation}
        where $\sigma_{\epsilon}^{2}$ denotes the noise variance.
        In Appendix~\ref{appendix:subsec:proof-noise-error-term} we present an elementary derivation of the related statement that
        \begin{equation}\label{eq:noise error term bound expectation over training data - isotropic Gaussian}
            \EE_{\mathcal{D}}[\testerrnoise] = \frac{\sigma_{\epsilon}^2 n}{p - n - 1}
        \end{equation}
        provided that $p > n + 1$ and the feature vector $\vec{\phi}$ is isotropic Gaussian.
        The results in (\ref{eq:noise error term bound - isotropic covariance}) and~\eqref{eq:noise error term bound expectation over training data - isotropic Gaussian} display an explicit benefit of overparameterization in fitting noise in a harmless manner. In more detail, this result is (i) a special case of the more general (non-asymptotic) results for arbitrary covariance by Bartlett et al.~\cite{bartlett2020benign}, (ii) one of the closed-form precise asymptotic results presented as $n,p \to \infty$ by Hastie et al.~\cite{hastie2019surprises}, and (iii) a consequence of the more general characterization of the minimum test error interpolating solution, or ``ideal interpolator", derived by Muthukumar et al.~\cite{muthukumar2020harmless}.
        
         On the other hand, the signal error term can be bounded as 
         \begin{equation}
             \label{eq:signal error term bound - isotropic covariance}
             \testerrsignal = \Theta\left(||\vecgreek{\beta}||_2^2 \left(1 - \frac{n}{p}\right)\right),
         \end{equation}
         displaying an explicit \textit{harm} in overparameterization from the point of view of signal recovery. In Appendix~\ref{appendix:subsec:proof-signal-error-term}, we prove the related result 
         \begin{equation}
            \label{eq:signal error term bound - isotropic covariance - well specified - in expectation}
             \EE_{\mathcal{D}}[\testerrsignal] = \|\vecgreek{\beta}\|_2^2 \left(1 - \frac{n}{p}\right)
         \end{equation}
         provided that $p \ge n$ and the feature vector $\vec{\phi}$ is isotropic Gaussian  in a well specified feature space. 
         In more detail, this result is (i) a non-asymptotic upper bound provided by Bartlett et al.~\cite{bartlett2020benign} (Tsigler and Bartlett~\cite{tsigler2020benign} show that this term is matched by a sharp lower bound), (ii) precise asymptotics provided by Hastie et al.~\cite{hastie2019surprises}.
    
    As discussed later in Section \ref{subsec:signal recovery}, the behavior of $\testerrsignal$ has negative implications for statistical \emph{consistency}\footnote{Statistical consistency implies that $\testerr \to \testerrirred = \sigma_{\epsilon}^2$ as $n \to \infty$.} in a certain high-dimensional sense.

    \item More generally, the case of \textit{anisotropic covariance} and features that are independent in their principal component basis is significantly more complicated. Nevertheless, it turns out that a high amount of ``effective overparameterization" can be defined, formalized, and shown to be sufficient and necessary for interpolating noise in a harmless manner.
    This formalization was pioneered\footnote{Asymptotic (not closed-form) results are also provided by the concurrent work by Hastie et al.~\cite{hastie2019surprises}. These asymptotics match the non-asymptotic scalings provided by Bartlett et al.~\cite{bartlett2020benign} for the case of the spiked covariance model and Gaussian features in the regime where $p/n = \gamma > 1$. These results can also be used to recover in part the benign overfitting results as shown in \cite{bartlett2021deep}.~Muthukumar et al.~\cite{muthukumar2020harmless} recover these scalings in a toy Fourier feature model through elementary calculations that will be reviewed in Section~\ref{subsec:spexplanation} using cosine features.} by Bartlett et al.~\cite{bartlett2020benign}, who provide a definition of effective overparameterization solely as a function of $n$, $p$ and the spectrum of the covariance matrix $\mtx{\Sigma}_{\vecgreek{\phi}}$, denoted by the eigenvalues ${\lambda_1 \geq \ldots \geq \lambda_p}$.
    We refer the reader to~\cite{bartlett2021deep} for a detailed exposition of these results and their consequences. Here, we present the essence of these results in a popular model used in high-dimensional statistics, the spiked covariance model.
    Under this model (recall that $p>n$), we have a feature covariance matrix $\mtx{\Sigma}_{\vecgreek{\phi}}$ with ${s < p - n}$ high-energy eigenvalues and ${p - s}$ low-energy (but non-zero) eigenvalues.
    In other words, ${\lambda_1 \geq \ldots \geq \lambda_s = \lambda_H}$ and ${\lambda_{s+1} = \ldots = \lambda_p = \lambda_L}$ for some ${\lambda_H \gg \lambda_L}$
    (for example, notice that Figure~\ref{fig:Example1_trueDCT_featureDCT_betaAligned_truebeta_FeatureCovarianceMatrix} is an instance of this model). 
    In this special case 
    (assuming sufficient spectral separation so that the effective-rank cutoff of \cite{bartlett2020benign} occurs after the $s$ high-energy directions), the results of~\cite{bartlett2020benign} can be simplified to show that the test error arising from overfitting noise is given by
    \begin{align}\label{eq:testerrornoiseaniso}
        \testerrnoise = \Theta\left(\sigma_{\epsilon}^{2} \left(\frac{s}{n} + \frac{n}{p - s}\right)\right).
    \end{align}
    Equation~\eqref{eq:testerrornoiseaniso} demonstrates a remarkably simple set of necessary and sufficient conditions for harmless interpolation of noise in this anisotropic setting of the above described spiked covariance model:
    \begin{enumerate}[label=(\alph*)]
        \item The number of high-energy directions, given by $s$, must be significantly smaller than the number of examples $n$; more formally, we want $s = o(n)$ that, by the standard little-o definition, implies $\frac{s}{n} \rightarrow 0$. This represents a need for sufficiently low ``effective dimensionality'' of the high-energy part of the data.
        \item The number of low-energy directions, given by $p - s$, must be significantly larger than the number of examples $n$; more formally, we want $p - s = \omega(n)$ that, by the standard little-omega definition, implies $\frac{n}{p - s}\rightarrow 0$. 
        This represents a requirement for sufficiently high ``effective overparameterization" to be able to absorb the noise (from the training data), and fit it in a relatively harmless manner.
    \end{enumerate}
\end{itemize}
Interestingly, the work by Kobak et al.~\cite{kobak2020optimal} considered a related anisotropic model for the special case of a single spike (i.e., for $s = 1$, which usually implies that the first condition is trivially met) and showed that the error incurred by interpolation is comparable to the error incurred by ridge regression.
These results are related in spirit to the above framework, and so we can view the conditions derived by Bartlett et al.~\cite{bartlett2020benign} as sufficient and necessary for ensuring that \emph{interpolation of (noisy) training data does not lead to sizably worse performance than ridge regression}.
Of course, good generalization of ridge regression itself is not always a given.
In this vein, the central utility of studying this anisotropic case is that it is now possible to also identify situations under which the test error arising from fitting pure signal, i.e., $\testerrsignal$, is sufficiently small.
We elaborate on this point next.

\subsubsection{When does the minimum $\ell_2$-norm solution enable signal recovery?\nopunct}\label{subsec:signal recovery}

The above results paint a compelling picture for an explicit benefit of overparameterization in harmless interpolation of noise (via minimum $\ell_2$-norm interpolation).
However, good generalization requires the term $\testerrsignal$ to be sufficiently small as well.
As we now demonstrate, the picture is significantly more mixed for signal recovery.

As shown in Eq.~(\ref{eq:signal error term bound - isotropic covariance}), for the case of isotropic covariance, the signal error term $\testerrsignal$ can be upper and lower bounded up to universal constants as $||\vecgreek{\beta}||_2^2 \left(1 - \frac{n}{p}\right)$, displaying an explicit \textit{harm} in overparameterization from the point of view of signal recovery.
The ills of overparameterization in signal recovery \textit{with the minimum $\ell_2$-norm interpolator} are classically documented in statistical signal processing; see, e.g., the experiments in the seminal survey  \cite{chen2001atomic}.
A fundamental consequence of this adverse signal recovery is that the minimum $\ell_2$-norm interpolator cannot have statistical \emph{consistency} (i.e., we cannot have $\testerr \to \testerrirred$ as $n \to \infty$) under isotropic covariance for \emph{any} scaling of $p$ that grows with $n$, whether constant (i.e., $p = \gamma n$ for some fixed $\gamma > 1$), or ultra high dimensional (i.e., $p = n^q$ for some $q > 1$).
Indeed, in the ultra high dimensional regime or if $p/n \to \infty$, the test error converges to the ``null risk"\footnote{terminology used in \cite{hastie2019surprises}}, i.e., the test error that would be induced by the ``zero-fit"\footnote{terminology used in \cite{muthukumar2020harmless}} $\alphamintwo_0 \triangleq \vec{0}$.

These negative results clarify that anisotropy of features is \emph{required} for the signal-specific error term $\testerrsignal$ to be sufficiently small (together with a sufficiently small noise-specific error term $\testerrnoise$); in fact, this is the main reason for the requirement that $\lambda_H \gg \lambda_L$ in the spiked covariance model to ensure good generalization in regression tasks.
In addition to this effective low dimensionality in data, we also require a low-dimensional assumption on the signal model.
In particular, the signal energy must be almost entirely aligned with the directions (i.e., eigenvectors) of the top $s \ll n$ eigenvalues of the feature covariance matrix (representing high-energy directions in the data).
Finally, we note as a matter of detail that the original work by Bartlett et al.~\cite{bartlett2020benign} precisely characterizes the contribution from $\testerrnoise$ up to constants independent of $p$ and $n$; however, they only provide upper bounds on the bias term.
In follow-up work, Tsigler et al.~\cite{tsigler2020benign} provide more precise expressions for the bias that show that the bias is non-vanishing in ``tail" signal components outside the first $s$ ``preferred" directions.
Their upper and lower bounds match up to a certain condition number.
The additional studies \cite{mei2019generalization,liao2020random,derezinski2020exact,mei2022generalization} also provide precise asymptotics for the bias and variance terms in certain random feature models, which can be shown to fall under the category of anisotropic covariance models.
See Section~\ref{sec:kernels} for a more detailed exposition on random feature models and their associated kernel methods.

\subsubsection{When does double descent occur?\nopunct}
The above subsections show that the following conditions are sufficient \emph{and} necessary for good generalization of the minimum $\ell_2$-norm interpolator \emph{under a well-specified model}:
\begin{itemize}
    \item low-dimensional signal structure, with the non-zero components of the signal aligned with the $s$ eigenvectors corresponding to the highest-value eigenvalues (i.e., high-energy directions in the data)
    \item low effective dimension in data
    \item an \emph{overparameterized} number of low-value (but non-zero) directions.
\end{itemize}

Given these pieces, the ingredients for double descent become clear. 
In addition to the aforementioned signal and noise-oriented error terms in (\ref{eq:signalnoisedecomposition}), we will in general have a \textit{misspecification} error term, denoted by $\testerrmissp$, which will usually decrease with overparameterization (i.e., nonincreasing monotonicity for an increasing $p$ in the overparameterized regime) and therefore further contributes to the double descent phenomenon. For example, recall Section \ref{subsec:misspecification of learned models} and the example in Fig.~\ref{fig:misspec_Example1_trueDCT_featureDCT_betaAligned_curves} where the misspecification bias and variance terms decrease with $p$ in the overparameterized range. 
Optimization aspects can also affect the double descent behavior, for example, it is shown in \cite{dascoli2020double} that the optimization initialization can contribute to the double descent phenomenon in a lazy training setting  (i.e., where the learned parameters are element-wise close in the infinity-norm sense to their initializations \cite{chizat2019lazy}).
Furthermore, it is shown in \cite{lin2021causes,adlam2020understanding} that the double descent peak can be caused by a variance term that diverges due to the interaction between the training data and the initialization.

Examples of the approximation-theoretic benefits of overparameterization are also provided in \cite{hastie2019surprises}, but these do not go far enough to recover the double descent behavior, as they are still restricted to the isotropic setting in which signal recovery becomes more harmful as the number of parameters increases. 

Belkin et al.~\cite{belkin2020two} provided one of the first explicit characterizations of double descent under two models of randomly selected ``weak features".
In their model, the $p$ features are selected uniformly at random from an ensemble of $d$ features (which are themselves isotropic in distribution and follow either a Gaussian or Fourier model).
This idea of ``weak features" is conceptually related to the classically observed benefits of overparameterization in ensemble models like random forests and boosting, which were discussed in an insightful unified manner in~\cite{wyner2017explaining}. 
This connection was first mentioned in~\cite{belkin2019reconciling}.

In addition to the first theoretical works on this topic, Mei and Montanari~\cite{mei2019generalization} also showed precise asymptotic characterizations of the double descent behavior in a random features model.

\subsection{A signal processing perspective on harmless interpolation}\label{subsec:spexplanation}

In this section, we present an elementary explanation for harmless interpolation using the concepts of aliasing and Parseval's theorem for a ``toy" case of overparameterized linear regression with cosine features on \textit{regularly spaced}, one-dimensional input data.
This elementary explanation was first introduced by Muthukumar et al.~\cite{muthukumar2020harmless}, and inspired subsequent work for the case of classification~\citep{muthukumar2021classification}.
While this calculation illuminates the impact of interpolation in both isotropic and anisotropic settings, we focus here on the isotropic case for simplicity.

This toy setting can be perceived as a special case of Example \ref{example:function class is nonlinear mappings} with one-dimensional input and cosine feature maps. Note that our example here for real-valued cosine feature maps differs in some low-level mathematical details compared to the analysis originally given in \cite{muthukumar2020harmless} for Fourier features.

\begin{example}[Cosine features for one-dimensional input]\label{example:fourieralias}
Throughout this example, we will consider $p$ to be an integer multiple of $n$.
We also consider one-dimensional data $x \in [0,1]$ and a family of cosine feature maps 
\begin{equation}
\label{eq:cosine feature functions - definition}
\varphi_j (x) = \kappa_j \cos((j-1) \pi x),~j=1,\dots,\infty,    
\end{equation}
where the normalization constant $\kappa_j$ equals $1$ for $j=1$ and $\sqrt{2}$ for $j\ge 2$. Then, the $p$-dimensional cosine feature vector is given by
\begin{align}
\label{eq:cosine feature vector definition - definition}
\vecgreek{\phi} = \varphi(x) = \begin{bmatrix}
1 \\ \sqrt{2} \cos(\pi x) \\ \sqrt{2} \cos(\pi (2x)) \\ \vdots \\ \sqrt{2} \cos(\pi (p-1) x)
\end{bmatrix} \in \mathbb{R}^p.
\end{align}
This is an \textit{orthonormal}, i.e., \textit{isotropic} feature family in the sense that 
\begin{equation}
\EE_{x \sim \UNIF[0,1]}\left[\varphi_j (x) \varphi_k (x) \right] = \int_{x\in[0,1]} \varphi_j (x)\varphi_k (x)dx = \delta_{jk},    
\end{equation}
where $\delta_{jk}$ denotes the Kronecker delta.
Furthermore, in this ``toy" model we assume $n$ regularly spaced training data points, i.e., ${x_i = \frac{i-\frac{1}{2}}{n}} \text{ for } i = 1,\ldots,n$. 
\end{example}

While the more standard model in the ML setup would be to consider $\{x_i\}_{i=1}^{n}$ to be i.i.d. draws from the input distribution (which is uniform over $[0,1]$ in this example), the toy model of Example \ref{example:fourieralias} with deterministic, regularly-spaced $\{x_i\}_{i=1}^{n}$ will yield a particularly elementary and insightful analysis owing to the presence of exact \emph{aliases}. Informally, aliases consist of a multitude of different basis functions that perfectly fit a set of samples, and are associated with sampling below the Shannon-Nyquist rate and therefore poor signal recovery in classical signal processing contexts. In the forthcoming treatment of this example we will precisely define the aliases that arise as a result of overparameterization. 

In the regime of overparameterized learning, we have $p > n$. 
The case of regularly-spaced training inputs allows us to see overparameterized learning as reconstructing an \textit{undersampled} signal: the number of examples (given by $n$) is much less than the number of features (given by $p$).
Thus, the undersampling perspective implies that \textit{aliasing}, in some approximate\footnote{The reason that this statement is in an approximate sense in practice is because the training data is typically random, and we can have a wide variety of feature families.} sense, is a core issue that underlies model behavior in overparameterized learning.
We illustrate this through Example~\ref{example:fourieralias}, which is an idealized toy model and a special case of Example \ref{example:function class is nonlinear mappings}.

\begin{figure*}
    \centering
    \subfloat[]{\label{fig:example3_cosine_combination_reconstruction_example}\includegraphics[width=0.42\textwidth]{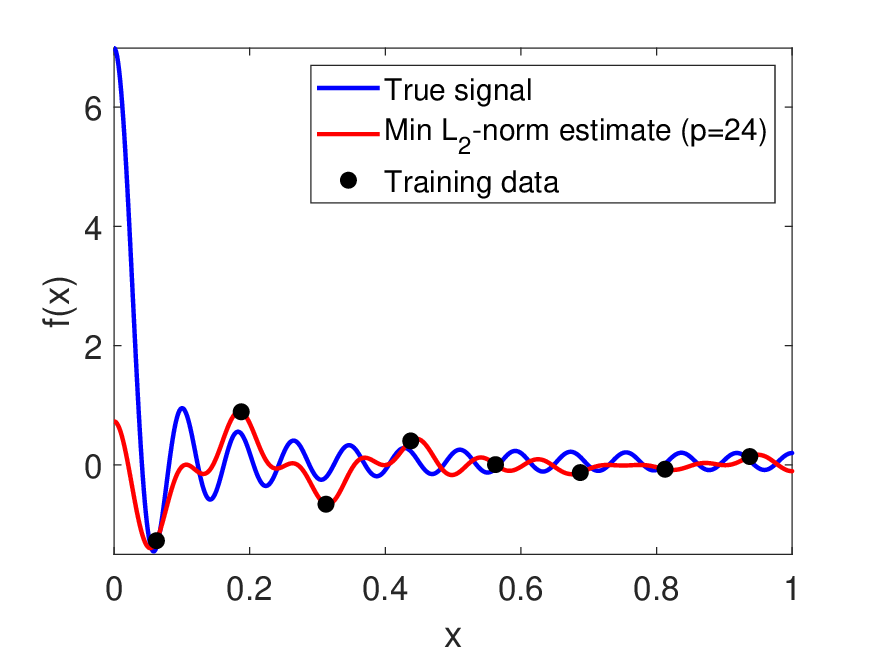}}
    \subfloat[]{\label{fig:example3_cosine_combination_error_curves}\includegraphics[width=0.42\textwidth]{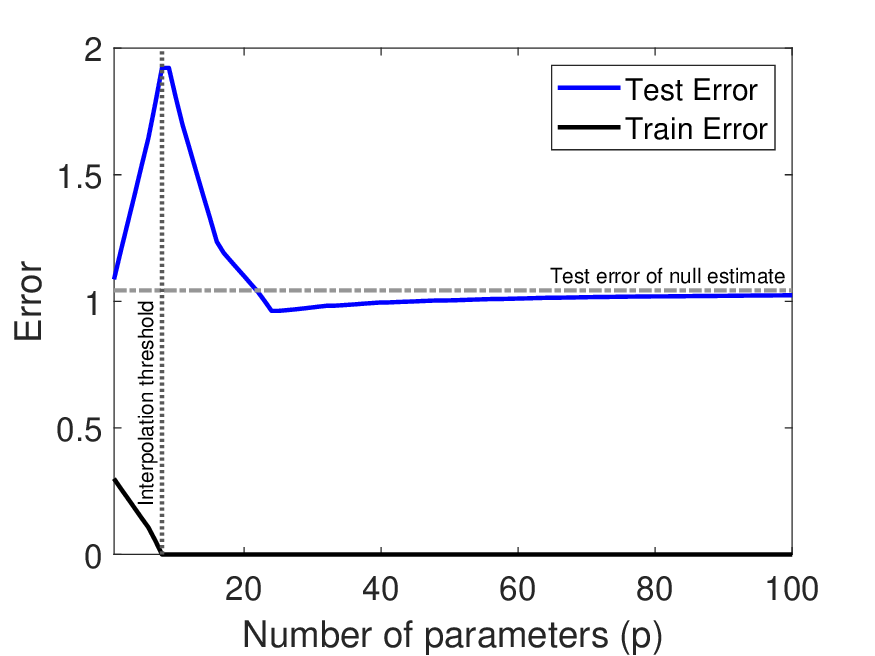}}
    \caption{Evaluation of LS regression using orthonormal cosine features as in Example \ref{example:fourieralias}. The true signal is a linear combination of the first 25 cosine functions (see blue curve in \textit{(a)}). The standard deviation of the noise is $\sigma=0.2$. The number of training data points is $n=8$. \textit{(a)}~Example of the minimum $\ell_2$-norm solution for $p=24$. \textit{(b)}~Test and training error curves for the min $\ell_2$-norm solutions with varying number of parameters $p$.}
    \label{fig:example3_cosine_combination}
\end{figure*}

Here, the true signal is of the form ${f_{\sf true}: [0,1]\rightarrow\mathbb{R}}$.
Then, we have the noisy training data 
	\begin{align}\label{eq:constantsignal}
	y_i = f_{\sf true}(x_i)  + \epsilon_i \text{ for } i = 1,\ldots,n,
	\end{align}
where ${x_i = \frac{i-\frac{1}{2}}{n}}$.
The overparameterized estimator has to reconstruct a signal over the continuous interval $[0,1]$ while interpolating the $n$ given data points using a linear combination of ${p>n}$ orthonormal cosine functions $\{\varphi_j(x)\}_{j=1}^p$, that were defined in Example \ref{example:fourieralias} for ${x\in[0,1]}$.
In other words, various interpolating function estimates can be induced by plugging different parameters ${\{\alpha_j\}_{j=1}^{p}\in\mathbb{R}}$ in 
\begin{align}
    \label{eq:example 3 estimate form}
    f(x)&=\sum_{j=1}^{p}{\alpha_j \varphi_j(x) }
    \nonumber\\
    &=\alpha_1 + \sqrt{2}\sum_{j=2}^{p}{\alpha_j \cos\left((j-1) \pi x\right) }
\end{align}
such that $f(x_i)=y_i$ for $i=1,\dots,n$. 
The assumed uniform distribution on the \textit{test} input implies that the performance of an estimate $f$ is evaluated as 
\begin{equation}
{\mathcal{E}_{\sf test} (f) = \sigma_{\epsilon}^{2} + \int_{x=0}^{1} {(f(x)-f_{\sf true}(x))^2 dx}}.    
\end{equation}
In Figure \ref{fig:example3_cosine_combination} we demonstrate results in the setting of Example \ref{example:fourieralias}. 
Figure \ref{fig:example3_cosine_combination_reconstruction_example} illustrates the minimum $\ell_2$-norm solution of the form (\ref{eq:example 3 estimate form}) for $p=24$ parameters and $n = 8$ training examples. The training and test error curves (as a function of $p$) in Figure \ref{fig:example3_cosine_combination_error_curves} show that the global minimum of test error is obtained in the overparameterized regime for $p=24$ parameters.

\begin{figure*}
    \centering
    \subfloat[]{\label{fig:aliasing_signal}\includegraphics[width=0.42\textwidth]{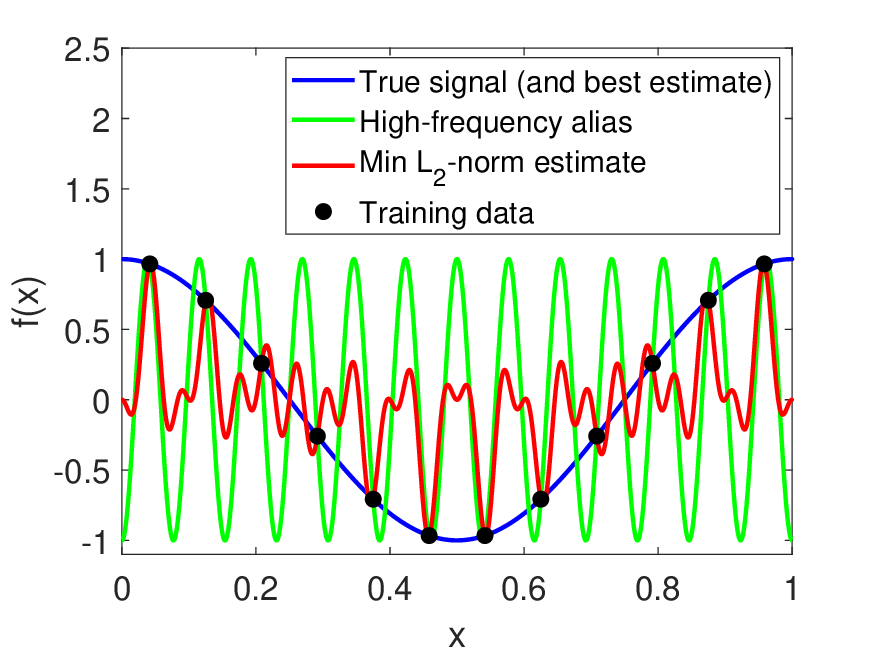}}
    \subfloat[]{\label{fig:aliasing_noise}\includegraphics[width=0.42\textwidth]{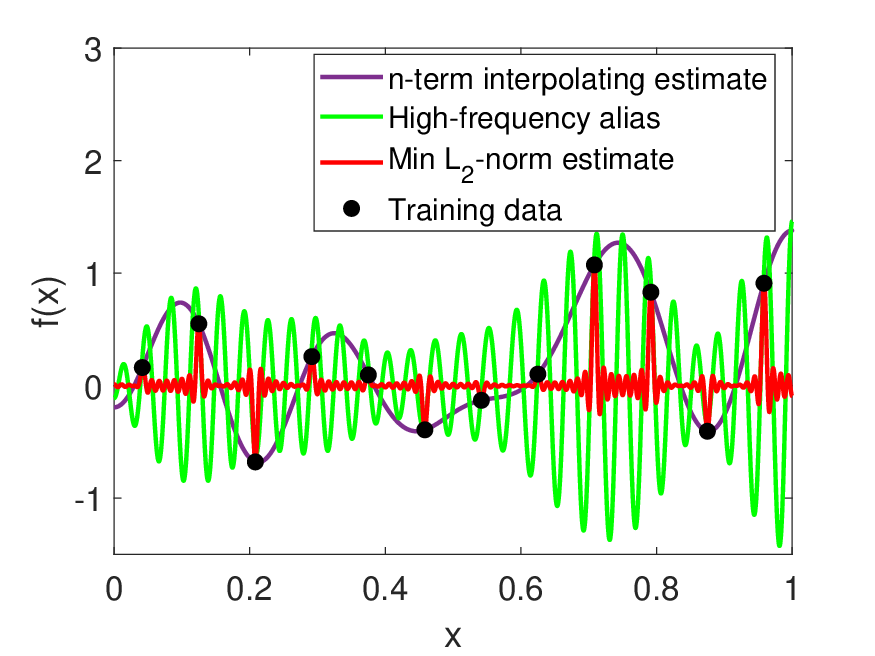}}
    \caption{Illustration of aliasing and its effect on the minimum $\ell_2$-norm interpolator. Two idealized settings of Example~\ref{example:fourieralias} are shown. 
    Left subfigure \textit{(a)}:~A noiseless case ($\sigma=0$) where the true signal is ${f_{\sf true}(x)=\cos(2 \pi x)}$ and the $n=12$ training data points are ${y_i = \cos(2\pi x_i)}$  for ${i = 1,\ldots,n}$. The trivial single-cosine alias, $f_{{\sf alias},0}(x)$ coincides with ${f_{\sf true}(x)}$ in the blue curve, a corresponding high-frequency alias $f_{{\sf alias},1}(x)$ is presented as the green curve. The minimum $\ell_2$-norm estimate for $p=48$ is presented as the red curve.
    Right subfigure \textit{(b)}:~~A noisy case where the noise standard deviation is $\sigma=0.3$, the true signal is ${f_{\sf true}(x)=0}$, and the $n=12$ training data points are ${y_i = \epsilon_i}$  for  ${i = 1,\ldots,n}$. The interpolating estimate using the first $n$ cosine functions is presented as the purple curve, a high-frequency alias of this estimate is presented as the green curve. The minimum $\ell_2$-norm estimate for $p=180$ is presented as the red curve.}
    \label{fig:aliasing}
\end{figure*}

In the discussion below, we will consider (\ref{eq:constantsignal}) in both noisy and noiseless settings (where $\epsilon_i = 0$ in the latter case). To demonstrate clearly the concept of aliasing, suppose the true signal is 
\begin{equation}
\label{eq:pure cosine function}
  f_{\sf true}(x)=\cos(k \pi x),  
\end{equation}
i.e., a single cosine function whose frequency is set by an integer ${k\in\{1,\dots,n-1\}}$. We start by considering the noiseless version of (\ref{eq:constantsignal}), which we illustrate in Figure \ref{fig:aliasing_signal}
(note that in this figure the true signal and the training data points are presented as blue curve and black circle markers, respectively).
Then, a trivial estimate that interpolates all the training data points is the unnormalized $(k+1)^{\rm th}$ cosine function: ${f_{{\sf alias},0}(x)=\frac{1}{\sqrt{2}}\varphi_{k+1}(x) = \cos(k\pi x)}$. 
Yet, $f_{{\sf alias},0}(x)$ is not the only interpolating estimate that relies on a single cosine function. 
Recall that in Example \ref{example:fourieralias} we assume $p/n$ to be an integer.
Then, using periodicity and other basic properties of cosine functions, one can prove that there is a total of $N_{\sf alias}\triangleq\frac{p}{n}$ \emph{orthogonal alias functions} 
(each in a form of a single cosine function from $\varphi_j(x)$, $j=1,\dots,p$) that interpolate the training data over the grid $x_i=\frac{i-\frac{1}{2}}{n}$, $i=1,\dots,n$. Specifically, one subset of single-cosine interpolating estimates is given by 
\begin{align}
    \label{eq:example 3 - first subset of single-cosine interpolating estimates}
    f_{{\sf alias},\ell}(x) &=\frac{(-1)^{\ell}}{\sqrt{2}}\varphi_{1+k+2\ell n}(x) 
    \nonumber\\
    &= (-1)^{\ell}\cos((k+2\ell n)\pi x) 
\end{align}
for $\ell=0,\dots,\left\lceil{\frac{N_{\sf alias}}{2}}\right\rceil-1$, 
and the second subset of single-cosine interpolating estimates is given by 
\begin{align}
    \label{eq:example 3 - second subset of single-cosine interpolating estimates}
    f_{{\sf alias},-\ell}(x)&=\frac{(-1)^{\ell}}{\sqrt{2}}\varphi_{1-k+2\ell n}(x) 
    \nonumber\\
    &= (-1)^{\ell}\cos((-k+2\ell n)\pi x)  
\end{align}
for $\ell=1,\dots,\left\lfloor{\frac{N_{\sf alias}}{2}}\right\rfloor$.
One of these aliases is shown as the green curve in Figure~\ref{fig:aliasing_signal}.
(Note that the true signal, which is shown as the blue curve in Figure~\ref{fig:aliasing_signal}, corresponds to the specific ``alias" indexed by $\ell = 0$).
We define the set of signed indices for the aliases defined in (\ref{eq:example 3 - first subset of single-cosine interpolating estimates}) and (\ref{eq:example 3 - second subset of single-cosine interpolating estimates}) together as $\mathcal{I}_{\sf alias}\triangleq\left\{-\left\lfloor{\frac{N_{\sf alias}}{2}}\right\rfloor,\dots,\left\lceil{\frac{N_{\sf alias}}{2}}\right\rceil-1\right\}$. 
Clearly, the total number of such aliases, given by $N_{\sf alias} \triangleq \lvert\mathcal{I}_{\sf alias}\rvert= \frac{p}{n}$, \textit{increases} with overparameterization. 

Importantly, \emph{any} linear combination of these aliases whose coefficients sum to one is an interpolating solution, i.e., 
\begin{equation}
f_{\vec{c}}(x) = \sum_{\ell\in\mathcal{I}_{\sf alias}} c_{\ell} f_{\sf alias,\ell}(x) \text{ such that } \sum_{\ell\in\mathcal{I}_{\sf alias}} c_\ell = 1 \end{equation}
is an interpolating solution.

Note that the $N_{\sf alias}$ alias functions in (\ref{eq:example 3 - first subset of single-cosine interpolating estimates}), (\ref{eq:example 3 - second subset of single-cosine interpolating estimates}) are scaled versions of $N_{\sf alias}$ out of the $p$ orthonormal cosine functions in our feature map (\ref{eq:cosine feature functions - definition}), (\ref{eq:cosine feature vector definition - definition}). Hence, by defining the $\alpha_j$ coefficients for the feature-based representation (\ref{eq:example 3 estimate form}) as $\alpha_{1+k+2\ell n}=\frac{(-1)^{\ell}}{\sqrt{2}}c_{\ell}$ for $\ell=0,\dots,\left\lceil{\frac{N_{\sf alias}}{2}}\right\rceil-1$; $\alpha_{1-k+2\ell n}=\frac{(-1)^{\ell}}{\sqrt{2}}c_{-\ell}$ for $\ell=1,\dots,\left\lfloor{\frac{N_{\sf alias}}{2}}\right\rfloor$; and $\alpha_j=0$ otherwise, we get the representation equivalence of 
\begin{align}
\label{eq:equality of fc and alpha representation}
     f_{\vec{c}}(x) = \sum_{\ell\in\mathcal{I}_{\sf alias}} c_{\ell} f_{\sf alias,\ell}(x)= \sum_{j=1}^{p}\alpha_j \varphi_j(x).
\end{align}
By Parseval's theorem, which uses orthonormality of the basis functions $\{\varphi_j\}_{j=1}^p$ to equate the squared norm of the continuous function $f_{\vec{c}}(x)$ and the squared norm of its discrete coefficient vector $\vecgreek{\alpha}$, we get
\begin{equation}
\label{eq:parsevel for alpha and fc}
    \Ltwonorm{\vecgreek{\alpha}} = \left\|f_{\vec{c}}\right\|_{L_2}^2.
\end{equation}
By the orthogonality of the alias functions and (\ref{eq:equality of fc and alpha representation})-(\ref{eq:parsevel for alpha and fc}), we then have
\begin{align}
    \left\|f_{\vec{c}}\right\|_{L_2}^2 &= \sum_{\ell\in\mathcal{I}_{\sf alias}} c_\ell^2 \|f_{\sf alias,\ell}\|_{L_2}^2 = \frac{1}{2}\sum_{\ell\in\mathcal{I}_{\sf alias}} c_\ell^2
\end{align}
where we used $\|f_{\sf alias,\ell}\|_{L_2}^2 = \frac{1}{2}$ that corresponds to the alias functions as defined in (\ref{eq:example 3 - first subset of single-cosine interpolating estimates}), (\ref{eq:example 3 - second subset of single-cosine interpolating estimates}). Hence, 
\begin{equation}
    \Ltwonorm{\vecgreek{\alpha}} = \frac{1}{2}\sum_{\ell\in\mathcal{I}_{\sf alias}} c_\ell^2.
\end{equation}
Consequently, the minimum $\ell_2$-norm solution corresponds to the specific linear combination of alias functions whose coefficients are 
\begin{align}\label{eq:beta-optimization-problem}
    \{\widehat{c}_\ell\}_{\ell\in\mathcal{I}_{\sf alias}} = \underset{\{c_\ell\}_{\ell\in\mathcal{I}_{\sf alias}}}{\argmin} \sum_{\ell\in\mathcal{I}_{\sf alias}} c_\ell^2 ~\text{ subject to } \sum_{\ell\in\mathcal{I}_{\sf alias}} c_\ell = 1.
\end{align}
The solution of \eqref{eq:beta-optimization-problem} is given by $\widehat{c}_\ell = \frac{1}{N_{\sf alias}}$, $\forall\ell\in\mathcal{I}_{\sf alias}$, i.e.,~a function fit that \emph{equally weights all the aliases}.
Translating back into $\alpha_j$ coefficients, the minimum $\ell_2$-norm solution (among linear combinations of the considered alias functions) is given by 
\begin{align}
    \label{eq:example 3 - parameters for min l2-norm}
    &\widehat{\alpha}_{j} = \begin{cases}
    \frac{(-1)^{\ell}}{N_{\sf alias}\sqrt{2}},& \text{for~~} j=1+k+2\ell n,~\ell=0,\dots,\left\lceil{\frac{N_{\sf alias}}{2}}\right\rceil-1 \vspace{0.05in}\\
    ~& \text{and for~~} j=1-k+2\ell n,~\ell=1,\dots,\left\lfloor{\frac{N_{\sf alias}}{2}}\right\rfloor\\
    0,              & \text{otherwise.}
\end{cases}
\end{align}
It can be proved that the coefficients in (\ref{eq:example 3 - parameters for min l2-norm}) define the minimum $\ell_2$-norm solution among all the interpolating solutions and not only among linear combinations of the set of $N_{\sf alias}$ alias functions that we consider here. 

According to (\ref{eq:example 3 - parameters for min l2-norm}), the $\ell_2$-norm of $\widehat{\vecgreek{\alpha}}$ is given by
\begin{align*}
    \Ltwonorm{\widehat{\vecgreek{\alpha}}} = N_{\sf alias} \left(\frac{1}{N_{\sf alias} \cdot \sqrt{2}}\right)^2 =  \frac{1}{2 N_{\sf alias}}  = \frac{n}{2 p}
\end{align*}
Noting that ${\expectationwrt{\vecgreek{\phi} \vecgreek{\phi}^{T}}{x}=\mtx{I}_{p}}$, we can again use Parseval's theorem on the learned function form from (\ref{eq:learned linear mapping of feature vector}) to obtain
\begin{align}\label{eq:harmlessinterpolation}
  \expectationwrt{\left({\widehat{f}(x)}\right)^2 }{x} =  \Ltwonorm{\widehat{\vecgreek{\alpha}}} = \frac{n}{2 p} 
\end{align}
where $\widehat{f}(x)=\sum_{j=1}^{p}\widehat{\alpha}_j\varphi_j(x)$. 
Note that the $\ell_2$-norm of the minimum $\ell_2$-norm solution in (\ref{eq:harmlessinterpolation}) decreases as overparameterization increases. This behavior affects the corresponding test error and also occurs in more general cases where there is noise and the true signal is more complex than a single cosine function.

In Figure \ref{fig:aliasing_signal} we consider a ``pure cosine" for the output (i.e., $f_{\sf true}$ is a single cosine function as in (\ref{eq:pure cosine function}) but in a noiseless setting where $\epsilon_i = 0$ for $i = 1,\ldots,n$) to illustrate the exact orthogonal aliases clearly in (\ref{eq:example 3 - first subset of single-cosine interpolating estimates})-(\ref{eq:example 3 - second subset of single-cosine interpolating estimates}). However, this aliasing effect would happen for arbitrary output, including when the output is pure noise (i.e., the signal is given by ${f_{\sf true}(x)=0}$ and $y_i = \epsilon_i$ for all $i = 1,\ldots,n$) as demonstrated in Figure \ref{fig:aliasing_noise}.
Thus, the results in \cite{muthukumar2020harmless} show that minimum $\ell_2$-norm interpolation provides a mechanism for combining aliases to interpolate noise in a harmless manner. 
Specifically, the noise-related error term $\testerrnoise$ from the decomposition in (\ref{eq:signalnoisedecomposition}) can be shown to scale \textit{exactly} as $\frac{\sigma_{\epsilon}^{2} n}{p}$, implying that interpolation of noise is less of an issue as overparameterization increases. 
On the other hand, the illustration in Fig.~\ref{fig:aliasing_signal} clearly shows that minimum $\ell_2$-norm interpolation is not always desirable from the perspective of signal preservation.
This aliasing perspective also recovers in an elementary manner the scalings provided by Bartlett et al.~\cite{bartlett2020benign} for the more complex anisotropic case; see \cite{muthukumar2020harmless} for further details on this calculation.
~

\subsection{Beyond minimum $\ell_2$-norm interpolation}\label{sec:l1}

Of course, the minimum $\ell_2$-norm interpolation is just \textit{one} candidate solution that could be applied in the overparameterized regime.
Research interest in this specific solution has been largely driven by  
\begin{enumerate}[label=(\alph*)]
    \item the convergence of gradient descent to the minimum $\ell_2$-norm interpolation when initialized at zero~\citep{engl1996regularization},
    \item the observation of double descent with this solution in particular, and 
    \item its relation to minimum Hilbert-norm interpolation via kernel machines~\citep{scholkopf2002learning}.
\end{enumerate}
However, as we have just seen~\citep{hastie2019surprises,muthukumar2020harmless}, the minimum $\ell_2$-norm interpolation suffers from fundamental vulnerabilities for signal reconstruction that have been known for a long time, particularly in the isotropic setting where data is effectively very high-dimensional.
It is natural to examine the ramifications of using alternative interpolators of noisy data.

One such choice would be the class of \textit{parsimonious}, or sparsity-seeking interpolators (Definition $4$ in ~\cite{muthukumar2020harmless}), which includes the minimum $\ell_1$-norm interpolator (basis pursuit~\citep{chen2001atomic}) and orthogonal matching pursuit run to completion~\citep{pati1993orthogonal}.
Such interpolators can arise from the implicit bias of other iterative optimization algorithms (such as AdaBoost~\citep{rosset2004boosting} or gradient descent run at a large initialization~\citep{woodworth2020kernel}), and comprise the gold standard for signal recovery in noiseless high-dimensional problems.
This is primarily because their sparsity-seeking inductive bias enables us to avoid the issue of signal absorption created by the minimum $\ell_2$-norm interpolator that is illustrated clearly by the red curve in Figure~\ref{fig:aliasing_signal}.
In statistical terms, these interpolators have vanishing bias.
However, their \textit{robustness} to non-zero noise constitutes a more challenging study, primarily because such interpolators do not possess a closed-form expression.
It turns out that such parsimonious interpolators, as defined above, suffer from a greater overfitting effect on noise in comparison to minimum $\ell_2$-norm interpolation. In particular,~\cite{muthukumar2020harmless} show that \emph{any} parsimonious interpolator must suffer test MSE on the order of $\frac{\sigma_{\epsilon}^{2}}{\log (p/n)}$ as a consequence of the noise alone. 
Interestingly, this lower bound was matched by an upper bound in the non-asymptotic sense (i.e., up to multiplicative constant factors) by~\cite{wang2022tight} for the case of minimum $\ell_1$-norm interpolation when the covariates are isotropic and further Gaussian.
It is interesting to note that this scaling of the MSE \emph{does} imply harmless interpolation in that the MSE arising from interpolating noise goes to $0$ as $\frac{p}{n} \to \infty$, just as in the case of the minimum $\ell_2$-norm interpolator.
However, the rate of decay of the MSE is logarithmic, which is an extremely slow rate that is often considered to be impractical for statistical purposes.

Inspired by the above,~\cite{donhauser2022fast} further provided a separate sharp analysis of minimum $\ell_q$-norm interpolators of noisy data when $1 < q \leq 2$, again for isotropic Gaussian covariates.
They uncovered an interesting tradeoff between the signal error term and the noise error term: as $q$ increases in the range $(1,2]$, the noise error term \emph{decreases} while the signal error term \emph{increases}. 
They further showed that the choice $q = 1 + \mathcal{O}\left(\frac{1}{\log \log p}\right)$ minimizes the overall MSE. 
It is also interesting to consider interpolators of a more qualitative nature: for example, do \emph{low $\ell_2$-norm interpolators} admit partially the benefits of interpolation of noise of the minimum $\ell_2$-norm interpolator?
The works~\cite{zhou2020uniform,koehler2021uniform} answered this question in the affirmative.
Finally, one can ask what the \emph{minimum test-error interpolation} looks like, and what error it incurs.
Given the tradeoff between preserving signal and absorbing noise, one would expect the best possible interpolator to use different mechanisms to interpolate the signal and noise.
Indeed, Muthukumar et al.~\cite{muthukumar2020harmless} showed that such a \emph{hybrid} interpolator, which uses minimum $\ell_1$-norm interpolation to fit the signal and minimum $\ell_2$-norm interpolation to fit the residual noise, achieves the best MSE among all possible interpolating solutions in the sparse linear model under a standard assumption of restricted isometry on the covariates.
While this hybrid interpolator is somewhat artificial,~\cite{zhou2023implicit} showed that a similar rate (up to logarithmic factors) can be obtained more naturally via running the gradient descent algorithm on a special \emph{reparameterized} deep linear architecture in the same setting as~\cite{muthukumar2020harmless}.

\subsection{Interpolation vs.~regularization: Is interpolation simply harmless or actually beneficial?\nopunct}
\label{sec:shouldweinterpolate}

The setting of overparameterized linear regression allows us to make an explicit mathematical comparison between interpolating and regularized solutions.
Consider the $s$-sparse high-dimensional linear model, one of the simplest examples of overparameterized linear regression.
In Section~\ref{sec:l1}, we mentioned that the best test error among all interpolating solutions will scale as $\Theta\left(\frac{s \log (p/s)}{n} + \frac{n}{p} \right)$.
On the other hand, \emph{optimally regularized} solutions (such as the Lasso) would incur a strictly better test (squared) error of $\Theta\left(\frac{s \log (p/s)}{n}\right)$.
Hastie et al.~\cite{hastie2019surprises} also show that optimally regularized ridge regression (where the regularization parameter can be chosen via cross-validation) dominates the minimum $\ell_2$-norm interpolation in test performance.
As one would expect, they also show that this regularization mitigates the peak in the double descent curve.
Indeed, Nakkiran et al.~\cite{nakkiran2020optimal} also demonstrate this benefit of regularization, including empirically for DNNs. Furthermore, Mignacco et al.~\cite{mignacco2020role} study how regularization mitigates the double descent peak in a classification task of separating data from a mixture of two Gaussians. 

These results suggest that optimally tuned regularization will always dominate interpolation.
From this perspective, the foundational TOPML results show that interpolation is \emph{relatively harmless}, rather than being relatively beneficial.
However, to obtain the full extent of benefit of regularization over interpolation, we need to optimally tune the regularization parameter. This is usually unknown and needs to be estimated from data.
While this can be done via estimating the noise variance, or more generally cross-validation (as formally shown in~\citep{hastie2019surprises}), such techniques usually require a strong i.i.d.~assumption on the noise in the data and will not easily work for less friendly noise models, such as adversarial, correlated, or heteroskedastic noise.
In such situations, an overly conservative choice of regularizer might preclude the benefits afforded by interpolation in settings with minimal noise.
From a practitioner's point of view, running iterative algorithms to completion (thus interpolating the data) may be more convenient (although more computationally costly) than formulating a complex early stopping rule that will admit only minimally beneficial performance in practice~\citep{neyshabur2014search}.
In summary, the decision of whether to interpolate or regularize contains several theoretical and practical nuances and often depends on the context surrounding model deployment.

\section{Overparameterized linear classification}
\label{sec:classification}

The basic regression task highlighted two central questions regarding the success of overparameterized models in practice: a) when and why there is a benefit to adding more parameters? and b)~why interpolating noise need not cause harmful overfitting?
However, most of the empirical success of neural networks is documented for \textit{classification} tasks, which pose their own intricacies and challenges.
Accordingly, in this section we review answers to these questions in the context of classification problems.
Throughout this section we will assume that the data is linearly separable (i.e., there exists a linear classifier that would attain zero classification error on the training data). 
Linear separability typically holds almost surely or with high probability in the overparameterized regime (see, e.g.~\cite{hsu2021proliferation}).

\subsection{The scope of data-dependent generalization bounds}

A benefit of overparameterization in classification problems was observed well before the double descent phenomenon. In fact, this was observed classically in the case of boosting: adding more weak learners led to better generalization empirically~\citep{drucker1996boosting,breiman1996arcing}.
An insightful classical work by Schapire et al.~\cite{schapire1998boosting} provided a candidate explanation for this behavior: they showed that the \textit{training data margin}, roughly the minimum distance to the decision boundary of any training data point, increases as more parameters are added into the model.
Intuitively, an increased training data margin means that points are classified with higher confidence, and should lead to better generalization guarantees.
More formally, generalization bounds that scale inversely with the margin (which is a training-data dependent quantity) and directly proportional to the Rademacher complexity (which is a parameter-dependent quantity, and is typically small for low norm solutions) can explain good generalization \textit{for classification} in some cases of overparameterization, where the ``effective dimension" (as we discussed for the anisotropic case in Section~\ref{sec:consistency results}) is sufficiently small with respect to $n$ \textit{and} there is no label noise in the data. In fact, these techniques can be used to explain the good generalization behavior of deep neural networks possessing a low spectral norm~\citep{bartlett1998sample,bartlett2017spectrally}.

These types of generalization bounds, coupled with the by now well-known implicit bias of first-order methods such as gradient descent towards the max-margin SVM~\citep{soudry2018implicit,ji2019implicit}, were sometimes touted as an explanation for the benefit of overparameterization and harmlessness of ``interpolation" in the sense that the logistic/cross-entropy loss is driven to zero (which is in general a stronger condition than simply achieving zero 0-1 loss on the training examples, i.e., achieving correct label predictions for all the training data). 
However, these data-dependent bounds are far from universally accepted to fully explain the good generalization behavior of overparameterized neural networks; in fact,~\citep{dziugaite2017computing,nagarajan2019uniform} show that they fall short empirically of explaining good generalization behavior.
Theoretically as well, there are several gaps that these techniques do not fill.
For example, while the margin can trivially be shown to increase as a function of overparameterization, quantitative rates on this increase do not exist and in general are difficult to show.
Even in the classic case of boosting, the \textit{worst-case} training data margin could be a too pessimistic measure: \textit{average-case} margins might do a better job in explaining good generalization behavior~\citep{gao2013doubt,banburski2021distribution}.

In the simplest cases of linear and kernel methods, the margin-based bounds can be shown to fall short from explaining this good generalization behavior in a number of ways.
First, they cannot explain good generalization behavior of the SVM in the presence of non-zero label noise (as noted in \cite{belkin2018understand}).
Second, they are tautological when the number of independent degrees of freedom in the data (or the ``effective dimension") exceeds the number of examples.
See \cite{bartlett2020failures} for a formal description of these shortcomings not only of margin-based bounds, but \textit{all} possible bounds that are functionals of the training dataset.

In summary, we see that despite the elegance and attractive generality of these classic data-dependent bounds, they fail to explain at least two types of good generalization behavior in overparameterized regimes with noisy training data.
An alternative avenue of analysis involves leveraging the developed techniques in foundational TOPML for the case of regression to bring to bear on the technically more complex classification task.
As we will summarize below, this avenue has yielded sizable dividends in resolving the above questions for linear models.

\subsection{Double descent and harmless interpolation}

In addition to~Belkin et al.~\cite{belkin2019reconciling}, Nakkiran et al.~\cite{nakkiran2019deep} first empirically observed the double descent phenomenon for deep learning architectures, such as ResNet, trained on classification tasks with \emph{noisy} labels.
Subsequently, Deng et al.~\cite{dengmodel} developed a simple logistic regression model under which the double descent phenomenon could be mathematically predicted in the proportional asymptotic regime (where $p, n \to \infty$ and $\frac{p}{n} \to \gamma$ for a fixed positive $\gamma$).
Concurrent to this, Montanari et al.~\cite{montanari2019generalization} provide a sharp formula for the classification test error of the SVM, also in the proportional regime but under more general assumptions on the data distribution and signal model.
This formula could be used to identify situations in which double descent occurs, although~\cite{montanari2019generalization} do not explore this aspect in detail.

Analyzing the classification error of the max-margin SVM is a technically significantly more difficult undertaking than, say, the regression error of the minimum $\ell_2$-norm interpolator.
This is for two reasons: a) unlike the minimum $\ell_2$-norm interpolator, the max-margin SVM typically does not have a closed-form expression, and b) the 0-1 test loss function (i.e., the frequency of wrong label prediction) is a significantly more difficult object to analyze than the regression squared loss function.
To handle these difficulties, both the works~\cite{dengmodel,montanari2019generalization} use the powerful technique of Gordon's comparison theorem~\cite{gordon1985some} along with the convex Gaussian min-max theorem~\cite{thrampoulidis2015regularized}, which characterizes the \emph{value} of the solution to a primal optimization problem in terms of a separable auxiliary optimization problem.
The ensuing expressions for classification test error are typically not closed form.
However, these works clearly show double descent behavior in the corresponding scenarios to where double descent occurs for regression.
Moreover, they demonstrate that the classical margin-based bounds are significantly more loose, and do not accurately explain the double descent behavior.
~\cite{montanari2019generalization} augmented their framework with a \emph{non-asymptotic upper bound} on their formula for test error that \emph{is} closed form and matches the 
\emph{regression} test MSE for the minimum $\ell_2$-norm interpolator~\cite{bartlett2020benign}.

Light has also been shed on the good generalization behavior of the minimum $\ell_2$-norm interpolator \textit{of binary labels} in the presence of label noise.
Concretely, Muthukumar et al.~\cite{muthukumar2021classification} showed that the minimum $\ell_2$-norm interpolator generalizes well for classification tasks in the same regimes where benign overfitting is shown to occur in linear regression~\citep{bartlett2020benign}.
More generally, Liang and Recht~\cite{liang2021interpolating} derived mistake bounds for this classifier that match the well-known bounds for the SVM.
These insights are both intuitively and formally connected to the foundational phenomenon of benign overfitting that is associated with overparameterization and minimum $\ell_2$-norm interpolation.

\subsection{Beyond regression}

Recall that we needed two ingredients for benign overfitting in linear regression: a) sufficiently low bias (which required low effective dimension in data), and b) sufficiently low noise fitting error (which required many low-energy directions in data compared to the number of examples). 
The above analyses essentially show that benign overfitting will also occur for a classification task under these assumptions. However, we can go significantly further and identify a strictly broader class of regimes under which classification will generalize even when regression may not in overparameterized regimes.
This insight started with the concurrent works \cite{muthukumar2021classification,chatterji2021finite} and has been expanded and improved upon in subsequent works~\cite{wang2020benign,wang2021benign,cao2021risk}.

The main ingredient behind this insight is the fact that for binary classification tasks, \textit{we can tolerate very poor signal recovery --- in the sense of convergence of test error to the ``null risk" --- as long as the rate of this convergence is slower than the rate of decay of the variance term arising from fitting noise}. This conceptual insight is classical and goes back to~\cite{friedman1997bias}, where it was leveraged in local nonparametric methods.
Its application to the overparameterized/interpolating regime unveils a particularly dramatic separation in performance between classification and regression tasks: in some sufficiently high-dimensional regimes, the classification error can be shown to go to $0$, but the regression error goes to the non-zero null risk, as $n \to \infty$ (and $p,d \to \infty$ with the desired proportions with respect to $n$ in order to maintain overparameterization).
While the works \cite{muthukumar2021classification} and \cite{chatterji2021finite} are concurrent, both their results and their approach are quite different.
While ~\cite{muthukumar2021classification} analyzes the minimum $\ell_2$-norm interpolator of the binary labels (which turns out to always induce effective misspecification noise\footnote{\cite{muthukumar2021classification} deal with this misspecification noise in a manner that is reminiscent of analyses of $1$-bit compressed sensing~\cite{boufounos20081}.} even when the labels themselves are clean) and makes an explicit connection to benign overfitting analyses,~\cite{chatterji2021finite} analyzes the max-margin SVM directly and directly connects to implicit bias analyses of gradient descent~\cite{soudry2018implicit,ji2019implicit}.
Nevertheless, these solutions were shown to be intricately connected in the following sense: in this same high-dimensional regime that admits a separation between classification and regression tasks, \emph{the solutions obtained by the max-margin SVM and minimum $\ell_2$-norm interpolation of binary labels exactly coincide} with high probability over the training data.
This phenomenon of support vector proliferation was first shown in \cite{muthukumar2021classification}, and further sharply characterized in \cite{hsu2021proliferation,ardeshir2021support}.
Support vector proliferation has intriguing implications for the choice of training loss functions (in particular, the square loss function) for classification that are of independent interest. 
For more details on these implications, see \cite{muthukumar2021classification} for a general discussion and \cite{hui2020evaluation} for a large-scale empirical comparison of training loss functions.

\section{Beyond overparameterized linear models}\label{sec:nonlinear}

The insights of the preceding two sections were restricted to linear models.
In this section, we provide a partial exposition of some mathematical and empirical work that aims to understand the implications of overparameterization and interpolation on more complex nonlinear models.

\subsection{Kernels and ``linearized" neural networks}\label{sec:kernels}

Kernel methods essentially learn a (potentially infinite-dimensional) linear model on a \emph{feature map} $\varphi\left({\vec{x}}\right)$ which is a nonlinear, vector-valued function of the input data $\vec{x}$, producing the feature vector $\vecgreek{\phi} \triangleq \varphi\left({\vec{x}}\right)$.
Indeed, Examples~\ref{example:function class is linear mappings with p parameters} and~\ref{example:function class is nonlinear mappings} in this survey constitute specific examples of learning with $p$-dimensional feature maps (linear and nonlinear, respectively).
When the feature map is overly high-dimensional (or even infinite-dimensional), we cannot explicitly keep track of the coefficients of the linear model in the feature space; kernel methods resolve this by writing a closed-form solution in terms of evaluations of the kernel function whose basic definition, $K(\vec{x},\vec{y}) \triangleq \left\langle \varphi\left({\vec{x}}\right), \varphi\left({\vec{y}}\right) \right\rangle$, is an inner-product between the feature vectors of two inputs (where the inner product is as defined in the relevant Hilbert space). 
Importantly, in practice the feature maps are often not explicitly invoked and the inner-product definition of $K(\vec{x},\vec{y})$ is conveniently replaced by a computationally-efficient kernel evaluation formula (such formula is not available for every feature map).
Indeed, the most popular kernel methods in practice are specified entirely by their efficient kernel formula (e.g., Gaussian kernel and other radial basis kernel functions, Laplace kernel, polynomial kernel).

We note that although the \emph{input data dimension} $d$ may be smaller than $n$, all of the kernels considered in this subsection will consist of either infinite-dimensional or $p$-dimensional feature maps where $p > n$; therefore, we continue to operate in the overparameterized regime.
Further, the results that we consider here fix the dimension of the feature map (whether infinite or finite).

A traditional kernel method with a closed-form solution for regression is the \emph{kernel ridge regression} method, which minimizes a combination of the least-squares error and a penalty on the norm of the learned function in the associated \emph{reproducing kernel Hilbert space} (RKHS).
The norm penalization, as in ridge regression in linear models, is employed to mitigate potential overfitting.
Akin to the double descent studies for linear models, it was demonstrated in \cite{belkin2018understand} that interpolating kernel methods (also called kernel \emph{ridgeless} regression) can, in fact, achieve competitive test performance to their regularized counterparts and need not always significantly overfit to noise.
It was a natural question to ask in this context whether the characterizations for harmless interpolation in overparameterized linear regression, as described in Section~\ref{sec:regression} of this survey, might have anything to say about kernel interpolation.
Recall that~\cite{bartlett2020benign} provided expressions for the bias and variance of the minimum $\ell_2$-norm interpolator for overparameterized linear regression with an arbitrary covariance matrix $\mtx{\Sigma}$ in terms of its \emph{ordered eigenvalues} $\lambda_1 \geq \lambda_2 \geq \ldots \geq \lambda_p$.
A compelling hypothesis would be that these expressions also apply to kernel interpolation where the (potentially infinite-dimensional) feature covariance matrix is now given by $\mtx{\Sigma}_{\vecgreek{\phi}} \triangleq \EE[\varphi\left({\vec{x}}\right)\varphi\left({\vec{x}}\right)^T]$, and the eigenvalues of $\mtx{\Sigma}_{\vecgreek{\phi}}$ are the eigenvalues of the associated kernel integral operator\footnote{The kernel integral operator acts on a function and is given by $\mathcal{K} f(\vec{y}) \triangleq \int_{\vec{x} \in \mathbb{R}^d} f(\vec{x}) K(\vec{x},\vec{y}) dP_{\vec{x}}$, and eigenfunctions $\{\phi_j\}_{j \geq 1}$ are defined such that $\mathcal{K} \phi_j = \lambda_j \phi_j$.}. 
The story turns out not to be quite as simple as this --- the results of~\cite{bartlett2020benign} were proved under the assumption that the coordinates of $\varphi\left(\vec{x}\right)$ were independent and sub-Gaussian, which is not the case for typical kernel methods.
Moreover, explicit characterizations of the eigenvalues $\{\lambda_j\}_{j \geq 1}$ are not always available in practical instantiations specified entirely by the kernel function (i.e., lacking in explicit definition of the feature maps). 

Despite these technical challenges, a partial understanding of kernel interpolation and its effect on noise overfitting has emerged, with the dimensionality of the \emph{input data} playing a key role. 
In a nutshell, \emph{higher dimensional data appears conducive to and necessary for harmless interpolation with typical kernel methods.}
Indeed, data dimension $d$ that grows with sample size $n$ was shown to be sufficient and necessary for harmless interpolation with the Laplace kernel~\cite{rakhlin2019consistency,mallinar2022benign}.
Interestingly, the Gaussian kernel always interpolates in a harmful manner, even on high dimensional data~\cite{mallinar2022benign}.
In~\cite{liang2020just,bartlett2021deep}, a general class of kernels was considered for showing that harmless interpolation occurs when the data dimension $d$ is proportional to the sample size $n$ and the learning is effectively overparameterized.
A \emph{polynomially proportional regime} $n \propto d^q$ has also been considered and upper bounds on the noise overfitting error of a more complex ``multiple-descent type" have been derived for \emph{inner product kernels} of the form $K(\vec{x},\vec{y}) = h(\langle \vec{x}, \vec{y} \rangle)$~\cite{liang2020multiple} (a matching lower bound remains to be obtained). It was empirically and theoretically shown in \cite{mallinar2022benign} that a variety of kernel methods on constant-dimensional data exhibit a more ``tempered" overfitting effect, rather than a benign one.
In addition to the above analyses, which directly analyze the test error of specific kernels, a plethora of works have established equivalence between the test error of a general kernel method and an equivalent linear model with the same eigenvalues (see, e.g.,~\cite{tsigler2020benign,mei2022generalization,cheng2024dimension,gavrilopoulos2024geometrical,mcrae2022harmless,misiakiewicz2024non}).
In essence, these works significantly weaken the independent sub-Gaussian assumption of~\cite{bartlett2020benign}, and provide powerful predictions when the eigenvalues of the kernel integral operator are characterizable or can be estimated (e.g. for Gaussian and Laplace kernels~\cite{bietti2021deep}) --- of course, this is often not the case in practice.
A unified and sharp analysis that recovers all of the mentioned results in this section remains elusive.

The discovery that high-dimensional input data appears essential to avoid noise overfitting prompted a re-examination of the \emph{bias}, or \emph{approximation error}, of kernel methods on complex ground-truth functions or signals.
Indeed, the expressive power of kernel methods in general is known to degrade significantly when the data dimension grows with sample size (see, e.g.,~\cite{belkin2018approximation,ghorbani2019limitations,donhauser2021rotational}).
As a specific example, the expressive power of inner-product kernel methods turns out to be limited to that of a $d$-dimensional linear model in the proportional regime where $d \propto n$~\cite{el2010spectrum}.
Tight bounds on the bias have also been derived for the so-called polynomial regime $n \propto d^q$ under specific distributional assumptions on the input data (e.g. uniformly distributed on the sphere/Boolean hypercube) using tools from harmonic analysis~\cite{ghorbani2021linearized,mei2022generalization}. 
At a high level, much as in the case of linear models, an intriguing tradeoff arises between preserving signal and absorbing noise as we vary input data dimension: while higher data dimension can enable harmless interpolation of noise, it also severely limits the expressive power of the kernel method!
(Indeed, \cite{aerni2023strong} makes this tradeoff very explicit with convolutional kernels).
As in our study of various interpolators for linear models, there turns out to not be a free lunch.

Finally, we mention that precise test error predictions for kernel ridge/ridgeless regression have interesting implications for certain ``linearized" neural networks that are trained in the ``lazy" regime where the weights on the hidden neurons barely move from their initializations, and the neural network behaves akin to a finite-dimensional approximation of a kernel method given by the \emph{neural tangent kernel (NTK)}~\cite{jacot2018neural,chizat2019lazy}.
Indeed, the eigenspectrum of the NTK has been related to the Laplace kernel~\cite{chendeep,geifman2020similarity}.
An even simpler connection can be made to a training protocol for randomly initialized neural networks where only the last (linear) layer is trained, which amounts to a \emph{random features} model, also a finite-dimensional approximation of a kernel method~\cite{rahimi2007random}.
See~\cite{misiakiewicz2024six} for a tutorial on error analysis of these linearized neural networks.
Note that these connections to kernel methods only apply to sufficiently \emph{wide} overparameterized neural networks and generally are not believed to explain the remarkable generalization power of deep learning in practice.

\subsection{Empirical observations in deep learning/neural networks}
\label{subsec:Double Descent in Deep Learning for Classification}

Thus far, we have traced the foundational efforts to understand generalization of overparameterized modern ML models, from seminal empirical observations made in neural networks~\cite{zhang2017understanding,neyshabur2014search,belkin2019reconciling} to a theoretical foundation that was laid primarily for linear models (and some kernel methods, as mentioned in the preceding section).
What does any of this have to say for overfitting in overparameterized \emph{neural networks}?
The answer to this question remains unclear and murky, primarily because a comprehensive theory of generalization for neural networks (e.g., outside the aforementioned lazy regime) remains elusive.
In this section, we summarize insightful experiments conducted since the initial studies~\cite{zhang2017understanding,neyshabur2014search,belkin2019reconciling} and a few initial attempts at corresponding theory.

As in the case of kernel methods, it appears that data dimension plays a crucial role in determining whether neural networks can interpolate noisy data in a harmless manner.
A line of work~\cite{frei2022benign,frei2023benign,kornowski2023tempered} showed that \textit{neural networks that are overparameterized because they operate on very high-dimensional input data} (i.e., $d = \Omega(n^2)$) is conducive to harmless interpolation by a $2$-layer neural network (multi-layer perceptron) trained by gradient descent under the additional assumption that the data is comprised of sufficiently separated, linearly separable mixtures. 
While the dynamics of gradient descent here are nonlinear and exit the ``lazy regime", the ultimate performance of the neural network model resembles that of a linear classifier~\cite{frei2023benign}.
An arguably more realistic scenario is one where the data dimension $d$ is fixed and the sample size $n$ grows to infinity.
Here, experiments by~\cite{mallinar2022benign} on both real and synthetic datasets show that practical neural networks tend to exhibit a more \emph{tempered} overfitting effect\footnote{While they also show a parallel phenomenon for popular kernel methods, we do not expect theory from kernel methods to provide even intuitive or high level explanations for neural network generalization --- indeed, we know that neural networks outperform kernel methods in a variety of tasks involving high data dimension (see, e.g.~\cite{yehudai2019power,ghorbani2020neural} for early examples).}, wherein their binary classification test error goes to a constant that is strictly greater than $0$, but strictly less than $1/2$, as $n \to \infty$. 
The observation of tempered overfitting in neural networks was theoretically explained for the special case $d = 1$ in~\cite{kornowski2023tempered}.

We next turn to empirical observations of the double descent phenomenon on large-scale neural network architectures. The seminal paper by Belkin et al.~\cite{belkin2019reconciling} demonstrated double descent on multi-layer perceptron neural networks. Then, the empirical study of the double descent was significantly extended by Nakkiran et al.~\cite{nakkiran2019deep} to deeper neural networks, including architectures such as the residual network (ResNet) for image classification and also representative examples of transformer architectures for language translation. 

In \cite{nakkiran2019deep}, the train and test errors of convolutional networks (such as ResNet-18 and a 5-layer CNN) are evaluated as a function of two varying quantities: 
\begin{itemize}
    \item The proportional scaling of the DNN width, namely, a multiplicative factor that is applied on the number of convolution filters in all the convolution layers in the examined architecture. Indeed, the DNN width is a natural way to vary the number of parameters.
    \item The number of training epochs\footnote{The common practice in deep learning is to train via iterative optimization where in each iteration a batch (or mini-batch) of training examples (i.e., a randomly chosen subset of the complete training dataset) is used to approximate the training loss gradient. Accordingly, an \textit{epoch} is a training interval (usually of many iterations) that goes over all the examples in the training dataset.}. The number of training epochs reflects if, and how much, regularization through early stopping is applied. Specifically, in training via iterative optimization, the model gradually adapts its parameter values (also called weights) to the training dataset; a setting-dependent number of iterations is required to perfectly fit the training data, and therefore overfitting can be avoided by stopping the training at an earlier iteration. Accordingly, the number of training epochs effectively reflects regularization strength and, therefore, also the learned model complexity. 
\end{itemize}
This empirical analysis elucidated when double descent of the test error can be expected in deep models. Label noise was shown as a significant promoter of double descent. Moreover, two complementary results were established in \cite{nakkiran2019deep}: (i) A double descent of the test error as a function of the DNN width can emerge if the training duration is sufficiently long (i.e., the number of training epochs is sufficiently high). (ii) A double descent of the test error as a function of the number of training epochs (an ``epoch-wise'' double descent) can emerge if the DNN is sufficiently wide. 

Somepalli et al.~\cite{somepalli2022can} empirically elucidated how the fragmentation of decision regions in the learned image classifier (a ResNet-18 model) has a double descent trend that follows the test error's double descent as function of the model width. Specifically, the decision region fragmentation peaks together with the test error peak. This fragmentation of decision regions (also referred to as region count, e.g., in \cite{li2025understanding}) is associated with the inductive bias and model complexity that the DNN architecture and its parameterization level form. 

Motivated by the empirical elucidation of the epoch-wise double descent in DNN training in \cite{nakkiran2019deep}, Heckel and Yilmaz \cite{heckel2021early} theoretically analyzed this phenomenon for \textit{underparameterized} linear regression models and two-layer networks (plus empirical analysis for 5-layer CNN and ResNet-18 classification models). Their findings showed that the epoch-wise double descent can be due to the learning of different model parts (e.g., parameters corresponding to input components with different variances in a linear model, or different layers in a multi-layer model) at different times during the training process, which makes the epoch-wise test error curve a sum of several bias-variance trade-off shapes that together form a double descent shape. 
Beyond that, the foundational understanding of the epoch-wise double descent in overparameterized settings still requires further research. 

\section{Subspace learning for dimensionality reduction}
\label{sec:subspace learning}

Initial research on interpolating solutions focused on the supervised learning paradigm; specifically, regression and classification tasks. 
In this section, we overview another research direction that examines interpolation phenomena in unsupervised and semi-supervised settings for subspace learning tasks.

\subsection{Overparameterized subspace learning via PCA}
\label{subsec:subspace learning via PCA}

The study by Dar et al.~\cite{dar2020subspace} on overparameterized subspace learning was one of the first to explore interpolation phenomena beyond the realm of regression and classification. 
The study in \cite{dar2020subspace} starts by considering an overparameterized version of a linear subspace learning problem, which is commonly addressed via principal component analysis (PCA) of the training data. Recall that PCA is an unsupervised task, in contrast to the supervised nature of regression and classification. Let us examine the following non-asymptotic setting in which the learned subspace can interpolate the training data. 
\begin{example}[Subspace learning via PCA in linear feature space]
\label{example:Subspace learning via PCA in linear feature space}
The input is a $d$-dimensional random vector $\vec{x}$ that satisfies the following model 
\begin{equation}
    \label{eq:subspace learning data model}
    \vec{x} = \mtx{B}_m \vec{z} + \vecgreek{\epsilon}
\end{equation}
where $\mtx{B}_m$ is a $d\times m$ real matrix with orthonormal columns, $\vec{z}\sim\mathcal{N}(\vec{0},\mtx{I}_m)$ is a random latent vector of dimension $m$, and $\vecgreek{\epsilon}\sim\mathcal{N}(\vec{0},\sigma_{\epsilon}^{2}\mtx{I}_d)$ is a $d$-dimensional noise vector independent of $\vec{z}$. 
The training dataset $\mathcal{D}=\{\vec{x}_i\}_{i=1}^{n}$ includes $n$ examples of $\vec{x}$ after centering with respect to~their sample mean; specifically, $n$ examples $\widetilde{\vec{x}}_1,\dots,\widetilde{\vec{x}}_n$ are i.i.d.~drawn from the distribution defined by (\ref{eq:subspace learning data model}), then we set $\vec{x}_i\triangleq \widetilde{\vec{x}}_i - \widetilde{\vecgreek{\mu}}$ where $\widetilde{\vecgreek{\mu}}\triangleq \frac{1}{n}\sum_{i=1}^{n} \widetilde{\vec{x}}_i$.

The training examples from $\mathcal{D}$ are also organized as the rows of the $n\times d$ matrix $\mtx{X}$.

The true dimension $m$ of the linear subspace in the noisy model of Equation~\eqref{eq:subspace learning data model} is usually unknown. Hence, we will consider dimensionality reduction operators that map  a $d$-dimensional input to a $k$-dimensional representation, where $k<d$ and may differ from the unknown $m$.  For the definition of our unsupervised function class, we consider a set of ${\{\vec{u}_j\}_{j=1}^{d} \in \mathbb{R}^{d}}$ real orthonormal vectors that form a basis for $\mathbb{R}^{d}$. Further, we consider ${p\in\{1,\dots,d\}}$. The $d\times p$ matrix ${\mtx{U}_p\triangleq [\vec{u}_1,\dots,\vec{u}_p]}$ will be utilized to project inputs onto the $p$-dimensional feature space spanned by ${\{\vec{u}_j\}_{j=1}^{p}}$.
We define ${\mathcal{F}_{p,k}^{\sf DR,orth}(\{\vec{u}_j\}_{j=1}^{d})}$ as a class of dimensionality reduction operators from a high ($d$) dimensional space to a low ($k$) dimensional space.
Specifically, for $k\le p \le d$, we define
\begin{align}
\label{eq:function class for subspace learning via PCA}
&\mathcal{F}_{p,k}^{\sf DR,orth}(\{\vec{u}_j\}_{j=1}^{d}) \triangleq \Bigg\{ f(\vec{x})= \mtx{A}_{k}^T \mtx{U}_{p}^T \vec{x} ~\Bigg\vert~ \vec{x}\in\mathbb{R}^{d}, \mtx{A}_{k}\in\mathbb{R}^{p\times k}, \mtx{A}_{k}^{T}\mtx{A}_{k}=\mtx{I}_{k} \Bigg\}. 
\end{align}
Here, $\mathcal{F}_{p,k}^{\sf DR,orth}(\{\vec{u}_j\}_{j=1}^{d})$ is parameterized by matrices $\mtx{A}_{k}$ that have $k$ orthonormal columns and operate in the $p$-dimensional feature space. 
Observe that each function ${f\in \mathcal{F}_{p,k}^{\sf DR,orth}(\{\vec{u}_j\}_{j=1}^{d})}$ has a matching reconstruction operator ${g_f (\vec{z})=\mtx{U}_{p}\mtx{A}_{k}\vec{z}}$ that maps a low dimensional vector ${\vec{z}\in{\mathbb{R}^{k}}}$ back to the high ($d$) dimensional input space; moreover, note that $f(g_f(\vec{z})) = \vec{z}$. 
\end{example}

We denote the $p$-dimensional feature vector of $\vec{x}$ as $\vecgreek{\phi}\triangleq \mtx{U}_p^T \vec{x}$ and the training feature matrix as $\mtx{\Phi}\triangleq\mtx{X}\mtx{U}_{p}$. Then, the training optimization procedure 
\begin{equation}
\label{eq:example 4 - training optimization - formulation in f terms}
\widehat{f} \in \argmin_{f\in\mathcal{F}_{p,k}^{\sf DR,orth}(\{\vec{u}_j\}_{j=1}^{d})} \frac{1}{n}\sum_{i=1}^{n}{ \Ltwonorm{ g_f(f(\vec{x}_{i})) - \vec{x}_{i} } }, 
\end{equation}
(which aims to minimize the reconstruction error) can also be formulated as 
\begin{align}
\label{eq:example 4 - training optimization - formulation in A_k terms}
\widehat{\mtx{A}}_{k} &\in \argmin_{\mtx{A}_{k}\in\mathbb{R}^{p\times k}:~\mtx{A}_{k}^{T}\mtx{A}_{k}=\mtx{I}_{k}} \frac{1}{n}\sum_{i=1}^{n}{ \Ltwonorm{ \mtx{U}_{p}\mtx{A}_{k}\mtx{A}_{k}^T \mtx{U}_{p}^T \vec{x}_{i} - \vec{x}_{i} } } 
\\
\label{eq:example 4 - training optimization - formulation in A_k terms - feature domain}
&= \argmin_{\mtx{A}_{k}\in\mathbb{R}^{p\times k}:~\mtx{A}_{k}^{T}\mtx{A}_{k}=\mtx{I}_{k}} \frac{1}{n}\Frobnorm{ \mtx{\Phi} \left({\mtx{I}_{p}-\mtx{A}_{k}\mtx{A}_{k}^{T}}\right) }
\\
\label{eq:example 4 - training optimization - formulation in A_k terms - PCA form}
&= \argmax_{\mtx{A}_{k}\in\mathbb{R}^{p\times k}:~\mtx{A}_{k}^{T}\mtx{A}_{k}=\mtx{I}_{k}} \mtxtrace{\mtx{A}_{k}^{T}\mtx{\Phi}^{T} \mtx{\Phi} \mtx{A}_{k} }.
\end{align}
Note that the formulation in (\ref{eq:example 4 - training optimization - formulation in A_k terms - PCA form}) shows that the learning problem can be addressed via PCA of ${\widehat{\mtx{\Sigma}}_{\vecgreek{\phi}}\triangleq\frac{1}{n}\mtx{\Phi}^{T} \mtx{\Phi}}$, which is the sample covariance matrix of the training features. We denote 
\begin{equation}
\label{eq:subspace learning - sample covariance in feature space - rank upper bound}
{\rho\triangleq {\rm rank}\left(\widehat{\mtx{\Sigma}}_{\vecgreek{\phi}}\right)\le \min \{p,n-1\}}.    
\end{equation}
The maximal rank is the minimum of the feature space dimension $p$, and  $n-1$ that upper bounds the dimension of the span of the centered $n$ examples (that were independent before centering) in the training dataset; this maximal rank is attained almost surely under the noisy data model (\ref{eq:subspace learning data model}). If $k\le \rho$, then $\mtx{A}_{k}$ is formed by the $k$ principal eigenvectors of $\widehat{\mtx{\Sigma}}_{\vecgreek{\phi}}$; otherwise, $\mtx{A}_{k}$ is formed by the $\rho$ principal eigenvectors of $\widehat{\mtx{\Sigma}}_{\vecgreek{\phi}}$ and $k-\rho$ orthonormal vectors that span a $(k-\rho)$-dimensional subspace of the nullspace of $\widehat{\mtx{\Sigma}}_{\vecgreek{\phi}}$. 

The training error in the data domain is equivalent to the optimization costs in (\ref{eq:example 4 - training optimization - formulation in f terms}) and (\ref{eq:example 4 - training optimization - formulation in A_k terms}), i.e., 
\begin{equation}
\label{eq:subspace learning - train error}
  \mathcal{E}_{\sf train}\left(f\right)=\frac{1}{n}\sum_{i=1}^{n}{ \Ltwonorm{ g_f(f(\vec{x}_{i})) - \vec{x}_{i} } }.  
\end{equation}
Here, we are also interested in the training error in the $p$-dimensional feature space as given by the optimization cost in (\ref{eq:example 4 - training optimization - formulation in A_k terms - feature domain}), namely, 
\begin{equation}
\label{eq:subspace learning - train error in feature space}
  \mathcal{E}_{\sf train,feat}\left(\mtx{A}_{k}\right)=\frac{1}{n}\sum_{i=1}^{n}{ \Ltwonorm{ \mtx{A}_{k}\mtx{A}_{k}^T \vecgreek{\phi}_{i} - \vecgreek{\phi}_{i} } }.  
\end{equation}
where $\vecgreek{\phi}_{i}\triangleq\mtx{U}_p^T \vec{x}_i$ is the feature vector of the $i^{\sf th}$ training example.
A subspace, defined by $\mtx{A}_{k}$, is considered interpolating in the feature space if it achieves zero training error in the feature space, i.e., $\mathcal{E}_{\sf train,feat}\left(\mtx{A}_{k}\right)=0$.

The test error is given by expected reconstruction error (in the data domain) for a fresh example, 
\begin{equation}
\label{eq:subspace learning - test error}
  {\mathcal{E}_{\sf test}\left(f\right)=\expectationwrt{\Ltwonorm{ g_f(f(\vec{x})) - \vec{x} }}{\vec{x}}}.  
\end{equation}

\begin{figure*}
    \centering
    \subfloat[]{\label{fig:PCA_baswed_subspace_learning_DCT_features__train_errors}\includegraphics[width=0.46\textwidth]{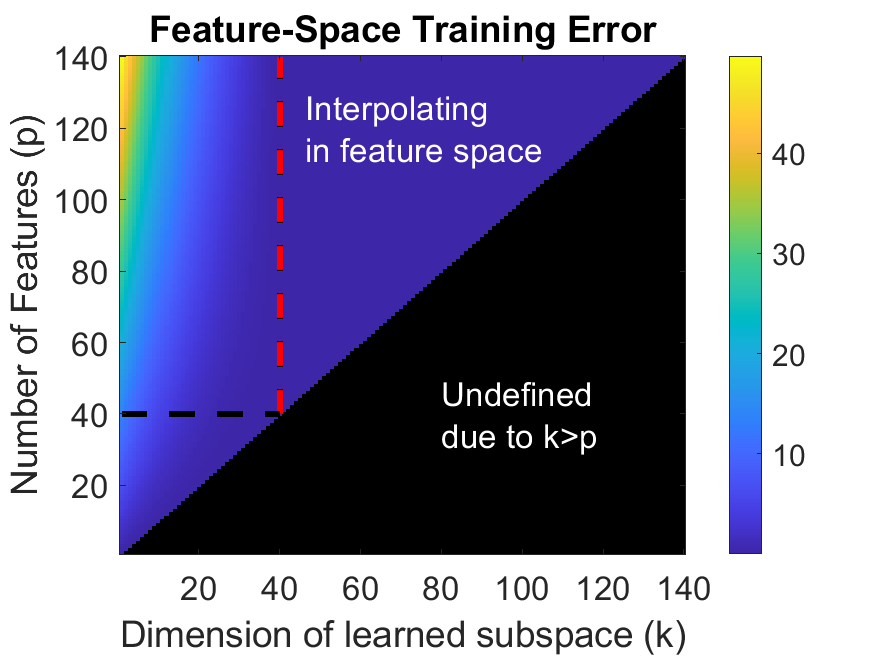}}
    \subfloat[]{\label{fig:PCA_baswed_subspace_learning_DCT_features__test_errors}\includegraphics[width=0.46\textwidth]{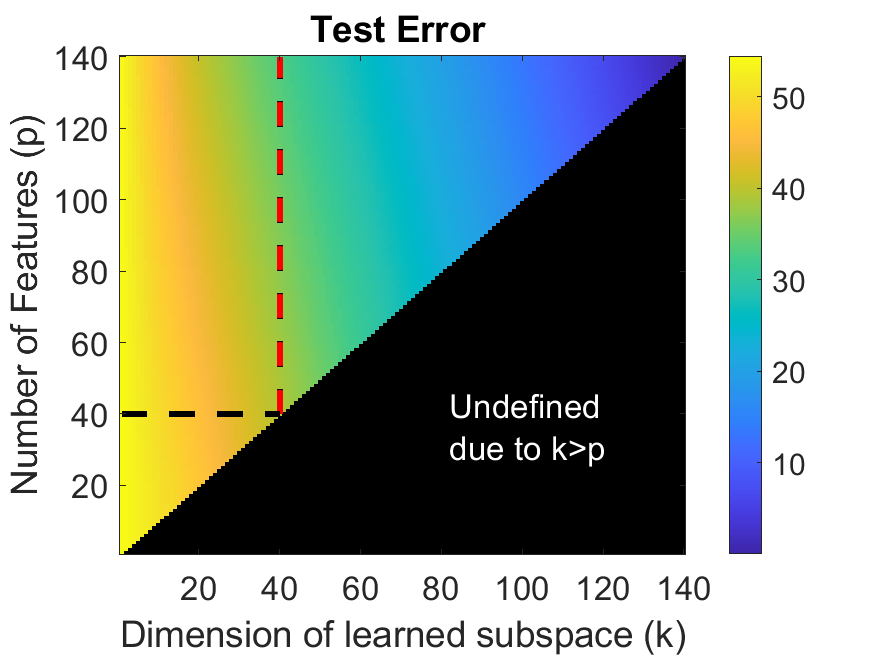}}
    \caption{Empirical results for subspace learning based on a class $\mathcal{F}_{p,k}^{\sf DR,orth}(\{\vec{u}_j\}_{j=1}^{d})$ of linear dimensionality reduction operators as described in Example \ref{example:Subspace learning via PCA in linear feature space}. Here, DCT features are employed, $d=140$, $m=20$, $n=40$. 
    Considering the data model in (\ref{eq:subspace learning data model}), ${\sigma_{\epsilon}=0.5}$ and the true subspace is determined by $\mtx{B}_m$ whose $j^{\sf th}$ column is the properly normalized sum of columns ${\{j+20s\}_{s=0}^{\frac{d}{m}-1}}$ from the ${d\times d}$ DCT matrix. 
    In \textit{(a)} we present the feature-space training error (i.e., as in the optimization cost in (\ref{eq:example 4 - training optimization - formulation in A_k terms - feature domain})). In \textit{(b)} we show the test error. Both errors are evaluated as a function of the learned subspace dimension $k$ and the number of features $p$. The $(p,k)$ pairs above the black dashed line correspond to overparameterized settings. The $(p,k)$ pairs on or to the right of the red dashed line (and also satisfy $k\le p$) correspond to rank-overparameterized settings. 
    }
    \label{fig:PCA-based subspace learning}
\end{figure*}

In~\cite{dar2020subspace}, two types of overparameterization are defined for PCA-based subspace learning as in Example \ref{example:Subspace learning via PCA in linear feature space}, which we demonstrate for DCT features in Figure \ref{fig:PCA-based subspace learning}: 
\begin{itemize}
	\item A learned linear subspace is \textit{overparameterized} if $p>n$, i.e., the number of features is larger than the number of training examples. This implies that the sample covariance matrix of the (centered) training features is rank deficient (which is guaranteed for $p\ge n$ due to using $n$ centered training examples). However, the learned subspace (of dimension $k$) is not necessarily interpolating in the feature space; this is because overparameterization in the sense of $p>n$ does not tell if $k$ is higher/equal or lower than $n$ (or the rank $\rho$). 
    In Fig.~\ref{fig:PCA-based subspace learning}, the domain of overparameterized solutions is located above the black dashed line.
	
	\item A learned linear subspace is \textit{rank overparameterized} if it is overparameterized (i.e., ${p>n}$) and its dimension is at least the number of centered training examples (i.e., ${k\ge n}$). Recall that, by (\ref{eq:subspace learning - sample covariance in feature space - rank upper bound}), for ${p>n}$, $n-1$ is the maximal rank of the sample covariance matrix of the training feature vectors, hence 
	${k\ge n}$ implies that the sample covariance matrix of the training features is rank deficient (due to ${p>n}$) in a way that introduces degrees of freedom to the construction of the learned subspace. Specifically, the learned subspace interpolates the training data in the $p$-dimensional feature space\footnote{\cite{zhuo2021computational} defines rank overparameterization in matrix sensing problems. However, the rank overparameterization of \cite{zhuo2021computational} is not related to interpolation nor the number of training examples, but only means that the solution matrix has a higher rank than the true matrix in the data model. This is in contrast to the rank overparameterization defined in \cite{dar2020subspace} for  studying overparameterization in the sense of interpolating solutions for subspace learning problems.}. Note that $k=n-1$ allows interpolation without degrees of freedom but we consider rank overparameterization as when there are also degrees of freedom in the learned subspace. In Fig.~\ref{fig:PCA-based subspace learning}, the domain of rank overparameterized solutions is located on and to the right of the red dashed line. 
\end{itemize}
Interpolating linear subspaces, which are constructed by standard PCA, do not exhibit double descent phenomena in their test errors. 
This is also aligned with earlier analytical results in \cite{paul2007asymptotics,johnstone2009consistency,shen2016general} from the related works in the area of high-dimensional statistics. Intuitively, the lack of a double descent peak in PCA is because there is no (actual or effective) inversion or pseudoinversion of the sample covariance matrix, which is the numerical mechanism that creates the peaking test error for linear regression with an ill-conditioned sample covariance matrix.

Motivated by the lack of double descent phenomena in PCA-based subspace learning, \cite{dar2020subspace} define a new family of subspace learning problems with varying levels of supervision and orthonormality constraints. By this, it was empirically shown that the double descent phenomenon becomes more evident as the  subspace learning problem is more supervised and less orthonormally constrained. Technically, the double descent emerges when the increased supervision induces a sufficiently strong effect (which requires sufficiently weak constraints/regularization) of matrix inversion whose numerical stability depends on the sample covariance matrix. 

Beyond linear subspace learning, nonlinear subspaces can be learned by multi-layer autoencoder networks with nonlinear activation functions. \cite{radhakrishnan2020overparameterized,abitbul2024how} studied the memorization in overparameterized autoencoders and the ability to recover training examples from their degraded versions and the trained autoencoder.

\section{Additional overparameterized learning problems}
\label{sec:additional problems}

In this section, we survey pioneering research on interpolating models in various other modern ML tasks, including data generation, transfer learning, pruning of learned models, and dictionary learning.

\subsection{Data generation}
\label{subsec:data generation}

Data generation applications aim to produce new instances from a data class whose accurate description is unknown and only a set of its examples is given.
Data generation is ubiquitous in deep learning practice, including various architectures of generative adversarial network (GAN) concept \citep{goodfellow2014generative} and diffusion models. 

\subsubsection{Overparameterization in GANs}
The idea of a GAN is to jointly train a generator network that produces synthetic instances from the desired class, and a discriminator network that evaluates the generator performance by identifying its synthetic outputs versus true examples from the data class. 
Practical GANs include highly complex generator and discriminator networks and therefore have overparameterization aspects to explore.

Luzi et al.~\cite{luzi2021interpolation} provide the first explicit study on interpolation phenomena (including double descent) in overparameterized GANs. 
The study \cite{luzi2021interpolation} is focused mainly on linear GANs that are overparameterized in the sense that their learned latent dimension is large with respect to the number of examples. 
This perspective on overparameterized GANs reveals a new behavior of test errors when examined with respect to~the learned latent dimension: all interpolating solutions have the same generalization ability, which is often much worse than the best generalization in the non-interpolating (i.e., underparameterized) range of solutions. 
This theoretical result applies to generative models beyond linear GANs: interpolating solutions generalize equivalently in any generative model that is trained through minimization of a distribution metric or $f$-divergence.
Moreover, \cite{luzi2021interpolation} propose the concept of pseudo-supervised training of GANs, which is done by associating the data examples with random latent representations drawn from an isotropic Gaussian distribution. This approach induces test error curves that have double descent and, sometimes, triple descent shapes that can promote improved generalization and reduced training time.

\subsubsection{Overparameterization in diffusion models}
In recent years, diffusion models have become the leading approach for learning data distributions and sampling new examples from them. The idea is to learn a model that can reverse the forward diffusion process of gradually adding noise to a data example until it becomes complete noise; then, the learned model is able to process a given pure noise into a new data example from the distribution. Popular models incorporate text guidance by which users can describe the specific data example they want that the pure noise would eventually transform into.

Memorization of training examples has various forms and implications in practical diffusion models for image generation. This was shown by early empirical research of this topic. 
 Carlini et al. \cite{carlini2023extracting} show that it is possible to extract some of the memorized training examples quite accurately. 
 Somepalli et al. \cite{somepalli2023diffusion} show that memorized training examples may appear in newly generated examples as semantic (non-exact) replications of original training examples. \cite{carlini2023extracting,somepalli2023diffusion} show that training data extraction and replication is promoted by training examples that are duplicated or near-duplicated in the large dataset of the (pretrained) diffusion model. \cite{somepalli2023understanding} emphasize the role of text conditioning on the training data replications in generated images, specifically showing that diversity of captions in training data can mitigate replications even for duplicated images in the training dataset.
Importantly, these extraction and replication phenomena are related to, but distinct from, interpolation as defined for supervised learning. In supervised learning, interpolation implies exactly matching all the input-output training examples, whereas extraction or replication in a generative model implies that generated examples coincide with or resemble particular training examples; such extraction/replication behavior may occur for only a subset of the training data.

On the theory side, George et al.~\cite{george2026denoising} provide precise asymptotic formulations for the test and train errors of diffusion models of the denoising score matching (DSM) form \cite{vincent2011connection}. Their analysis of generalization and memorization includes high-dimensional settings and Gaussian distributions, showing that the transition between the generalization and memorization regimes is at $p=n$ (i.e., where the number of random features equals the number of training examples). Moreover, they consider the statistical approximation in the DSM loss function that uses multiple noise examples (per data example) and show that increasing the number of noise realizations can improve generalization for $p<n$ whereas increase memorization in the high-dimensional case $p>n$. 

\subsection{Transfer learning}
\label{subsec:transfer learning}

Transfer learning is a popular approach for training a DNN to solve a desired target problem, based on another DNN that was already learned (i.e., pretrained) for a related source problem. 
In practice, one or more layers of the source DNN are transferred to the target DNN and fixed, fine-tuned, or used to initialize a full training process.
Provided that the source DNN is adequately trained using a large dataset, such transfer learning helps to address the challenges in training a highly overparameterized target DNN using a relatively smaller amount of data, and possibly faster (i.e., using less compute resources). 

A solid base for a fundamental understanding of transfer learning was laid by analytical results for transfer learning under linear models  \citep{lampinen2018analytic,dar2020double,obst2021transfer,dar2024common}.
The first study on double descent phenomena in transfer learning was provided by Dar and Baraniuk \cite{dar2020double} for a setting of two linear regression problems, where a subset of the target parameters is fixed to values transferred from a pretrained least-squares model of a related source task. 
In \cite{dar2020double}, they analytically characterize a two-dimensional double descent phenomenon of the test error of the target task when evaluated with respect to the number of free parameters in the target and source solutions.
They defined a noisy linear model to relate the true parameters of the target and source tasks and showed how the generalization error of the target task is affected by the interplay among the structure of the task relation, layout of the true parameters, and the subset of transferred parameters. 
Their non-asymptotic results describe when parameter transfer is useful, and emphasize a fundamental tradeoff between overparameterized learning and transfer learning. 
In particular, transferring more parameters from the source model can provide useful inductive bias (if the tasks are sufficiently related) but it also limits the number of free target parameters and therefore limits the actual parameterization level and the potential gains from high overparameterization. 

The idea that transfer learning counterbalances overparameterization is further emphasized in a second study by Dar et al.~\cite{dar2024common}, which utilize the pretrained least-squares source model to explicitly regularize the target linear regression problem. 
Specifically, the target task solution is penalized for its $\ell_2$-distance from the source task solution. 
This method can also be perceived as implementing a fine tuning procedure.
This approach for transfer learning is able to regularize overparameterized learning in a way that mitigates the test error peak that is typically present in double descent phenomena. The analysis includes precise asymptotic formulations of the test error. Moreover, the mean and covariance of the pretrained source model, which is a least squares solution (and specifically the minimum $\ell_2$-norm interpolating solution when the source learning is overparameterized) are used in the analysis. Remarkably, it is shown that it can be more beneficial to avoid using the true task relation in the transfer learning when a practically-convenient use of a functional identity relation provides better numerical stability. 

Dhifallah and Lu \cite{dhifallah2021phase} and Gerace et al.~\cite{gerace2021probing} provide transfer learning theories for high-dimensional models that involve nonlinearities. Their learning is regularized and therefore the source and target models do not accurately interpolate their training data, although somewhat-attenuated double descent phenomena can still be observed in \cite{gerace2021probing}.

More recently, Kim et al.~\cite{kim2025transfer} theoretically analyze linear transfer learning where the learned target model is the interpolating solution of the target training dataset of minimum distance from the source model (which by itself interpolates the source training dataset). Interestingly, their formulation can coincide with the limiting case shown in \cite{dar2024common} by taking the source-distance regularization hyperparameter to $0^+$. While \cite{dar2024common} provide a precise asymptotic analysis more focused on double descent aspects, \cite{kim2025transfer} provide a non-asymptotic analysis more focused on benign overfitting aspects other than double descent. 

The use of multiple pretrained models in overparameterized transfer learning has started to gain attention very recently. \cite{kim2025transfer} propose to select informative sources and to ensemble multiple interpolating transfer learning models. \cite{boharon2026transfer} propose to debias the overparameterization bias that  interpolating pretrained models have due to residing in the span of their training dataset; such debiasing allows the transfer learning of the target model to benefit from using many more pretrained models together.

\subsection{Pruning of learned models}
\label{subsec:pruning}

While DNNs are highly parameterized models, it is also common to prune them into less complex forms that better trade off generalization performance and computational/memory requirements. This is especially useful for applications with limitations on storage space for the learned model, as well as constraints on computation time and energy consumption for processing new inputs. 

In general, model pruning reduces the overparameterization level, and therefore the study of their interplay is of great interest. 
Chang et al.~\cite{chang2021provable} provided the first theoretical study on double descent phenomena arising in pruning of interpolating models. 
Their main focus is on a pruning approach where a dense model is trained, pruned and then retrained over the support of coordinates that were not pruned. Hence, their pruning is a sparsifying method that defines a support of nonzero values within the entire model. They consider three types of pruning rules: based on magnitude of values, based on the Hessian, and based on oracle knowledge of the best support that minimizes the test error for a desired model sparsity.

They show in~\cite{chang2021provable} that overparameterization of the dense model before pruning can lead to improved generalization in the pruned model (i.e., the improvement is with respect to~pruned models with the same sparsity level but with varying degrees of parameterization of the dense, pre-pruning model). 
Moreover, such a pruning strategy can induce better generalization than training from scratch a model of the same size as the pruned model.
They also demonstrate double descent phenomena of test errors of the pruned model as a function of the parameterization level of the dense (pre-pruning) model. 
The concrete setting that they analyze is the minimum $\ell_2$-norm solution in overparameterized linear regression on data that satisfies a spiked covariance model. 
Their theoretical results rely on an asymptotic distributional characterization of the minimum-norm solution of the dense model; specifically, they construct a Gaussian distribution with identical second-order statistics to the original data that may not be Gaussian.
They also consider a random feature model in which a first layer (of fixed weights) computes random projections that go into ReLU nonlinearities, followed by a second layer that is learned.
They consider pruning of the first layer of (non-learned) random features, which conceptually corresponds to pruning a hidden layer in a network, and show double descent phenomena of the test error as a function of the number of random features. 
This further emphasizes the finding that overparameterization can be beneficial even when pruning is applied. 

He et al.~\cite{he2022sparse} further explored the relation between pruning and overparameterization and empirically showed double descent of the generalization performance as a function of the pruning-induced sparsity level. Specifically, there is a critical phase of sparsity levels around the interpolation threshold where the corresponding test error can peak. This showcases that pruning of overparameterized models should be applied while considering the effective new parameterization level and its implications for generalization performance. 

\subsection{Dictionary learning for sparse representations}
\label{subsec:dictionary learning}

Dictionary learning is an unsupervised task of forming an overcomplete dictionary for sparse representations of a given set of data examples. 
Sulam et al.~\cite{sulam2020recovery} consider learned dictionaries that include more atoms (columns in the dictionary matrix) than the true dictionary that generated the training data. They refer to such learned dictionaries as \emph{over-realized}. Over-realized dictionaries can be learned to achieve zero training error, i.e., to provide perfect sparse representations of the training data. 
Hence, over-realized dictionaries constitute a particular type of overparameterized model.
They show that over-realized dictionaries can generalize better than those of the ground-truth size in sparse representations of new test data from the same model. Regarding estimation of the true dictionary, they show that learning a dictionary of the true size can be outperformed by first learning an over-realized dictionary and then distilling it into a dictionary of the true size. 

The results in \cite{sulam2020recovery} show that increasing the dictionary size is useful, but only up to a certain point, after which the performance continuously degrades with the increase in model size. Specifically, their results suggest that the best performance can be obtained by the smallest dictionary that interpolates the training data. Due to the underlying non-convex optimization problem in training, such an interpolating solution is achieved by an over-realized dictionary that is larger than the true dictionary.

The learned dictionary can be perceived as a mapping from sparse code vectors to their corresponding data vectors of interest (although the sparse codes are not given in the training data and, thus, the learning is unsupervised). Hence, it is interesting to note that the error curves of a learned over-realized dictionary \emph{do not exhibit double descent phenomena like in interpolating solutions of linear regression}. A possible reason for the lack of double descent in over-realized dictionaries is that their construction via well-established methods (such as K-SVD \cite{aharon2006ksvd} or  the online dictionary learning \cite{mairal2010online}) is numerically stable and structurally constrained (e.g., dictionary columns are normalized) -- this contrasts the (unconstrained) least squares from Section \ref{sec:Interpolation in Supervised Learning} that has numerical instabilities and a peaking test error around the interpolation threshold. Thus, despite the ability of the over-realized dictionaries in \cite{sulam2020recovery} to interpolate their training data, they are not necessarily minimum-norm solutions. 
An interesting research direction for future work would be to define and study over-realized dictionaries with minimum-norm properties. 

The main theoretical analysis in \cite{sulam2020recovery} is for a noiseless data model where data vectors are generated as sparse linear combinations of atoms from the unknown dictionary. Chidambaram et al.~\cite{chidambaram2023hiding} further explored over-realized dictionaries by focusing on a noisy data model where data vectors are generated as sparse linear combinations of dictionary atoms plus a noise vector. 
\cite{chidambaram2023hiding} show that, in the noisy setting, the standard reconstruction objective can fail to recover the ground-truth dictionary in the over-realized learning. They mitigate this problem by using masking in the optimization objective, in which only some of the data coordinates are used to infer the sparse representation and the remaining coordinates are used to evaluate the reconstruction loss. This demonstrates the crucial role of the optimization objective for the success of over-realized dictionary learning in noisy settings.

\section{Open frontiers}
\label{sec:Open Questions}

The TOPML research area has received considerable attention since its emergence in 2018.
While impressive progress has been made to significantly improve our understanding of overparameterized learning, some important scientific challenges remain to be resolved. 
In this section we briefly discuss a few open frontiers.

\subsection{Are there applications other than ML where interpolating solutions are useful?\nopunct}
Much of the TOPML theory of overparameterized interpolating solutions surveyed in this paper aims to explain empirical practice in modern ML, especially deep learning. Ideally, such fundamental insights may induce novel extensions to contemporary practice.
Another important question is whether the foundational theoretical findings on overparameterized interpolating solutions can lead to significant breakthroughs in domains other than ML. As an example, consider signal estimation tasks, such as signal deconvolution and inverse problems, in which one receives degraded measurements of a specific signal without any particular examples of other signals.  

In general, the relevance of interpolating solutions in various problems is not a simple wish. 
As a fundamental example, we note that ML research on interpolating solutions in supervised learning is focused on \textit{random design} settings where both the inputs and outputs are independently drawn from an unknown joint distribution; assume that the input marginal distribution has a density, which implies that all the drawn training and test inputs differ almost surely. 
Therefore, in this random design setting, interpolating the input-output mappings of the training examples does not prevent good performance on new inputs at test time. 
Supervised learning in random design settings is in accordance with contemporary ML frameworks where, indeed, interpolating solutions can provide remarkable performance. 
In contrast, classical statistics often considers a \textit{fixed design} setting, in which the inputs are prescribed (by the experiment designer) or conditioned upon while the outputs are random. Specifically, for prescribed inputs, performance may then be evaluated using test data that has the same inputs as the training data but matched to different random outputs. Hence, in fixed design settings, a mapping that interpolates the input-output training examples can fail when applied on test data. 

Let us consider the signal denoising problem to contrast a supervised denoiser-learning problem with a fixed-grid, single-signal denoising problem: 
\begin{itemize}
\item The denoising problem in the random design can take the following form. 
The given training dataset includes $n$ examples of noisy signals and their corresponding underlying, noiseless signals. Each of the $n$ examples corresponds to a \textit{different} signal from the same signal distribution. The noiseless and the noisy signals are both $p$-dimensional vectors. In this random design, the goal is to learn a denoising mapping that takes a $p$-dimensional noisy signal and outputs a $p$-dimensional estimate for the corresponding noiseless signal. It is of interest that this mapping will operate well on new noisy signals from the considered distribution, but that were not included in the training dataset. Following the main principles that we overviewed in this paper, in this case of random-design signal denoising, a learned function that interpolates the $n$ training examples does not necessarily imply poor performance on new data. 

    \item A fixed-grid, single-signal denoising problem can be defined as follows.
The given data include $n$ noisy measurements of the \textit{same} signal; each measurement is a $p$-dimensional vector of noisy values of the signal over a uniformly-spaced grid of coordinates (i.e., we consider a discrete version of a continuous-domain signal). Examples for noiseless or noisy versions of other signals are not given.
In this fixed design, the goal is to estimate the true, noiseless signal values (only) in the $p$ coordinates of the uniformly-spaced grid. 
The inputs are considered to be the $p$ coordinates of the uniformly-spaced grid. 
The estimate and the $n$ noisy measurements have the same discrete grid of $p$ coordinates. Hence, when $n>1$ and the training examples have different noisy outputs for the same inputs, interpolation is impossible; when $n=1$, interpolation provides a noise-dependent estimate that may be poor.

\end{itemize}

The above example simply demonstrates why interpolating solutions are not immediately meaningful in many applications and, accordingly, that it is not trivial to find applications other than ML where interpolating solutions are beneficial. Yet, the success of interpolation in modern ML suggests to explore its relevance to other areas.

\subsection{How should we define learned model complexity?\nopunct}
\label{subsec:Open Question - complexity definition}
The correct definition of learned model complexity is an important component of TOPML research. 
Recall that we defined a scenario to be overparameterized when the \emph{learned} model complexity is high compared to the number of training examples. 
Thus, the definition of learned model complexity is clearly crucial for understanding whether a specific learning setting is in fact overparameterized. 
While this basic decision problem (i.e., whether a setting of learning from noisy data is overparameterized or not) can be solved by identifying whether the learned model \emph{interpolates} or achieves zero training error, quantifying the actual extent of overparameterization is more challenging as well as more valuable as it allows us to distinguish various interpolating models.
The summary of results in Section~\ref{sec:regression} of this paper showed a diversity of possible generalization behaviors for interpolating models.
Whether we can interpret these results through a new overarching notion of learned model complexity remains a largely open question.

In linear models such as in LS regression, model complexity is often measured by the number of learned parameters in the solution. 
However, the double descent behavior clearly shows that the number of parameters, by itself, is a misleading measure of \textit{learned} model complexity; regularization in the learning process, as well as structural redundancies in the architecture may reduce the effective complexity of the learned model.

Moreover, defining complexity of  nonlinear and multi-layer models, including the extreme case of deep neural networks, becomes highly intricate. 
Nakkiran et al.~\cite{nakkiran2019deep} proposed to define the effective model complexity of a practical DNN training procedure as the largest dataset size for which it attains expected training error less than a prescribed numerical tolerance. However, this measure depends on data distribution, numerical tolerance, and training algorithm and indicates interpolation capacity rather than providing an intrinsic, generalization-informative complexity measure for an individual learned DNN.

Model complexity measures that depend on the training data (e.g., Rademacher complexity \citep{bartlett2002rademacher}) are useful to understand classical ML via uniform convergence analyses. In contrast, such complexity measures cannot sufficiently explain the good generalization ability in contemporary settings where the learned models interpolate their training data \citep{belkin2018understand,nagarajan2019uniform,bartlett2020failures}.
Researchers have also attempted, with relatively more success, to relate other classical notions of learned model complexity to the overparameterized regime.
Dwivedi et al.~\cite{dwivedi2020revisiting} adapted Rissanen's classical notion of minimum description length (MDL)~\citep{rissanen1978modeling,rissanen1983universal} to the overparameterized regime: their data-driven notion of MDL explains some of the behaviors observed in Section~\ref{sec:regression}. 
Algorithm-dependent notions of model complexity, such as algorithmic stability~\cite{bousquet2002stability}, are also interesting to consider; indeed, Rangamani et al.~\cite{rangamani2020interpolating} showed that in some cases, the minimum Hilbert-norm interpolation for kernel regression is also the most algorithmically stable.
Much work remains to be done to systematically study these complexity measures in the overparameterized regime and understand whether they are \emph{always} predictive of generalization behavior.
Accordingly, the definition of an appropriate complexity measure continues to pose a fundamental challenge that is at the heart of the TOPML research area.

\newpage
\section*{Acknowledgments}
\sloppy
We thank all the organizers and participants of the TOPML workshops for the extensive discussions and contributions that enriched the content in this survey.
Special thanks to Ryan Tibshirani for fruitful discussions on the definition of TOPML and the open questions of the field. 
Special thanks also to Misha Belkin for helpful comments on an early draft of this paper.

VM thanks her co-authors for several insightful discussions that inspired the treatment of TOPML in this survey; special thanks go to Anant Sahai for influencing the signal processing-oriented treatment of minimum-norm interpolation provided in Section~\ref{subsec:spexplanation} of this paper.

\sloppy
YD and RGB acknowledge support from NSF grants CCF-1911094, IIS-1838177, and IIS-1730574; ONR grants N00014-18-12571, N00014-20-1-2534, and MURI N00014-20-1-2787; AFOSR grant FA9550-18-1-0478; and a Vannevar Bush Faculty Fellowship, ONR grant N00014-18-1-2047.
RGB acknowledges support from ONR grant N00014-23-1-2714, DOI grant 140D0423C0076, and DOE grant DE-SC0020345.
VM acknowledges support from NSF grants CCF-2239151 and IIS-2212182, an Adobe Data Science Research Award, and an Amazon Research Award.

\bigskip
\appendix
\section{Proof Outline for the Bias--Variance Decomposition and the In-Class and Misspecification Components}
\label{appendix:sec:Proof Outline for the Bias--Variance Decomposition}

\subsection{The Expected Learned Predictor is Undefined for $p=n$}
\label{appendix:subsec:The Expected Learned Predictor is Undefined at the interpolation threshold}

Recall that (\ref{eq:learned linear mapping of feature vector}) formulates the learned predictor as $\widehat{f}(\vec{x}) = \vecgreek{\phi}^{T} \widehat{\vecgreek{\alpha}}$ where $\vecgreek{\phi}\triangleq \varphi\left({\vec{x}}\right)$ is the feature vector for an input $\vec{x}$, and  $\widehat{\vecgreek{\alpha}}\in\mathbb{R}^{p}$ is the learned parameter vector.

The expectation of the learned predictor satisfies $\expectationwrt{\widehat{f}(\vec{x})}{\mathcal{D}} =  \vecgreek{\phi}^{T}\expectationwrt{\widehat{\vecgreek{\alpha}}}{\mathcal{D}}$. Hence, the expected learned predictor $\expectationwrt{\widehat{f}(\vec{x})}{\mathcal{D}}$ is defined if the expected parameter vector $\expectationwrt{\widehat{\vecgreek{\alpha}}}{\mathcal{D}}$ is defined.

By (\ref{eq:example 1-2 - min-norm solution}), $\widehat{\vecgreek{\alpha}}=\mtx{\Phi}^+ \vec{y}$. Organize the training data in $\mathcal{D}=\{(\vec{x}_i, y_i)\}_{i=1}^n$ as $\mtx{X}=[\vec{x}_1,\dots,\vec{x}_n]^T$, $\vec{y}=[y_1,\dots,y_n]^T$ and write $\vec{y}=\mtx{X}\vecgreek{\beta}+\vecgreek{\epsilon}$ where $\vecgreek{\epsilon}=[\epsilon_1,\dots,\epsilon_n]^T$ are the error terms in the training examples. Recall that $\mtx{U}_{p}\triangleq[\vec{u}_1,\dots,\vec{u}_p]$; we define $\mtx{U}_{c(p)}\triangleq[\vec{u}_{p+1},\dots,\vec{u}_d]$ and get $\mtx{U}_{p}\mtx{U}_{p}^T + \mtx{U}_{c(p)} \mtx{U}_{c(p)}^T = \mtx{I}_d$. Also, recall our assumption that $\mtx{U}_d$ diagonalizes $\mtx{\Sigma}_{\vec{x}}$ and that $ \vecgreek{\xi}=\mtx{X}\mtx{U}_{c(p)} \mtx{U}_{c(p)}^T \vecgreek{\beta}$ is a zero-mean random vector independent of $\mtx{\Phi}$ and $\vecgreek{\epsilon}$. We assume here that $\sigma_{\xi}^2+\sigma_{\epsilon}^2>0$, i.e., the vector $\vecgreek{\xi}+\vecgreek{\epsilon}$ is not deterministically a zero vector. 

Moreover, recall our assumption of a full-rank feature covariance matrix $\mtx{\Sigma}_{\vecgreek{\phi}}$, which makes the Gaussian feature distribution nondegenerate and ensures that ${\rm rank}\left(\mtx{\Phi}\right) = \min\{n,p\}$ almost surely, hence the interpolation threshold is at $p=n$ rather than at a smaller effective feature dimension.

Then, the decomposition $\vec{y}=\mtx{\Phi}\mtx{U}_p^T\vecgreek{\beta} + \vecgreek{\xi}+\vecgreek{\epsilon}$ is useful for developing the expectation of the learned parameter vector:
\begin{equation}
    \label{appendix:eq:expectation of the learned predictor - intermediate}
    \expectationwrt{\widehat{\vecgreek{\alpha}}}{\mathcal{D}} = \expectationwrt{\mtx{\Phi}^+ \vec{y}}{\mtx{\Phi},\vec{y}} =  \expectationwrt{\mtx{\Phi}^+ \mtx{\Phi}\mtx{U}_p^T\vecgreek{\beta}}{\mtx{\Phi}} + \expectationwrt{\mtx{\Phi}^+ \left(\vecgreek{\xi}+\vecgreek{\epsilon}\right)}{\mtx{\Phi},\vecgreek{\xi},\vecgreek{\epsilon}} 
\end{equation}
When the entries of $\mtx{\Phi}^+$ are integrable, Fubini's theorem allows to separate the independent variables' expectations as  
\begin{equation}
\label{appendix:eq:expectation of the learned predictor - subterm for defined expectation}
\expectationwrt{\mtx{\Phi}^+ \left(\vecgreek{\xi}+\vecgreek{\epsilon}\right)}{\mtx{\Phi},\vecgreek{\xi},\vecgreek{\epsilon}}=\expectationwrt{\mtx{\Phi}^+}{\mtx{\Phi}}\expectationwrt{\vecgreek{\xi}+\vecgreek{\epsilon}}{\vecgreek{\xi},\vecgreek{\epsilon}}=\expectationwrt{\mtx{\Phi}^+}{\mtx{\Phi}}\vec{0}=\vec{0}   
\end{equation}
and the expectation of the learned parameter vector in (\ref{appendix:eq:expectation of the learned predictor - intermediate}) becomes 
\begin{equation}
    \label{appendix:eq:expectation of the learned predictor - well defined case}
    \expectationwrt{\widehat{\vecgreek{\alpha}}}{\mathcal{D}} = \expectationwrt{\mtx{\Phi}^+ \mtx{\Phi}}{\mtx{\Phi}} \mtx{U}_p^T\vecgreek{\beta}.
\end{equation}
However, for $p=n$ and nondegenerate Gaussian features, $\mtx{\Phi}^+ = \mtx{\Phi}^{-1}$ almost surely, and the entries of $\mtx{\Phi}^{-1}$ are not integrable. Then, in this Gaussian setting, if $\sigma_{\xi}^2+\sigma_{\epsilon}^2>0$ (i.e., $\vecgreek{\xi}+\vecgreek{\epsilon}$ is not a deterministically zero vector), the expectation $\expectationwrt{\mtx{\Phi}^+ \left(\vecgreek{\xi}+\vecgreek{\epsilon}\right)}{\mtx{\Phi},\vecgreek{\xi},\vecgreek{\epsilon}}$ is undefined instead of becoming zero as in (\ref{appendix:eq:expectation of the learned predictor - subterm for defined expectation}). Consequently, the expectations $\expectationwrt{\widehat{\vecgreek{\alpha}}}{\mathcal{D}}$ and $\expectationwrt{\widehat{f}(\vec{x})}{\mathcal{D}}$ are undefined for $p=n$.

\subsection{The Variance is Infinite for $p\in\{n-1,n+1\}$}
\label{appendix:subsec:Proof - variance diverges near interpolation threshold}

We will here develop the variance term from (\ref{eq:bias variance decomposition - general definiton for squared error - variance}), considering $p\neq n$ due to the undefined $\expectationwrt{\widehat{f}(\vec{x})}{\mathcal{D}}$ for $p=n$ (Appendix \ref{appendix:subsec:The Expected Learned Predictor is Undefined at the interpolation threshold}). We continue to consider the same setting, assumptions and notations (e.g., as in Appendix \ref{appendix:subsec:The Expected Learned Predictor is Undefined at the interpolation threshold}). Specifically, we use the decomposition $\vec{y}=\mtx{\Phi}\mtx{U}_p^T\vecgreek{\beta} + \vecgreek{\xi}+\vecgreek{\epsilon}$ and the expected learned predictor formula $\expectationwrt{\widehat{f}(\vec{x})}{\mathcal{D}}=\vecgreek{\phi}^T \expectationwrt{\mtx{\Phi}^+ \mtx{\Phi}}{\mtx{\Phi}} \mtx{U}_p^T\vecgreek{\beta}$. Then, 
\begin{align}
\label{appendix:eq:variance decomposition to for divergence proof}
{\sf var}\left(\widehat{f}\right) & = \mathbb{E}_{\vec{x}}\expectationwrt{\Big( \widehat{f}(\vec{x}) - \expectationwrt{\widehat{f}(\vec{x})}{\mathcal{D}} \Big)^2}{\mathcal{D}} \nonumber \\ \nonumber
& = \mathbb{E}_{\vecgreek{\phi}}\expectationwrt{\Big( \vecgreek{\phi}^T \mtx{\Phi}^+ \vec{y} - \vecgreek{\phi}^T \expectationwrt{\mtx{\Phi}^+ \mtx{\Phi}}{\mtx{\Phi}} \mtx{U}_p^T\vecgreek{\beta} \Big)^2}{\mathcal{D}} \nonumber \\
& = \mathbb{E}_{\vecgreek{\phi}}\expectationwrt{\Big( \vecgreek{\phi}^T \mtx{\Phi}^+ \mtx{\Phi}\mtx{U}_p^T\vecgreek{\beta} - \vecgreek{\phi}^T \expectationwrt{\mtx{\Phi}^+ \mtx{\Phi}}{\mtx{\Phi}} \mtx{U}_p^T\vecgreek{\beta} + \vecgreek{\phi}^T \mtx{\Phi}^+ \left(\vecgreek{\xi}+\vecgreek{\epsilon}\right) \Big)^2}{\mathcal{D}}. 
\end{align}
Note that the expectation w.r.t.~the training data $\mathcal{D}$ can be denoted here as expectation w.r.t.~$\mtx{\Phi}, \vecgreek{\epsilon}, \vecgreek{\xi}$. Then, further developments using the zero means of $\vecgreek{\epsilon}, \vecgreek{\xi}$ and their independence of $\vecgreek{\phi},\mtx{\Phi}$ yield 
\begin{align}
\label{appendix:eq:variance decomposition for divergence proof}
{\sf var}\left(\widehat{f}\right) &=  \mathbb{E}_{\vecgreek{\phi},\mtx{\Phi}}\expectationwrt{\Big( \vecgreek{\phi}^T \mtx{\Phi}^+ \mtx{\Phi}\mtx{U}_p^T\vecgreek{\beta} - \vecgreek{\phi}^T \expectationwrt{\mtx{\Phi}^+ \mtx{\Phi}}{\mtx{\Phi}} \mtx{U}_p^T\vecgreek{\beta} \Big)^2   }{\vecgreek{\epsilon}, \vecgreek{\xi} \vert\vecgreek{\phi},\mtx{\Phi}} \nonumber\\
&\quad + \mathbb{E}_{\vecgreek{\phi},\mtx{\Phi}}\expectationwrt{\Big( \vecgreek{\phi}^T \mtx{\Phi}^+ \left(\vecgreek{\xi}+\vecgreek{\epsilon}\right) \Big)^2   }{\vecgreek{\epsilon}, \vecgreek{\xi} \vert\vecgreek{\phi},\mtx{\Phi}}.
\end{align}
Note that the variance decomposition in (\ref{appendix:eq:variance decomposition for divergence proof}) is a sum of expectations over nonnegative random values.

Let us develop the second expectation from (\ref{appendix:eq:variance decomposition for divergence proof}), using that our assumptions imply $ \expectationwrt{\left(\vecgreek{\xi}+\vecgreek{\epsilon}\right)\left(\vecgreek{\xi}+\vecgreek{\epsilon}\right)^T}{\vecgreek{\epsilon}, \vecgreek{\xi}}=\left(\sigma_{\xi}^2 + \sigma_{\epsilon}^2\right)\mtx{I}_n$: 
\begin{align}
    \label{appendix:eq:variance lower bound development - for divergence proof}
    \mathbb{E}_{\vecgreek{\phi},\mtx{\Phi}}\expectationwrt{\Big( \! \vecgreek{\phi}^T \mtx{\Phi}^+ \left(\vecgreek{\xi}+\vecgreek{\epsilon}\right) \! \Big)^2  }{\vecgreek{\epsilon}, \vecgreek{\xi} \vert\vecgreek{\phi},\mtx{\Phi}} & =
    \expectationwrt{\vecgreek{\phi}^T \mtx{\Phi}^+      \expectationwrt{\left(\vecgreek{\xi}+\vecgreek{\epsilon}\right)\left(\vecgreek{\xi}+\vecgreek{\epsilon}\right)^T}{\vecgreek{\epsilon}, \vecgreek{\xi}}  \mtx{\Phi}^{+,T}\vecgreek{\phi}}{\vecgreek{\phi},\mtx{\Phi}}  \nonumber \\ \nonumber
    & =
    \left(\sigma_{\xi}^2 + \sigma_{\epsilon}^2\right)\expectationwrt{\vecgreek{\phi}^T \mtx{\Phi}^+   \mtx{\Phi}^{+,T}\vecgreek{\phi}}{\vecgreek{\phi},\mtx{\Phi}} \\ \nonumber
    & =
    \left(\sigma_{\xi}^2 + \sigma_{\epsilon}^2\right)\expectationwrt{\mtxtrace{\mtx{\Sigma}_{\vecgreek{\phi}}\left(\mtx{\Phi}^{T}\mtx{\Phi}\right)^{+}}}{\mtx{\Phi}} \\ 
    & \ge \left(\sigma_{\xi}^2 + \sigma_{\epsilon}^2\right)\lambda_{\sf min}\left(\mtx{\Sigma}_{\vecgreek{\phi}}\right)\expectationwrt{\mtxtrace{\left(\mtx{\Phi}^{T}\mtx{\Phi}\right)^{+}}}{\mtx{\Phi}} 
\end{align}
due to basic pseudoinverse properties that imply $\mtx{\Phi}^{+} \mtx{\Phi}^{+,T}=\left(\mtx{\Phi}^{T}\mtx{\Phi}\right)^{+}$, and a lower bound for a trace of matrix product (we assume here a full rank $\mtx{\Sigma}_{\vecgreek{\phi}}$ whose minimal eigenvalue is $\lambda_{\sf min}\left(\mtx{\Sigma}_{\vecgreek{\phi}}\right)>0$). Then, properties of inverse Wishart matrices and pseudoinverse of Wishart matrices imply that the expected trace in (\ref{appendix:eq:variance lower bound development - for divergence proof}) diverges for $p=n-1$ and $p=n+1$; see \cite{belkin2020two} for a treatment of a related expression in the context of test error analysis for isotropic features. 
Therefore, if $\sigma_{\xi}^2 + \sigma_{\epsilon}^2>0$ and $p\in\{n-1,n+1\}$, the second term in the variance decomposition in (\ref{appendix:eq:variance decomposition for divergence proof}) is infinite and the variance is infinite.

\subsection{Proof Outline for Eqs.~(\ref{eq:bias decomposition})-(\ref{eq:in-class bias formula}): The Bias Decomposition to In-Class and Misspecification Components}
\label{appendix:subsec:Proof - The Bias Decomposition to In-Class and Misspecification Terms}

Let us assume $p\neq n$ and develop the squared bias term from (\ref{eq:bias variance decomposition - general definiton for squared error - bias}) as follows: 
\begin{align}
\label{appendix:eq:bias decomposition - proof sketch - the decomposition}
{{\sf bias}^{2}\left(\widehat{f}\right)}  \triangleq &  \expectationwrt{\left({\expectationwrt{\widehat{f}(\vec{x})}{\mathcal{D}} - f_{\sf opt}(\vec{x})}\right)^2}{\vec{x}} \\ \nonumber
= & \expectationwrt{\left({\expectationwrt{\widehat{f}(\vec{x})}{\mathcal{D}} - f_{{\sf opt},\mathcal{F}_{p}^{\sf lin}}(\vec{x}) + f_{{\sf opt},\mathcal{F}_{p}^{\sf lin}}(\vec{x}) - f_{\sf opt}(\vec{x})}\right)^2}{\vec{x}} \\ \nonumber
= & \expectationwrt{\left({\expectationwrt{\widehat{f}(\vec{x})}{\mathcal{D}} - f_{{\sf opt},\mathcal{F}_{p}^{\sf lin}}(\vec{x})}\right)^2}{\vec{x}} + \expectationwrt{\left( {f_{{\sf opt},\mathcal{F}_{p}^{\sf lin}}(\vec{x}) - f_{\sf opt}(\vec{x})} \right)^2}{\vec{x}} \\\nonumber
& + 2 \expectationwrt{\left({\expectationwrt{\widehat{f}(\vec{x})}{\mathcal{D}} - f_{{\sf opt},\mathcal{F}_{p}^{\sf lin}}(\vec{x})}\right)\left( {f_{{\sf opt},\mathcal{F}_{p}^{\sf lin}}(\vec{x}) - f_{\sf opt}(\vec{x})} \right)}{\vec{x}}\\
= & {\sf bias}_{\sf in class}^{2}\left(\widehat{f}\right) + {\sf bias}_{\sf misspec}^{2}
\label{appendix:eq:bias decomposition - proof sketch - the decomposition - decomposition sum}
\end{align}
where the last equality is due to the misspecification bias term ${\sf bias}_{\sf misspec}^{2}$ definition in (\ref{eq:misspecification bias definition}), the in-class bias term ${\sf bias}_{\sf in class}^{2}\left(\widehat{f}\right)$ definition in (\ref{eq:in-class bias definition}), and that 
\begin{equation}
\label{appendix:eq:bias decomposition - proof sketch - zero cross term}
\expectationwrt{\left({\expectationwrt{\widehat{f}(\vec{x})}{\mathcal{D}} - f_{{\sf opt},\mathcal{F}_{p}^{\sf lin}}(\vec{x})}\right)\left( {f_{{\sf opt},\mathcal{F}_{p}^{\sf lin}}(\vec{x}) - f_{\sf opt}(\vec{x})} \right)}{\vec{x}}=0    
\end{equation}
for independent features and existing expectation $\expectationwrt{\widehat{f}(\vec{x})}{\mathcal{D}}$, as we prove next. For an input $\vec{x}$, the feature vector is $\vecgreek{\phi}= \mtx{U}_p^T \vec{x}$. Recall that $\mtx{U}_{p}\triangleq[\vec{u}_1,\dots,\vec{u}_p]$. Accordingly, by defining $\mtx{U}_{c(p)}\triangleq[\vec{u}_{p+1},\dots,\vec{u}_d]$, the vector $\vecgreek{\phi}_{c}= \mtx{U}_{c(p)}^T \vec{x}$ includes the features that are not utilized in the feature space. As $\vec{x}\sim\mathcal{N}\left(\vec{0},\mtx{\Sigma}_{\vec{x}}\right)$ is a Gaussian vector, its linear transformations $\vecgreek{\phi}$ and $\vecgreek{\phi}_{c}$ are also Gaussian vectors (with zero mean). Thus, also by our assumption that $\mtx{U}_d$ diagonalizes $\mtx{\Sigma}_{\vec{x}}$, the vectors $\vecgreek{\phi}$ and $\vecgreek{\phi}_{c}$ are independent. 
Also, note that 
\begin{equation}
\label{appendix:eq:orthogonal projection matrix sum}
\mtx{U}_{p}\mtx{U}_{p}^T + \mtx{U}_{c(p)} \mtx{U}_{c(p)}^T = \mtx{I}_d,    
\end{equation}
therefore, 
\begin{equation}
f_{\sf opt}(\vec{x})=\vec{x}^T \vecgreek{\beta} = \vec{x}^T (\mtx{U}_{p}\mtx{U}_{p}^T + \mtx{U}_{c(p)} \mtx{U}_{c(p)}^T) \vecgreek{\beta}  =  \vecgreek{\phi}^{T}\mtx{U}_p^T\vecgreek{\beta} + \vecgreek{\phi}_c^{T}\mtx{U}_{c(p)}^T\vecgreek{\beta}, 
\end{equation}
and, using (\ref{eq:optimal estimate in misspecified case of Example 1}), 
\begin{align}
\label{appendix:eq:xi variable second definition}
f_{\sf opt}(\vec{x}) - f_{{\sf opt},\mathcal{F}_{p}^{\sf lin}}(\vec{x}) & = \left({\vecgreek{\phi}^{T}\mtx{U}_p^T\vecgreek{\beta} + \vecgreek{\phi}_c^{T}\mtx{U}_{c(p)}^T\vecgreek{\beta}}\right) - \vecgreek{\phi}^{T}\mtx{U}_p^T\vecgreek{\beta} \\\nonumber
& =  \vecgreek{\phi}_c^{T}\mtx{U}_{c(p)}^T\vecgreek{\beta} \\\nonumber 
& =  \xi
\end{align}
where the random variable $\xi$ is defined in (\ref{eq:definition of the misspecification xi variable}) and is independent of $\vecgreek{\phi}$.
Then, using (\ref{eq:optimal estimate in misspecified case of Example 1}), and 
\begin{align}
\label{appendix:eq:bias decomposition - proof sketch - zero cross term - proof}
& \expectationwrt{\left({\expectationwrt{\widehat{f}(\vec{x})}{\mathcal{D}} - f_{{\sf opt},\mathcal{F}_{p}^{\sf lin}}(\vec{x})}\right)\left( {f_{{\sf opt},\mathcal{F}_{p}^{\sf lin}}(\vec{x}) - f_{\sf opt}(\vec{x})} \right)}{\vec{x}} = \\\nonumber
& =-\expectationwrt{\left({ \vecgreek{\phi}^{T} \expectationwrt{\widehat{\vecgreek{\alpha}}}{\mathcal{D}} - \vecgreek{\phi}^{T}\mtx{U}_p^T\vecgreek{\beta} }\right)\xi}{\vecgreek{\phi},\xi} \\ 
\label{appendix:eq:bias decomposition - proof sketch - zero cross term - proof - using independence}
& =-\expectationwrt{\vecgreek{\phi}^{T}}{\vecgreek{\phi}}\left({  \expectationwrt{\widehat{\vecgreek{\alpha}}}{\mathcal{D}} - \mtx{U}_p^T\vecgreek{\beta} }\right) { \expectationwrt{\xi}{\xi}} \\ \nonumber
& = 0
\end{align}
where (\ref{appendix:eq:bias decomposition - proof sketch - zero cross term - proof - using independence}) uses the statistical independence of $\vecgreek{\phi}$ and $\xi$, and the last equality is due to their zero mean. This proves (\ref{appendix:eq:bias decomposition - proof sketch - zero cross term}) and therefore (\ref{appendix:eq:bias decomposition - proof sketch - the decomposition - decomposition sum}). 

The misspecification bias formula in (\ref{eq:misspecification bias}) can be easily obtained from its definition in (\ref{eq:misspecification bias definition}) and using (\ref{appendix:eq:xi variable second definition}): 
\begin{equation}
\label{appendix:eq:misspecification bias - proof}
{\sf bias}_{\sf misspec}^{2} \triangleq \expectationwrt{\left(f_{{\sf opt},\mathcal{F}_{p}^{\sf lin}}(\vec{x}) - f_{\sf opt}(\vec{x})\right)^2}{\vec{x}} = \expectationwrt{\xi^2}{\xi} = \sigma_{\xi}^2 
\end{equation}
where $\sigma_{\xi}^2$ is defined in (\ref{eq:variance of misspecification variable}) as the variance of the zero-mean random variable $\xi$.

The in-class bias term in (\ref{eq:in-class bias formula}) can be proved as follows. 
First, by (\ref{appendix:eq:expectation of the learned predictor - well defined case}), we have $\expectationwrt{\widehat{\vecgreek{\alpha}}}{\mathcal{D}} = \expectationwrt{\mtx{\Phi}^+ \mtx{\Phi}}{\mtx{\Phi}} \mtx{U}_p^T\vecgreek{\beta}$.
Then, 
\begin{align}
\label{appendix:eq:in-class bias - proof}
{\sf bias}_{\sf in class}^{2}\left(\widehat{f}\right) \triangleq & \expectationwrt{\left(\expectationwrt{\widehat{f}(\vec{x})}{\mathcal{D}} - f_{{\sf opt},\mathcal{F}_{p}^{\sf lin}}(\vec{x})\right)^2}{\vec{x}} \\ \nonumber
& = \expectationwrt{\left(\vecgreek{\phi}^{T}\left({  \expectationwrt{\widehat{\vecgreek{\alpha}}}{\mathcal{D}} - \mtx{U}_p^T\vecgreek{\beta} }\right)\right)^2}{\vecgreek{\phi}} \\ \nonumber
& = \expectationwrt{\left( \vecgreek{\phi}^{T}\left({  \expectationwrt{\mtx{\Phi}^+ \mtx{\Phi}}{\mtx{\Phi}} \mtx{U}_p^T\vecgreek{\beta} - \mtx{U}_p^T\vecgreek{\beta} }\right) \right)^2}{\vecgreek{\phi}} \\ \nonumber
& = \expectationwrt{\left( \vecgreek{\phi}^{T}\left({  \mtx{I}_p - \expectationwrt{\mtx{\Phi}^+ \mtx{\Phi}}{\mtx{\Phi}} }\right) \mtx{U}_p^T\vecgreek{\beta}  \right)^2}{\vecgreek{\phi}} \\ \nonumber
& = \vecgreek{\beta}^T \mtx{U}_p \left({  \mtx{I}_p - \expectationwrt{\mtx{\Phi}^+ \mtx{\Phi}}{\mtx{\Phi}} }\right) \expectationwrt{ \vecgreek{\phi} \vecgreek{\phi}^{T} }{\vecgreek{\phi}} \left({  \mtx{I}_p - \expectationwrt{\mtx{\Phi}^+ \mtx{\Phi}}{\mtx{\Phi}} }\right) \mtx{U}_p^T\vecgreek{\beta}  \\ \nonumber
& = \vecgreek{\beta}^T \mtx{U}_p \bar{\mtx{P}}_{\mathcal{N}\left(\mtx{\Phi}\right)} \mtx{\Sigma}_{\vecgreek{\phi}} \bar{\mtx{P}}_{\mathcal{N}\left(\mtx{\Phi}\right)} \mtx{U}_p^T\vecgreek{\beta}  
\end{align}
where $\bar{\mtx{P}}_{\mathcal{N}\left(\mtx{\Phi}\right)} \triangleq \mtx{I}_p - \expectationwrt{\mtx{\Phi}^{+}\mtx{\Phi}}{\mtx{\Phi}}$.

\subsection{Proof Outline for Eqs.~(\ref{eq:variance decomposition})-(\ref{eq:in-class variance}): The Variance Decomposition to In-Class and Misspecification Components}
\label{appendix:subsec:Proof - The Variance Decomposition to In-Class and Misspecification Terms}

First, recall the decomposition of the learned function, $\widehat{f}\left(\vec{x}\right) =   \widehat{f}_{\sf misspec}\left(\vec{x}\right) + \widehat{f}_{\sf inclass}\left(\vec{x}\right)$, as formulated in (\ref{eq:learned funcation decomposition for devariance decomposition})-(\ref{eq:learned funcation decomposition for devariance decomposition - inclass term}). This decomposition is an immediate consequence of decomposing the training data outputs as $\vec{y} = \mtx{\Phi}\mtx{U}_p^T\vecgreek{\beta} + \vecgreek{\xi} + \vecgreek{\epsilon}$; it is justified by (\ref{appendix:eq:orthogonal projection matrix sum}) and the definition of $\vecgreek{\xi}=\mtx{X}\mtx{U}_{c(p)} \mtx{U}_{c(p)}^T\vecgreek{\beta}$ that imply $\mtx{\Phi}\mtx{U}_p^T\vecgreek{\beta} + \vecgreek{\xi} = \mtx{X}\vecgreek{\beta}$.

Let us develop the variance term from (\ref{eq:bias variance decomposition - general definiton for squared error - variance}) as follows. We continue to consider $p\notin \{n-1,n,n+1\}$ due to the undefined $\expectationwrt{\widehat{f}(\vec{x})}{\mathcal{D}}$ for $p=n$ (Appendix \ref{appendix:subsec:The Expected Learned Predictor is Undefined at the interpolation threshold}) and infinite variance for $p\in\{n-1,n+1\}$ (Appendix \ref{appendix:subsec:Proof - variance diverges near interpolation threshold}). 
\begin{align}
\label{appendix:eq:variance decomposition to in-class and misspecification terms - proof outline}
{\sf var}\left(\widehat{f}\right) = & \mathbb{E}_{\vec{x}}\expectationwrt{\Big( \widehat{f}(\vec{x}) - \expectationwrt{\widehat{f}(\vec{x})}{\mathcal{D}} \Big)^2}{\mathcal{D}} \nonumber \\ \nonumber
= & \mathbb{E}_{\vec{x}}\expectationwrt{\Big( \widehat{f}_{\sf misspec}\left(\vec{x}\right) - \expectationwrt{\widehat{f}_{\sf misspec}(\vec{x})}{\mathcal{D}} + \widehat{f}_{\sf inclass}\left(\vec{x}\right) - \expectationwrt{\widehat{f}_{\sf inclass}(\vec{x})}{\mathcal{D}}\Big)^2}{\mathcal{D}} \\ \nonumber
= & \mathbb{E}_{\vec{x}}\expectationwrt{\Big( \widehat{f}_{\sf misspec}(\vec{x}) - \expectationwrt{\widehat{f}_{\sf misspec}(\vec{x})}{\mathcal{D}} \Big)^2}{\mathcal{D}} \\\nonumber
& + \mathbb{E}_{\vec{x}}\expectationwrt{\Big( \widehat{f}_{\sf inclass}(\vec{x}) - \expectationwrt{\widehat{f}_{\sf inclass}(\vec{x})}{\mathcal{D}} \Big)^2}{\mathcal{D}} \\
& + 2 \mathbb{E}_{\vec{x}}\expectationwrt{\Big( \widehat{f}_{\sf misspec}(\vec{x}) - \expectationwrt{\widehat{f}_{\sf misspec}(\vec{x})}{\mathcal{D}} \Big)\Big( \widehat{f}_{\sf inclass}(\vec{x}) - \expectationwrt{\widehat{f}_{\sf inclass}(\vec{x})}{\mathcal{D}} \Big)}{\mathcal{D}}
\end{align}
To further develop the last expression, the following results are of interest. 
The expectation of the $\widehat{f}_{\sf misspec}\left(\vec{x}\right)$ is zero: 
\begin{equation}
    \label{appendix:eq:expectation of fmisspec}
     \expectationwrt{\widehat{f}_{\sf misspec}(\vec{x})}{\mathcal{D}} =  \expectationwrt{ \vec{x}^T \mtx{U}_p \mtx{\Phi}^+ \vecgreek{\xi} }{\mtx{\Phi},\vecgreek{\xi}} = \vec{x}^T \mtx{U}_p\expectationwrt{ \mtx{\Phi}^+}{\mtx{\Phi}}\expectationwrt{ \vecgreek{\xi} }{\vecgreek{\xi}} = \vec{x}^T \mtx{U}_p\expectationwrt{ \mtx{\Phi}^+}{\mtx{\Phi}}\vec{0} = 0
\end{equation}
due to (\ref{eq:learned funcation decomposition for devariance decomposition - misspecification term}) and that $\vecgreek{\xi}$ is a zero-mean vector independent of $\mtx{\Phi}$. 

Next, the expectation of $\widehat{f}_{\sf inclass}\left(\vec{x}\right)$ is 
\begin{equation}
    \label{appendix:eq:expectation of finclass}
     \expectationwrt{\widehat{f}_{\sf inclass} (\vec{x})}{\mathcal{D}} =  \vec{x}^T \mtx{U}_p \expectationwrt{\mtx{\Phi}^+ \left(\mtx{\Phi}\mtx{U}_p^T\vecgreek{\beta} + \vecgreek{\epsilon} \right)}{\mtx{\Phi},\vecgreek{\epsilon}}   =  \vec{x}^T \mtx{U}_p \expectationwrt{\mtx{\Phi}^+ \mtx{\Phi} }{\mtx{\Phi}} \mtx{U}_p^T\vecgreek{\beta}.
\end{equation}
By (\ref{appendix:eq:expectation of fmisspec})-(\ref{appendix:eq:expectation of finclass}) and that $\vecgreek{\xi}$ is a zero-mean vector independent of the other random elements, it can be proved that the cross-product term in (\ref{appendix:eq:variance decomposition to in-class and misspecification terms - proof outline}) is zero when $p\notin\{n-1,n,n+1\}$: 
\begin{align}
    \label{appendix:eq:variance decomposition to in-class and misspecification terms - proof outline - cross product term}
    \mathbb{E}_{\vec{x}}\expectationwrt{\Big( \widehat{f}_{\sf misspec}(\vec{x}) - \expectationwrt{\widehat{f}_{\sf misspec}(\vec{x})}{\mathcal{D}} \Big)\Big( \widehat{f}_{\sf inclass}(\vec{x}) - \expectationwrt{\widehat{f}_{\sf inclass}(\vec{x})}{\mathcal{D}} \Big)}{\mathcal{D}} = 0.
\end{align}
This completes the proof outline for the variance decomposition in (\ref{eq:variance decomposition})-(\ref{eq:in-class variance - definition}). 

The misspecification variance formulation in (\ref{eq:misspecification variance}) is proved using (\ref{eq:misspecification variance - definition}), (\ref{eq:learned funcation decomposition for devariance decomposition - misspecification term}) and (\ref{appendix:eq:expectation of fmisspec}) as follows. 
\begin{align}
\label{appendix:eq:misspecification variance - formula - proof}
{\sf var}_{\sf misspec}\left(\widehat{f}\right) = & \mathbb{E}_{\vec{x}}\expectationwrt{\Big( \widehat{f}_{\sf misspec}(\vec{x})\Big)^2}{\mathcal{D}} = \mathbb{E}_{\vec{x}}\expectationwrt{\Big( \vec{x}^T \mtx{U}_p \mtx{\Phi}^+ \vecgreek{\xi}\Big)^2}{\mathcal{D}} \\\nonumber 
 = & \expectationwrt{  \vecgreek{\xi}^T \mtx{\Phi}^{+,T}  \mtx{\Sigma}_{\vecgreek{\phi}} \mtx{\Phi}^+ \vecgreek{\xi} }{\mtx{\Phi},\vecgreek{\xi}} \\\nonumber 
 = & \mtxtrace{\expectationwrt{ \vecgreek{\xi}\vecgreek{\xi}^T }{\vecgreek{\xi}} \expectationwrt{ \mtx{\Phi}^{+,T}   \mtx{\Sigma}_{\vecgreek{\phi}}  \mtx{\Phi}^+ }{\mtx{\Phi}} } \\\nonumber 
= & \sigma_{\xi}^{2} \mtxtrace{\mtx{\Sigma}_{\vecgreek{\phi}}\expectationwrt{\left(\mtx{\Phi}^{T}\mtx{\Phi}\right)^{+}}{\mtx{\Phi}} }    
\end{align}
due to the covariance matrix of $\vecgreek{\xi}$ being $\sigma_{\xi}^{2}\mtx{I}_n$, and basic pseudoinverse properties that imply $\mtx{\Phi}^{+} \mtx{\Phi}^{+,T}=\left(\mtx{\Phi}^{T}\mtx{\Phi}\right)^{+}$. 

The in-class variance formulation in (\ref{eq:in-class variance}) is proved using (\ref{eq:in-class variance - definition}), (\ref{eq:learned funcation decomposition for devariance decomposition - inclass term}) and (\ref{appendix:eq:expectation of finclass}) as follows. 
\begin{align}
{\sf var}_{\sf inclass}\left(\widehat{f}\right) = & \mathbb{E}_{\vec{x}}\expectationwrt{\Big( \widehat{f}_{\sf inclass}\left(\vec{x}\right) - \expectationwrt{\widehat{f}_{\sf inclass}\left(\vec{x}\right)}{\mathcal{D}} \Big)^2}{\mathcal{D}} \nonumber\\ \nonumber
= & \mathbb{E}_{\vec{x}}\expectationwrt{ \Big( \vec{x}^T \mtx{U}_p \mtx{\Phi}^{+}\vecgreek{\epsilon} \Big)^2}{\mathcal{D}}  \\ \nonumber
&+ \mathbb{E}_{\vec{x}}\expectationwrt{ \Big( \vec{x}^T \mtx{U}_p \left(\mtx{\Phi}^{+}\mtx{\Phi} - \expectationwrt{\mtx{\Phi}^{+}\mtx{\Phi}}{\mtx{\Phi}}\right) \mtx{U}_p^T \vecgreek{\beta} \Big)^2}{\mathcal{D}}\\ \nonumber
&+ 2 \mathbb{E}_{\vec{x}}\expectationwrt{ \Big( \vec{x}^T \mtx{U}_p \left(\mtx{\Phi}^{+}\mtx{\Phi} - \expectationwrt{\mtx{\Phi}^{+}\mtx{\Phi}}{\mtx{\Phi}}\right) \mtx{U}_p^T \vecgreek{\beta} \Big) \vec{x}^T \mtx{U}_p \mtx{\Phi}^{+}\vecgreek{\epsilon}}{\mathcal{D}} \\ 
\label{appendix:eq:inclass variance - formula - proof - equality with multiple explanations}
= & \sigma_{\epsilon}^{2} \mtxtrace{\mtx{\Sigma}_{\vecgreek{\phi}}\expectationwrt{\left(\mtx{\Phi}^{T}\mtx{\Phi}\right)^{+}}{\mtx{\Phi}}} 
\\ \nonumber
& + \vecgreek{\beta}^T \mtx{U}_p {\expectationwrt{ \left(\mtx{\Phi}^{+}\mtx{\Phi} - \expectationwrt{\mtx{\Phi}^{+}\mtx{\Phi}}{\mtx{\Phi}}\right) \mtx{\Sigma}_{\vecgreek{\phi}}    \left(\mtx{\Phi}^{+}\mtx{\Phi} - \expectationwrt{\mtx{\Phi}^{+}\mtx{\Phi}}{\mtx{\Phi}}\right)   }{\mtx{\Phi}}} \mtx{U}_p^T \vecgreek{\beta}
\end{align}
where the last equality in (\ref{appendix:eq:inclass variance - formula - proof - equality with multiple explanations}) stems from proving that 
\begin{align}
    \label{appendix:eq:inclass variance - formula - proof - auxiliary result 1}
    \mathbb{E}_{\vec{x}}\expectationwrt{ \Big( \vec{x}^T \mtx{U}_p \mtx{\Phi}^{+}\vecgreek{\epsilon} \Big)^2}{\mathcal{D}} = \sigma_{\epsilon}^{2} \mtxtrace{\mtx{\Sigma}_{\vecgreek{\phi}}\expectationwrt{\left(\mtx{\Phi}^{T}\mtx{\Phi}\right)^{+}}{\mtx{\Phi}}},
\end{align}
\begin{align}
    \label{appendix:eq:inclass variance - formula - proof - auxiliary result 2}
    &\mathbb{E}_{\vec{x}}\expectationwrt{ \Big( \vec{x}^T \mtx{U}_p \left(\mtx{\Phi}^{+}\mtx{\Phi} - \expectationwrt{\mtx{\Phi}^{+}\mtx{\Phi}}{\mtx{\Phi}}\right) \mtx{U}_p^T \vecgreek{\beta} \Big)^2}{\mathcal{D}} = \\\nonumber
    &\quad\vecgreek{\beta}^T \mtx{U}_p {\expectationwrt{ \left(\mtx{\Phi}^{+}\mtx{\Phi} - \expectationwrt{\mtx{\Phi}^{+}\mtx{\Phi}}{\mtx{\Phi}}\right) \mtx{\Sigma}_{\vecgreek{\phi}}    \left(\mtx{\Phi}^{+}\mtx{\Phi} - \expectationwrt{\mtx{\Phi}^{+}\mtx{\Phi}}{\mtx{\Phi}}\right)   }{\mtx{\Phi}}} \mtx{U}_p^T \vecgreek{\beta},
\end{align}
and that 
\begin{align}
    \label{appendix:eq:inclass variance - formula - proof - auxiliary result 3}
    \mathbb{E}_{\vec{x}}\expectationwrt{ \Big( \vec{x}^T \mtx{U}_p \left(\mtx{\Phi}^{+}\mtx{\Phi} - \expectationwrt{\mtx{\Phi}^{+}\mtx{\Phi}}{\mtx{\Phi}}\right) \mtx{U}_p^T \vecgreek{\beta} \Big) \vec{x}^T \mtx{U}_p \mtx{\Phi}^{+}\vecgreek{\epsilon}}{\mathcal{D}} = 0
\end{align}
due to the zero-mean random vector $\vecgreek{\epsilon}$ which is independent of other random elements.

\section{Proof Outline for the Noise and Signal Error Terms}
\label{appendix:sec:Proof Outline for the Noise and Signal Error Terms}

In this section, we present the proofs of the expressions for the expectation of the noise and signal error terms over the training dataset for the following setting:
\begin{itemize}
    \item The input data vector is $d$-dimensional isotropic Gaussian $\vec{x}\sim\mathcal{N}(\vec{0},\mtx{I}_d)$.
    \item The feature space is defined by Example \ref{example:function class is linear mappings with p parameters}. The orthonormal feature map implies that the feature vector is $p$-dimensional isotropic Gaussian $\vecgreek{\phi}\sim\mathcal{N}(\vec{0},\mtx{I}_p)$.
    \item For the signal error proof in Section \ref{appendix:subsec:proof-signal-error-term}, we further assume a value of $p\in\{1,\dots,d\}$ for which the $p$-dimensional feature space defines a well-specified model in the sense that, for all $\vec{x}\in\mathbb{R}^d$, $f_{\sf true}(\vec{x}) = \vec{x}^T \vecgreek{\beta}=\vecgreek{\phi}^T \widetilde{\vecgreek{\beta}}$ where $\vecgreek{\phi}=\mtx{U}_p^T\vec{x}$, $\widetilde{\vecgreek{\beta}}=\mtx{U}_p^T\vecgreek{\beta}$, and $\vecgreek{\beta}=\mtx{U}_p \mtx{U}_p^T\vecgreek{\beta}=\mtx{U}_p \widetilde{\vecgreek{\beta}}$.
\end{itemize}

\subsection{Proof of noise error term Eq.~\eqref{eq:noise error term bound expectation over training data - isotropic Gaussian}}\label{appendix:subsec:proof-noise-error-term}

Recall that we defined $\widehat{f}_{\sf noise}(\vec{x}) \triangleq \vecgreek{\phi}^T \alphamintwo_{\sf noise}$, where $\alphamintwo_{\sf noise} \triangleq \mtx{\Phi}^+ \vecgreek{\epsilon}$.
Therefore, by the definition of $\testerrnoise$, 
\begin{align*}
    \testerrnoise \triangleq \EE_{\vec{x}} \left[(\widehat{f}_{\sf noise}(\vec{x}))^2\right] &= \EE_{\vecgreek{\phi}} \left[\vecgreek{\epsilon}^T (\mtx{\Phi}^+)^T \vecgreek{\phi} \vecgreek{\phi}^T \mtx{\Phi}^+ \vecgreek{\epsilon}\right] \\
    &= \vecgreek{\epsilon}^T (\mtx{\Phi}^+)^T \mtx{\Phi}^+ \vecgreek{\epsilon},
\end{align*}
where $\vecgreek{\phi}$ is a test feature vector independent of training data elements such as $\mtx{\Phi}$ and $\vecgreek{\epsilon}$; the last equality follows due to the isotropy of $\vecgreek{\phi}$ (note that this step holds for any isotropic distribution on $\vecgreek{\phi}$, not necessarily only Gaussian).
Now, we will take the outer expectation of $\testerrnoise$ over the training dataset.
Because we assume i.i.d. training examples (and therefore i.i.d.~noise components for these examples), we note that $\EE_{\mathcal{D}}[\vecgreek{\epsilon}\vecgreek{\epsilon}^T] = \sigma_{\epsilon}^2 \mtx{I}_n$.
Then, we have
\begin{align}
    \EE_{\mathcal{D}}\left[\testerrnoise\right] &= \EE_{\mathcal{D}}\left[\vecgreek{\epsilon}^T (\mtx{\Phi}^+)^T \mtx{\Phi}^+ \vecgreek{\epsilon}\right] \\\nonumber
    &= \EE_{\mathcal{D}} \left[\mtxtrace{(\mtx{\Phi}^+)^T \mtx{\Phi}^+ \vecgreek{\epsilon} \vecgreek{\epsilon}^T}\right] \\\nonumber
    &= \EE_{\mtx{\Phi}} \left[\mtxtrace{(\mtx{\Phi}^+)^T \mtx{\Phi}^+ \EE_{\vecgreek{\epsilon}}\left[\vecgreek{\epsilon} \vecgreek{\epsilon}^T\right]}\right] \\ \nonumber
    &= \sigma_{\epsilon}^2 \EE_{\mtx{\Phi}} \left[\mtxtrace{(\mtx{\Phi}^+)^T \mtx{\Phi}^+}\right] \nonumber \\ \nonumber
    &= \sigma_{\epsilon}^2 \mtxtrace{\EE_{\mtx{\Phi}}\left[(\mtx{\Phi}^+)^T \mtx{\Phi}^+\right]}
\end{align}
where this chain of equalities uses the cyclic property and linearity of the expectation of the trace as well as the aforementioned isotropic property of the noise vector $\vecgreek{\epsilon}$ and the independence of $\vecgreek{\epsilon}$ and $\mtx{\Phi}$.
Next, by using a Moore-Penrose pseudoinverse identity, observe that $(\mtx{\Phi}^+)^T \mtx{\Phi}^+ = (\mtx{\Phi}\mtx{\Phi}^T)^{-1} \mtx{\Phi} \mtx{\Phi}^T (\mtx{\Phi} \mtx{\Phi}^T)^{-1} = (\mtx{\Phi} \mtx{\Phi}^T)^{-1}$, which holds almost surely due to rank $n$ of $\mtx{\Phi}$. Finally, we will use the fact that $\{\vecgreek{\phi}_i\}_{i=1}^n$ are i.i.d. isotropic Gaussian in order to compute the expectation of the resulting \emph{inverse Wishart matrix} $(\mtx{\Phi} \mtx{\Phi}^T)^{-1}$.
In particular, since the $n\times p$ matrix $\mtx{\Phi}$ consists of i.i.d.~$\mathcal{N}(0,1)$ entries, its expectation is $\EE[(\mtx{\Phi}\mtx{\Phi}^T)^{-1}] = \frac{\mtx{I}_n}{p - n - 1}$ provided that $p > n + 1$.
Substituting this in the above display yields
\begin{equation}
\EE_{\mathcal{D}}\left[\testerrnoise\right] = \sigma_{\epsilon}^2 \mtxtrace{\frac{\mtx{I}_n}{p - n - 1}} = \frac{\sigma_{\epsilon}^2 n}{p - n - 1},    
\end{equation}
which is the desired statement.

\subsection{Proof of signal error term Eq.~\eqref{eq:signal error term bound - isotropic covariance - well specified - in expectation}}\label{appendix:subsec:proof-signal-error-term}

Recall that we defined $\widehat{f}_{\sf sig}(\vec{x}) \triangleq \vecgreek{\phi}^T \alphamintwo_{\sf sig}$, where $\alphamintwo_{\sf sig} \triangleq \mtx{\Phi}^+ (\vec{y} - \vecgreek{\epsilon})$.
We consider here a well-specified linear model where $f_{\sf true}(\vec{x}) = \vec{x}^T \vecgreek{\beta}= \vecgreek{\phi}^T \widetilde{\vecgreek{\beta}}$, $\widetilde{\vecgreek{\beta}}=\mtx{U}_p^T\vecgreek{\beta}$; therefore, the data model can be written as $\vec{y} = \mtx{X} \vecgreek{\beta} + \vecgreek{\epsilon} = \mtx{\Phi} \widetilde{\vecgreek{\beta}} + \vecgreek{\epsilon}$.
Hence, we have $\widehat{f}_{\sf sig}(\vec{x}) = \vecgreek{\phi}^T \mtx{\Phi}^+ \mtx{\Phi} \widetilde{\vecgreek{\beta}}$.

Now, recall that we defined 
\begin{align*}
    \testerrsignal \triangleq \EE_{\vec{x}} \left[(\widehat{f}_{\sf sig}(\vec{x}) - f_{\sf true}(\vec{x}))^2\right] &= \EE_{\vecgreek{\phi}} \left[ \widetilde{\vecgreek{\beta}}^T (\mtx{\Phi}^+ \mtx{\Phi} - \mtx{I}_p)^T \vecgreek{\phi}\vecgreek{\phi}^T (\mtx{\Phi}^+ \mtx{\Phi} - \mtx{I}_p) \widetilde{\vecgreek{\beta}}\right] \\
    &= \widetilde{\vecgreek{\beta}}^T (\mtx{\Phi}^+ \mtx{\Phi} - \mtx{I}_p)^T (\mtx{\Phi}^+ \mtx{\Phi} - \mtx{I}_p) \widetilde{\vecgreek{\beta}}
\end{align*}
where the last equality follows due to the isotropy of $\vecgreek{\phi}$.
Now, we will take the outer expectation of $\testerrsignal$ over the training dataset.
Note that the expression for $\testerrsignal$ depends on the training dataset only through the feature matrix $\mtx{\Phi}$.
Therefore, we have
\begin{align*}
    \EE_{\mathcal{D}}[\testerrsignal] &= \EE_{\mtx{\Phi}}\left[\widetilde{\vecgreek{\beta}}^T (\mtx{\Phi}^+ \mtx{\Phi} - \mtx{I}_p)^T (\mtx{\Phi}^+ \mtx{\Phi} - \mtx{I}_p) \widetilde{\vecgreek{\beta}}\right] \\
    &= \EE_{\mtx{\Phi}} \left[\mtxtrace{\widetilde{\vecgreek{\beta}}\widetilde{\vecgreek{\beta}}^T (\mtx{\Phi}^+ \mtx{\Phi} - \mtx{I}_p)^T (\mtx{\Phi}^+ \mtx{\Phi} - \mtx{I}_p)}\right] \\
    &= \mtxtrace{\widetilde{\vecgreek{\beta}}\widetilde{\vecgreek{\beta}}^T \EE_{\mtx{\Phi}}[(\mtx{\Phi}^+ \mtx{\Phi} - \mtx{I}_p)^T (\mtx{\Phi}^+ \mtx{\Phi} - \mtx{I}_p)]},
\end{align*}
where we used the cyclic and linearity properties of the trace and expectation.
Now, we will calculate the inner expectation $\EE_{\mtx{\Phi}}[(\mtx{\Phi}^+ \mtx{\Phi} - \mtx{I}_p)^T (\mtx{\Phi}^+ \mtx{\Phi} - \mtx{I}_p)]$.
First, observe that the matrix $\mtx{\Phi}^+ \mtx{\Phi} - \mtx{I}_p$ is symmetric.
Moreover, it is easy to verify that $(\mtx{\Phi}^+ \mtx{\Phi})^2 = \mtx{\Phi}^+\mtx{\Phi}$; i.e., this is an orthogonal projection matrix on the row space of $\mtx{\Phi}$. 
Therefore, $\mtx{I}_p - \mtx{\Phi}^+ \mtx{\Phi}$ is also an orthogonal projection matrix. 
\begin{equation}
(\mtx{\Phi}^+ \mtx{\Phi} - \mtx{I}_p)^T (\mtx{\Phi}^+ \mtx{\Phi} - \mtx{I}_p) = \mtx{\Phi}^+\mtx{\Phi} - \mtx{\Phi}^+\mtx{\Phi} - \mtx{\Phi}^+\mtx{\Phi} + \mtx{I}_p = \mtx{I}_p - \mtx{\Phi}^+ \mtx{\Phi}.    
\end{equation}

Since the entries of $\mtx{\Phi}$ are i.i.d. $\mathcal{N}(0,1)$, the rotational symmetry of the isotropic Gaussian distribution implies \cite{belkin2020two} that 
\begin{equation}
\label{appendix:eq:expected projection matrix for isotropic gaussian features}
\EE_{\mtx{\Phi}}\left[\mtx{\Phi}^+\mtx{\Phi}\right] = \frac{n}{p} \mtx{I}_p.    
\end{equation}

Also, due to the well-specification property of $\vecgreek{\beta}=\mtx{U}_p \mtx{U}_p^T\vecgreek{\beta}=\mtx{U}_p \widetilde{\vecgreek{\beta}}$, we get that 
\begin{equation}
\label{appendix:eq:equality of true signal norms for well specified settings}
    \Ltwonorm{\vecgreek{\beta}}=\Ltwonorm{\widetilde{\vecgreek{\beta}}}.
\end{equation}
Using (\ref{appendix:eq:expected projection matrix for isotropic gaussian features}) and (\ref{appendix:eq:equality of true signal norms for well specified settings}) in the above display for $\EE_{\mathcal{D}}[\testerrsignal]$ yields
\[
\EE_{\mathcal{D}}[\testerrsignal] = \mtxtrace{\widetilde{\vecgreek{\beta}}\widetilde{\vecgreek{\beta}}^T \left(1 - \frac{n}{p}\right) \mtx{I}_p} = \Ltwonorm{\widetilde{\vecgreek{\beta}}} \left(1 - \frac{n}{p}\right) = \Ltwonorm{\vecgreek{\beta}} \left(1 - \frac{n}{p}\right) ,
\]
which is the desired statement.

\bibliographystyle{siamplain}
\bibliography{topml_overview_references}
\end{document}